\documentclass{article}

\usepackage{microtype}
\usepackage{graphicx}
\usepackage{subfigure}
\usepackage{xcolor}
\usepackage[hyphens]{url}
\usepackage{booktabs} % for professional tables

\usepackage{listings}
\usepackage{hyperref}

\usepackage[accepted]{sty/icml2020}

\usepackage{amsmath,amsfonts,amsthm,amssymb}
\usepackage{bm}
\usepackage{xspace}
\usepackage{sty/notation}
\usepackage{enumitem,array,booktabs}
\usepackage{todonotes}
\usepackage{dsfont}
\usepackage{natbib}
\usepackage{subfiles}
\usepackage{makecell}

\DeclareMathOperator*{\argmax}{argmax}
\DeclareMathOperator*{\argmin}{argmin}
\DeclareMathOperator{\N}{\mathbb N}
\DeclareMathOperator{\KL}{\mathrm{KL}}
\DeclareMathOperator{\kl}{\mathrm{kl}}

\let\inf\undefined
\let\min\undefined
\let\max\undefined
\DeclareMathOperator*{\inf}{\vphantom{sup}inf}
\DeclareMathOperator*{\min}{\vphantom{sup}min}
\DeclareMathOperator*{\max}{\vphantom{sup}max}

\let\top\intercal

\DeclareBoldMathCommand{\vmu}{$\mu$}
\DeclareBoldMathCommand{\vlambda}{$\lambda$}
\DeclareBoldMathCommand{\vtheta}{$\theta$}
\DeclareBoldMathCommand{\vxi}{$\xi$}
\DeclareBoldMathCommand{\veta}{$\eta$}
\DeclareBoldMathCommand{\vsigma}{$\sigma$}
\DeclareBoldMathCommand{\a}{a}
\DeclareBoldMathCommand{\b}{b}
\DeclareBoldMathCommand{\e}{e}
\DeclareBoldMathCommand{\p}{p}
\DeclareBoldMathCommand{\q}{q}
\DeclareBoldMathCommand{\u}{u}
\DeclareBoldMathCommand{\w}{w}
\DeclareBoldMathCommand{\x}{x}
\DeclareBoldMathCommand{\y}{y}

\newtheorem{lemma}{Lemma}
\newtheorem{theorem}{Theorem}
\newtheorem{remark}{Remark}
\theoremstyle{definition}
\newtheorem{definition}{Definition}

\icmltitlerunning{Gamification of Pure Exploration for Linear Bandits}

\begin{document}

\twocolumn[
\icmltitle{Gamification of Pure Exploration for Linear Bandits}



\icmlsetsymbol{equal}{*}

\begin{icmlauthorlist}
\icmlauthor{R\'emy Degenne}{equal,a}
\icmlauthor{Pierre M\'enard}{equal,b}
\icmlauthor{Xuedong Shang}{c}
\icmlauthor{Michal Valko}{d}
\end{icmlauthorlist}

\icmlaffiliation{a}{INRIA - DIENS - PSL Research University, Paris, France}
\icmlaffiliation{b}{INRIA}
\icmlaffiliation{c}{INRIA - Universit\'e de Lille}
\icmlaffiliation{d}{DeepMind Paris}

\icmlcorrespondingauthor{R\'emy Degenne}{remydegenne@gmail.com}

%\icmlkeywords{Machine Learning, ICML, Bandits, Linear}

\vskip 0.3in
]



\printAffiliationsAndNotice{\icmlEqualContribution} % otherwise use the standard text.

\begin{abstract}
  We investigate an active \emph{pure-exploration} setting, that includes \emph{best-arm identification}, in the context of \emph{linear stochastic bandits}.  While asymptotically optimal algorithms exist for standard \emph{multi-arm bandits}, the existence of such algorithms for the best-arm identification in linear bandits has been elusive despite several attempts to address it.  First, we provide a thorough comparison and new insight over different notions of optimality in the linear case, including G-optimality, transductive optimality from optimal experimental design and asymptotic optimality.  Second, we design the first asymptotically optimal algorithm for fixed-confidence pure exploration in linear bandits. As a consequence, our algorithm naturally bypasses the pitfall caused by a simple but difficult instance, that most prior algorithms had to be engineered to deal with explicitly.  Finally, we avoid the need to fully solve an optimal design problem by providing an approach that entails an efficient implementation.
  \end{abstract}

% We investigate \emph{pure exploration} game, among others, \emph{best-arm identification}, in the context of \emph{finite-armed linear stochastic bandits}. That being long considered as a challenging problem where a simple but sinister problem instance can fool easily usual techniques employed in the classical multi-armed bandits (MAB). Moreover, numerous a
%
%
% Ad hoc workarounds are often required to overcome that pitfall in previous work, and do not achieve \emph{asymptotic optimality}. In this paper, we design the first asymptotically optimal algorithm for general pure exploration in linear bandits. We show that a such asymptotic optimality is able to avoid the aforementioned pitfall naturally. Moreover, we provide a thorough comparison and new insight over different notions of optimality in the linear case, including G-optimality, transductive optimality and asymptotic optimality.
%!TEX root = ../lin_bandit_explo.tex
\vspace{-0.1cm}
\section{Introduction}\label{sec:introduction}
Multi-armed bandits (MAB) probe fundamental \emph{exploration-exploitation} trade-offs in sequential decision learning. We study the pure exploration framework, from among different MAB models, which is subject to the maximization of information gain after an exploration phase. We are particularly interested in the case where noisy linear payoffs depending on some regression parameter $\theta$ are assumed. Inspired by~\citet{degenne2019game}, we treat the problem as a \emph{two-player zero-sum} game between the agent and the nature (in a sense described in Section~\ref{sec:lower_bound}), and we search for algorithms that are able to output a correct answer with high confidence to a given query using as few samples as possible.

Since the early work of~\citet{robbins1952}, a great amount of literature explores MAB in their standard stochastic setting with its numerous extensions and variants. \citet{even-dar2002pac} and \citet{bubeck2009pure} are among the first to study the pure exploration setting for stochastic bandits. A non-exhaustive list of pure exploration game includes best-arm identification (BAI), top-m identification~\cite{kalyanakrishnan2010}, threshold bandits~\cite{locatelli2016thresholding}, minimum threshold~\cite{kaufmann2018murphy}, signed bandits~\cite{menard2019lma}, pure exploration combinatorial bandits~\cite{chen2014combinatorial}, Monte-Carlo tree search \cite{teraoka2014mc}, etc.

In this work, we consider a general pure-exploration setting (see Appendix~\ref{app:examples} for details). Nevertheless, for the sake of simplicity, in the main text we primarily focus on BAI.  For stochastic bandits, BAI has been studied within two major theoretical frameworks. The first one,  \emph{fixed-budget} BAI, aims at minimizing the probability of misidentifying the optimal arm within a given number of pulls~\citep{audibert2010budget}. In this work, we consider another setting, \emph{fixed-confidence} BAI, introduced by~\citet{even-dar2003confidence}. Its goal is to ensure that the algorithm returns a wrong arm with probability less than a given risk level, while using a small total number of samples before making the decision. %Note that these two frameworks are very different in general and do not share transferable regret bounds (see~\citealt{carpentier2016budget} for an additional discussion).
%Existing fixed-confidence algorithms mostly depend on the risk parameter $\delta \in (0,1)$.
Existing fixed-confidence algorithms are either elimination-based such as \SE~\citep{karnin2013sha}, rely on confidence intervals such as \UGapE~\citep{gabillon2012ugape}, or follow plug-in estimates of the optimal pulling proportions by a lower bound such as \Track~\citep{garivier2016tracknstop}. We pay particular attention to the first two since they have been extended to the linear setting, which is the focus of this paper. In particular, a natural extension of pure exploration to linear bandits. Linear bandits were first investigated by~\citet{auer2002linear} in the stochastic setting for \emph{regret minimization} and later considered for fixed-confidence BAI problems by~\citet{soare2014linear}.
%\vspace{-0.4cm}
\paragraph{Linear bandits.}
We consider a \emph{finite-arm linear bandit} problem, where the collection of arms $\cA\subset \R^d$ is given  with $|\cA|=A$, and spans $\R^d$. We assume that $\forall a\in\cA, \normm{a}\leq L$, where $\normm{a}$ denotes the Euclidean norm of the vector $a$. The learning protocol goes as follows: for each round $1\leq t \leq T,$ the agent chooses an arm $a_t\in\cA$ and observes a noisy sample
\[
Y_t =\langle \theta,a_t\rangle +\eta_t\,,
\]
where $\eta_t \sim \cN(0,\sigma^2)$ is conditionally independent from the past and $\theta$ is some unknown regression parameter. For the sake of simplicity, we use $\sigma^2 = 1$ in the rest of this paper.

\paragraph{Pure exploration for linear bandits.}
We assume that $\theta$ belongs to some set $\cM\subset\R^d$ known to the agent. % s.t.\,$\forall \theta\in\cM, \normm{\theta}\leq M$.
For each parameter a \emph{unique correct answer} is given by the function $\istar: \cM \to \cI$ among the $I = |\cI|$ possible ones (the extension of pure exploration to multiple correct answers is studied by~\citealt{degenne2019pure}). Given a parameter~$\theta$, the agent then aims to find the correct answer $\istar(\theta)$ by interacting with the finite-armed linear bandit environment parameterized by~$\theta$.

In particular, we detail the setting of BAI for which the objective is to identify the arm with the largest mean. That is, the correct answer is given by $\istar(\theta)=\astar(\theta)\eqdef \argmax_{a\in\cA} \langle\theta,a\rangle$ for $\theta\in\cM = \R^d$ and the set of possible correct answers is $\cI = \cA$. We provide other pure-exploration examples in Appendix~\ref{app:examples}.

\paragraph{Algorithm.}
Let $\cF_{t}=\sigma (a_1,Y_1,\ldots, a_t,Y_t)$ be the information available to the agent after $t$ round. A deterministic pure-exploration algorithm under the fixed-confidence setting is given by three components: (1) a \emph{sampling rule} $(a_t)_{t\geq 1}$, where $a_t\in\cA$ is $\cF_{t-1}$-measurable, (2) a \emph{stopping rule} $\tau_\delta$, a stopping time for the filtration $(\cF_t)_{t\geq 1}$, and (3) a \emph{decision rule} $\hi\in \cI$ which is $\cF_{\tau_\delta}$-measurable.
Non-deterministic algorithms could also be considered by allowing the rules to depend on additional internal randomization. The algorithms we present are deterministic.

\paragraph{$\delta$-correctness and fixed-confidence objective.}
An algorithm is $\delta$-correct if it predicts the correct answer with probability at least $1-\delta$, precisely if $\P_\theta \big(\hi \neq \istar(\theta)\big) \leq \delta$ and $\tau_\delta < +\infty$ almost surely for all $\theta \in\cM$. Our goal is to find a $\delta$-correct algorithm that minimizes the \emph{sample complexity}, that is,  $\E_\theta[\tau_\delta]$ the expected number of sample needed to predict an answer.

Pure exploration (in particular BAI) for linear bandits has been previously studied by
~\citet{soare2014linear,tao2018alba,xu2018linear,zaki2019maxoverlap,fiez2019transductive,kazerouni2019glb}. They all consider the fixed-confidence setting. To the best of our knowledge, only~\citet{hoffman2014bayesgap} study the problem with a fixed-budget.

Beside studying fixed-confidence sample complexity,~\citet{garivier2016tracknstop} and some subsequent works~\citep{qin2017ttei,shang2019t3c} investigate a general criterion of judging the optimality of a BAI sampling rule: Algorithms that achieve the minimal sample complexity when~$\delta$ tends to zero are called asymptotically optimal. \citet{menard2019lma} and~\citet{degenne2019game} further study the problem in a game theoretical point of view, and extend the asymptotic optimality to the general pure exploration for structured bandits. Note that a naive adaptation of the algorithm proposed by~\citet{degenne2019game} may not work smoothly in our setting. In this paper we use some different confidence intervals that benefit better from the linear structure.

\paragraph{Contributions.}
\textbf{1)}
We provide new insights on the complexity of linear pure exploration bandits. In particular, we relate the asymptotic complexity of the BAI problem and other measures of complexity inspired by optimal design theory, which were used in prior work.
\textbf{2)}
We develop a saddle-point approach to the lower bound optimization problem, which also guides the design of our algorithms. In particular we highlight a new insight on a convex formulation of that problem. It leads to an algorithm with a more direct analysis than previous lower-bound inspired methods.
\textbf{3)}
We obtain two algorithms for linear pure exploration bandits in the fixed-confidence regime. Their sample complexity is asymptotically optimal and their empirical performance is competitive with the best existing algorithms.

%!TEX root = ../lin_bandit_explo.tex
\section{Asymptotic Optimality}\label{sec:lower_bound}

In this section we extend the lower bound of \citet{garivier2016tracknstop}, to hold for \emph{pure exploration in finite-armed linear bandit} problems.

\paragraph{Alternative.}
For any answer $i\in\cI$ we define \emph{the alternative to i}, denoted by $\neg i$ the set of parameters where the answer $i$ is not correct, i.e.
$%\[
\neg i \eqdef \{\theta\in\cM:\ i\neq\istar(\theta)\}\,.
$%\]

We also define, for any $w\in(\R^+)^A$, the design matrix
\[V_w \eqdef \sum_{a\in \cA} w^a a a^\top\,.\]
Further, we define $\normm{x}_V\eqdef \sqrt{x^\top Vx}$ for $x\in\R^d$ and a symmetric positive matrix $V\in\R^{d\times d}$. Note that it is a norm only if $V$ is positive definite. We also denote by $\Sigma_K$ the probability simplex of dimension $K-1$ for all $K\ge 2$.

\paragraph{Lower bound.}
We have the following non-asymptotic lower bound, proved in Appendix~\ref{app:lower_bound}, on the sample complexity of any $\delta$-correct algorithm. This bound was already proved by \citet{soare2014linear} for the BAI example.
\begin{theorem}
\label{th:lb_genral}
For all $\delta$-correct algorithms, for all $\theta\in \cM$,
\begin{equation*}
  \label{eq:lb_general}
  \liminf_{\delta\to 0}\frac{\E_\theta[\tau_{\delta}]}{\log(1/\delta)} \geq \Tstar(\theta) \,,
\end{equation*}
where the \emph{characteristic time} $\Tstar(\theta)$ is defined by
\[
\Tstar(\theta)^{-1} \eqdef \max_{w \in \Sigma_A} \inf_{\lambda\in \neg \istar(\theta)} \frac{1}{2}\normm{\theta - \lambda}_{V_w}^2\,.
\]
\end{theorem}
In particular, we say that a $\delta$-correct algorithm is asymptotically optimal if for all $\theta\in\cM,$
\[
\limsup_{\delta\to 0}\frac{\E_\theta[\tau_{\delta}]}{\log(1/\delta)} \leq \Tstar(\theta)\,.
\]
As noted in the seminal work of \citet{chernoff1959}, the complexity $\Tstar(\theta)^{-1}$ is the value of a fictitious zero-sum game between the agent choosing an optimal proportion of allocation of pulls $w$ and a second player, the nature, that tries to fool the agent by choosing the most confusing alternative $\lambda$ leading to an incorrect answer.
%In fact using the Sion's minimax theorem we can rewrite the characteristic times in various ways which will be useful to prove the lower bound of Theorem~\ref{th:lb_genral}.

\textbf{Minimax theorems.} Using Sion's minimax theorem we can invert the order of the players if we allow nature to play mixed strategies
\begin{align}
\Tstar(\theta)^{-1} &= \max_{w \in \Sigma_A} \inf_{\lambda\in \neg \istar(\theta)} \frac{1}{2} \normm{\theta - \lambda}_{V_w}^2 \label{eq:sion} \\
%&= \max_{w \in \Sigma_A} \inf_{q\in \cP(\neg \istar(\theta))} \frac{1}{2} \E_{\lambda\sim q}\normm{\theta - \lambda}_{V_w}^2\nonumber\\
&= \inf_{q\in \cP(\neg \istar(\theta))} \max_{a\in\cA}\frac{1}{2}\E_{\lambda\sim q}\normm{\theta - \lambda}_{aa^\top}^2\nonumber\,,
\end{align}
where $\cP(\cX)$ denotes the set of probability distributions over the set $\cX$. The annoying part in this formulation of the characteristic time is that the set $\neg \istar(\theta)$ where the nature plays is a priori unknown (as the parameter is unknown to the agent). Indeed, to find an asymptotically optimal algorithm one should somehow solve this minimax game. But it is easy to remove this dependency noting that $\inf_{\lambda\in\neg i}\normm{\theta -\lambda}=0$ for all $i\neq \istar(\theta)$,
\[
\Tstar(\theta)^{-1} = \max_{i\in\cI}\max_{w \in \Sigma_A} \inf_{\lambda\in \neg i } \frac{1}{2}\normm{\theta - \lambda}_{V_w}^2\,.
\]
Now we can see the characteristic time $\Tstar(\theta)^{-1}$ as the value of an other game where the agent plays a proportion of allocation of pulls $w$ \emph{and} an answer $i$. The agent could also use mixed strategies for the answer which leads to
\begin{align*}
\Tstar(\theta)^{-1} &= \max_{\rho\in\Sigma_I}\max_{w \in \Sigma_A}  \frac{1}{2}\sum_{i\in \cI }\inf_{\lambda^i\in\neg i}\rho_i\normm{\theta - \lambda^i}_{V_w}^2\\
& =\max_{\rho\in\Sigma_I}\max_{w \in \Sigma_A}\inf_{\tlambda\in\prod_i(\neg i)}  \frac{1}{2}\sum_{i\in \cI }\rho_i\normm{\theta - \tlambda^i}_{V_w}^2\,,
\end{align*}
where $\prod_{i\in\cI}(\neg i)$ denotes the Cartesian product of the alternative sets $\neg i$. But the function that appears in the value of the new game is not anymore convex in $(w,\rho)$ and Sion's minimax theorem does not apply anymore. We can however convexify the problem by letting the agent to play a distribution $\tw\in\Sigma_{AI}$ over the arm-answer pairs $(a,i)$, see Lemma~\ref{lem:sion_convexify} below proved in Appendix~\ref{app:lower_bound}.
\begin{lemma}
\label{lem:sion_convexify} For all $\theta\in\cM$,
\begin{align*}
  \Tstar(\theta)^{-1} &= \max_{\tw \in \Sigma_{AI}} \inf_{\tlambda\in \prod_i (\neg i) }\frac{1}{2} \sum_{(a,i)\in\cA\times\cI}\tw^{a,i}\normm{\theta - \tlambda^i}_{aa^\top}^2\\
   %&= \max_{\tw \in \Sigma_{AI}}\inf_{\tq\in \prod_i\cP(\neg i) }\!\sum_{(a,i)\in\cA\times\cI}\!\!\!\tw^{a,i}\E_{\lambda^i\sim \tq^i}\normm{\theta - \lambda^i}_{aa^\top}^2\\
   &= \inf_{\tq\in \prod_i\cP(\neg i) } \frac{1}{2}\max_{(a,i)\in\cA\times\cB}\E_{\tlambda^i\sim \tq^i}\normm{\theta - \tlambda^i}_{aa^\top}^2\,.
\end{align*}
\end{lemma}
Thus in this formulation the characteristic time is the value of a fictitious zero-sum game where the agent plays a distribution $\tw\in\Sigma_{AI}$ over the arm-answer pairs $(a,i)\in\cA\times\cI$ and nature chooses an alternative $\tlambda^i\in\neg i$ for all the answers $i\in\cI$. The algorithm \algoWCvx that we propose in Section~\ref{sec:algorithm} is based on this formulation of the characteristic time whereas algorithm \algoW is based on the formulation of Theorem~\ref{th:lb_genral}.

\textbf{Best-arm identification complexity.}
The inverse of the characteristic time of Theorem~\ref{th:lb_genral} specializes to
\[
\Tstar(\theta)^{-1} = \max_{w\in\Sigma_A} \min_{a\neq \astar(\theta)} \frac{\big\langle \theta, \astar(\theta)-a\big\rangle^2}{2 \normm{\astar(\theta)-a}_{V_w^{-1}}^2}
\]
for BAI (see Appendix~\ref{app:bai} for a proof). It is also possible to explicit the characteristic time
\[
\Tstar(\theta) = \min_{w\in\Sigma_A} \max_{a\neq \astar(\theta)} \frac{2\normm{\astar(\theta)-a}_{V_w^{-1}}^2}{\big\langle \theta, \astar(\theta)-a\big\rangle^2}\,.
\]
Since the characteristic time involves many problem dependent quantities that are unknown to the agent, previous papers target loose problem-independent upper bounds on the characteristic time. \citet{soare2014linear} (see also \citealt{tao2018alba}, \citealt{fiez2019transductive}) introduce the G-complexity (denoted by $\gopt$) which coincides with the G-optimal design of experimental design theory (see \citealt{pukelsheim2006optimal}) and the $\xyopt$-complexity\footnote{This complexity is denoted as $\cX\cY$ by \citet{soare2014linear}.} (denoted by $\xyopt$) inspired by the transductive experimental design theory  \citep{yu2006active},
\begin{align*}
\gopt &=\min_{w\in\Sigma_A} \max_{a\in\cA} \normm{a}_{V_w^{-1}}^2\,,\\
\xyopt &=\min_{w\in\Sigma_A} \max_{b\in\cB_{\texttt{dir}}} \normm{b}_{V_w^{-1}}^2\,,
\end{align*}
where $\cB_{\texttt{dir}}\eqdef\{a-a':\ (a,a')\in\cA\times\cA\}$. For the G-optimal complexity we seek for a proportion of pulls $w$ that explores \emph{uniformly} the means of the arms, since the statistical uncertainty for estimating $\langle \theta,a\rangle$ scales roughly with $\normm{a}_{V_w^{-1}}$. In the $\cA\cB$-complexity we try to estimate \emph{uniformly} all the \emph{directions} $a-a'$. On the contrary in this paper we try to maximize directly the characteristic times, that is try to estimate all the \emph{directions} $\astar(\theta) - a$ scaled by the squared gaps $\langle\theta,\astar(\theta)-a\rangle$.
Note that the characteristic time can also be seen as a particular optimal transductive design. Indeed for $\cBstar \eqdef \left\{ (\astar(\theta)- a)/\left|\big\langle \theta, \astar(\theta)-a\big\rangle\right|: a\in\cA/\big\{\astar(\theta)\big\}  \right\}$, it holds
\[
\Tstar(\theta) = 2 \cA\cBstar(\theta) \eqdef 2 \min_{w\in\Sigma_A} \max_{b\in\cBstar(\theta)} \normm{b}_{V_w^{-1}}^2\,.
\]
We have the following ordering on the complexities
\begin{align}\label{eq:complexities}
\Tstar(\theta) \leq 2 \frac{\xyopt}{\DeltaMin(\theta)^2}\leq 8 \frac{\gopt}{\DeltaMin(\theta)^2} = \frac{8d}{\DeltaMin(\theta)^2}\,\CommaBin
\end{align}
where $\DeltaMin = \min_{a\neq\astar(\theta)}\langle\theta,\astar(\theta)-a\rangle$ and  the last equality follows from the Kiefer-Wolfowitz equivalence theorem~\citep{kiefer1959}. Conversely the $\gopt$-complexity and the $\xyopt$-complexity are linked to an \emph{other} pure exploration problem, the thresholding bandits (see Appendix~\ref{app:threshold_bandits}).

\begin{remark}
In order to compute all these complexities, it is sufficient to solve the following generic optimal transductive design problem: for $\cB$ a finite set of elements in $\R^d$,
\[
\cA\cB=\min_{w\in\Sigma_K}\max_{b\in\cB}\normm{b}^2_{V_w^{-1}}\,.
\]
When $\cB=\cA$ we can use an algorithm inspired by Frank-Wolfe \citep{frank1956algorithm} which possesses convergence guarantees~\citep{atwood1969optimal,ahipasaoglu2008fw}. But in the general case, up to our knowledge, there is no algorithm with the same kind of guarantees. Previous works used an heuristic based on a straightforward adaptation of the aforementioned algorithm for general sets $\cB$ but it seems to not converge on particular instances, see Appendix~\ref{app:implem}. We instead propose in the same appendix an algorithm based on Saddle point Frank-Wolfe algorithm that seems to converge on the different instances we tested.
\end{remark}

%!TEX root = ../lin_bandit_explo.tex
\section{Algorithm}\label{sec:algorithm}

We present two asymptotically optimal algorithms for the general pure-exploration problem. We also make the additional assumption that the set of parameter is bounded, that is we know $M>0$ such that for all $\theta\in\cM,\, \normm{\theta}\leq M$. This assumption is shared by most of the works on %\todo{rather usual? or used everywhere? Evrywhere i would say since they target minimax regret}
linear bandits (e.g. \citealt{abbasi-yadkori2011linear, soare2014linear}).

We describe primarily \algoWCvx, detailed in Algorithm~\ref{alg:algoWCvx}. The principle behind \algoW, detailed in Algorithm~\ref{alg:algoW}, is similar and significant differences will be highlighted.

\subsection{Notations}
\paragraph{Counts.} At each round $t$ the algorithms will play an arm $a_t$ and choose (fictitiously) an answer $i_t$. We denote by $N_t^{a,i} \eqdef\sum_{s=1}^t \ind_{\{(a_t,i_t)=(a,i)\}}$ the number of times the pair $(a,i)$ is chosen up to and including time $t$, and by $N_t^a =\sum_{i\in\cI} N_s^{a,i}$ and $N_t^i =\sum_{a\in\cI} N_s^{a,i}$ the partial sums. The vectors of counts at time $t$ is denoted by $N_t \eqdef (N_t^a)_{a\in\cA}$
and when it is clear from the context we will also denote by $N_t^a = (N_t^{a,i})_{i\in\cI}$ and $N_t^i = (N_t^{a,i})_{i\in\cI}$ the vectors of partial counts.

\paragraph{Regularized least square estimator.} We fix a regularization parameter $\eta > 0$. The regularized least square estimator for the parameter $\theta\in \mathcal M$ at time $t$ is
\[
\htheta_{t} = (V_{N_t} + \eta I_d)^{-1} \sum_{s=1}^t Y_s a_s\,,
\]
where $I_d$ is the identity matrix. By convention $\htheta_0 = 0$.

\subsection{Algorithms}
\begin{algorithm}[tb]
   \caption{\algoW}
   \label{alg:algoW}
\begin{algorithmic}
   \STATE {\bfseries Input:} Agent learners for each answers $(\cL^i_w)_{i\in\cI}$, threshold $\beta(\cdot,\delta)$
   \FOR{t = 1 \ldots}
    \STATE \textit{// Stopping rule}
    \IF{{\small$\max_{i\in\cI}\inf_{\lambda\in\neg i} \frac{1}{2}\normm{\htheta_{t-1}-\lambda}^2_{V_{N_{t-1}}}\geq\! \beta(t-1,\delta)$}}
      \STATE {\bfseries stop} and {\bfseries return} $\hi = \istar(\hat{\theta}_{t-1})$
    \ENDIF
    \STATE \textit{// Best answer}
    \STATE $i_{t} = \istar(\hat{\theta}_{t-1})$
    \STATE \textit{// Agent plays first}
    \STATE Get $w_{t}$ from $\cL^{i_t}_w$ and update $W_{t}=W_{t-1}+w_{t}$
    \STATE \textit{// Best response for the nature}
    \STATE $\lambda_{t} \in \argmin_{\lambda\in\neg i_t}\normm{\htheta_{t-1}-\lambda}^2_{V_{w_{t}}}$
    \STATE \textit{// Feed optimistic gains}
    \STATE Feed learner $\cL_w^{i_{t}}$ with $g_{t}(w) = \sum_{a\in\cA}w^a U_t^a/2$
    \STATE \textit{// Track the weights}
    \STATE Pull $a_{t}\in \argmin_{a\in \mathcal A } N_{t-1}^{a} - W_{t}^{a}$
   \ENDFOR
\end{algorithmic}
\end{algorithm}

\begin{algorithm}[tb]
   \caption{\algoWCvx}
   \label{alg:algoWCvx}
\begin{algorithmic}
   \STATE {\bfseries Input:} Agent learner $\cL_{\tw}$, threshold $\beta(\cdot,\delta)$
   \FOR{t = 1 \ldots}
    \STATE \textit{// Stopping rule}
    \IF{ {\small $\max_{i\in \cI} \inf_{\lambda\in\neg i} \frac{1}{2}\normm{\htheta_{t-1}-\lambda}^2_{V_{N_{t-1}}}\geq \beta(t-1,\delta)$}}
      \STATE {\bfseries stop} and {\bfseries return} $\hi = \istar(\hat{\theta}_{t-1})$.
    \ENDIF
    \STATE \textit{// Agent plays first}
    \STATE Get $\tw_{t}$ from $\cL_{\tw}$ and update $\tW_{t}=\tW_{t-1}+\tw_{t}$
    \STATE \textit{// Best response for the nature}
    \STATE For all $i\in\cI$, $\tlambda^i_{t} \in \argmin_{\lambda\in\neg i}\normm{\htheta_{t-1}-\lambda}^2_{V_{\tw^i_{t}}}$
    \STATE \textit{// Feed optimistic gains}
    \STATE Feed learner $\cL_{\tw}$ with {\small $g_{t}(\tw) =\sum_{(a,i)\in\cA\times\cI} \tw^{a,i} U_t^{a,i}/2$}
    \STATE \textit{// Track the weights}
    \STATE Pull $(a_{t},i_{t})\in \argmin_{(a,i)\in \mathcal A \times \mathcal I} N_{t-1}^{a,i} - \tW_{t}^{a,i}$
   \ENDFOR
\end{algorithmic}
\end{algorithm}
\vspace{-0.4cm}
\paragraph{Stopping rule.}
Our algorithms share the same stopping rule. Following \citet{garivier2016tracknstop}, our algorithms stop if a generalized likelihood ratio exceeds a threshold. It stops if
\begin{align}
\label{eq:def_chernoff_stopping}
\max_{i\in \mathcal I} \inf_{\lambda_i \in \neg i}\frac{1}{2}\Vert \htheta_t - \lambda_i \Vert^2_{V_{N_t}}
> \beta(t,\delta)\,,
\end{align}
and return {\small $\displaystyle \istar_t\! \in\! \argmax_{i\in \mathcal I}\! \inf_{\lambda_i \in \neg i}\Vert \htheta_t - \lambda_i \Vert^2_{V_{N_t}}/2$}.
This stopping and decision rules ensures that the algorithms \algoW and \algoWCvx are $\delta$-correct regardless of the sampling rule used, see lemma below\footnote{The fact that $\tau_\delta <+\infty$ is a consequence of our analysis, see Appendix~\ref{app:proof}.} proved in Appendix~\ref{app:concentration}.
\begin{lemma}
\label{lem:chernoff_stopping rule_pac}
Regardless of the sampling rule, the stopping rule~\eqref{eq:def_chernoff_stopping} with the threshold
{\small\begin{equation} \label{eq:def_beta}
\beta(t,\delta) =\left( \sqrt{\log\!\left( \frac{1}{\delta}\right)+\frac{d}{2}\log\!\left(1+\frac{t L^2}{\eta d} \right)} +\sqrt{\frac{\eta}{2}}M\right)^2\!\!\!,
\end{equation}}
satisfy $ \P_{\theta}\big(\tau_{\delta} < \infty \wedge \istar_{\tau_\delta} \neq \istar(\theta)\big) \leq \delta$.
\end{lemma}
% This stopping and decision rules ensures that the algorithm is $\delta$-correct regardless of the sampling rule used (see \citealt{garivier2016tracknstop} for a proof), hence the following theorem is immediate.
% \begin{theorem}
% Algorithms~\ref{alg:algoW} and~\ref{alg:algoWCvx} are $\delta$-correct.
% \end{theorem}
Our contribution is a sampling rule that minimizes the sample complexity when combined with these stopping and decision rules.
We now explain our sampling strategy to ensure that the stopping threshold is reached as soon as possible.

\paragraph{Saddle point computation.}
Suppose in this paragraph, for simplicity, that the parameter vector $\theta$ is known. By the definition of the stopping rule and the generalized likelihood ratio, as long as the algorithm does not stop,
\begin{align*}
\beta(t,\delta)
&\ge \inf_{\lambda\in \neg i^\star(\theta)} \sum_{a\in \cA} N_t^a \Vert \theta - \lambda \Vert^2_{a a^\top}/2 \: .
\end{align*}
If we manage to have $N_t \approx t w^\star(\theta)$ (the optimal pulling proportions at $\theta$), then this leads to $\beta(t,\delta) \ge t T^\star(\theta)^{-1}$ and, solving that equation, we have asymptotic optimality.

Since there is only one correct answer, the parameter $\theta$ belongs to all sets $\neg i$ for $i\neq \istar(\theta)$. Hence
\begin{align*}
&\inf_{\lambda\in \neg i^\star(\theta)}\frac{1}{2} \sum_{a\in \cA} N_t^a \Vert \theta - \lambda \Vert^2_{a a^\top}
\\&\geq \inf_{\tlambda_t\in \prod_i (\neg i)}\frac{1}{2}\sum_{(a,i)\in \cA\times\cI}\!\!\!\!\! N_t^{a,i} \Vert \theta - \tlambda^i \Vert^2_{a a^\top}.
\end{align*} Introducing the sum removes the dependence in the unknown $i^*(\theta)$. \algoWCvx then uses an agent playing weights w in $\Sigma_{\cA\cI}$. \algoW does not use that sum over answers, but uses a guess for $\istar(\theta)$. Its analysis involves proving that the guess is wrong only finitely many times in expectation.

Our sampling rule implements the lower bound game between an agent, playing at each stage $s$ a weight vector $\tw_s$ in the probability simplex $\Sigma_{A\times I}$, and nature, who computes at each stage a point $\lambda_s^i \in \neg i$ for all $i\in \mathcal I$. We additionally ensure that $N_t^{a,i} \approx \sum_{s=1}^t \tw_s^{a,i}$. Suppose that the sampling rule is such that at stage $t$, a $\varepsilon_t$-approximate saddle point is reached for the lower bound game, see Lemma~\ref{lem:sion_convexify}. That is,
\begin{align*}
&\inf_{\tlambda \in \prod_{i}(\neg i)} \sum_{s=1}^t \sum_{(a,i)\in \cA\times\cI} \tw_s^{i,a} \Vert \theta - \tlambda^i \Vert^2_{a a^\top}/2 +\varepsilon_t
\\
&\ge \sum_{s=1}^t \sum_{(a,i)\in \cA\times\cI} \tw_s^{i,a} \Vert \theta - \tlambda_s^i \Vert^2_{a a^\top}/2
\\
&\ge\max_{(a,i)\in \cA\times\cI} \sum_{s=1}^t \Vert \theta - \tlambda_s^i \Vert^2_{a a^\top}/2 - \varepsilon_t \: .
\end{align*}
Then if the algorithm did not stop, it verifies, using Lemma~\ref{lem:sion_convexify},
\begin{align*}
\beta(t,\delta)
&\ge t \max_{(a,i)\in \cA\times\cI} \frac{1}{t}\sum_{s=1}^t \Vert \theta - \tlambda_s^i \Vert^2_{a a^\top}/2 - 2\varepsilon_t
\\
&\ge t \inf_{\tq \in \mathcal \prod_{i\in \mathcal I}\cP(\neg i)} \! \max_{(a,i)\in \cA\times\cI} \!\!\!\! \mathbb{E}_{\lambda^i\sim q^i}\Vert \theta \! - \! \tlambda^i \Vert^2_{a a^\top}/2 \! - \! 2\varepsilon_t\\
&= t T^\star(\theta)^{-1} - 2 \varepsilon_t \: .
\end{align*}
Solving that equation, we get asymptotically the wanted $t\lesssim T^\star(\theta) \log(1/\delta)$.

We implement the saddle point algorithm by using AdaHedge for the agent (a regret minimizing algorithm of the exponential weights family), and using best-response for the nature, which plays after the agent. Precisely the learner $\cL_w$ for \algoWCvx is AdaHedge on $\Sigma_{AI}$ with the gains
\[
g_t^\theta(\tw) = \frac{1}{2} \sum_{(a,i)\in\cA\times\cI}  \tw^{a,i} \Vert \theta - \tlambda_s^i \Vert^2_{a a^\top}\,.
\]
 Whereas \algoW uses $I$ learners $\cL_w^i$, one for each possible guess of $\istar(\theta)$ with the gains. For $i\in\cI$, the learner $\cL_w^i$ is also AdaHedge but only on $\Sigma_A$ with the gains (when the guess is $i$)
 \[
 g_t^\theta(w) = \frac{1}{2} \sum_{a\in\cA}  w^{a} \Vert \theta - \lambda_s^i \Vert^2_{a a^\top}\,.
 \]
 $\varepsilon_t$ is then the sum of the regrets of the two players. Best-response has regret 0, while the regret of AdaHedge is $O(\sqrt{t})$ for bounded gains, as seen in the following lemma, taken from \citet{derooij2014hedge}.
%\todo{define the learners $\cL_w^i$ here}
\begin{lemma}\label{lem:adahedge}
On the online learning problem with $K$ arms and gains $g_s(w) = \sum_{k\in[K]} w^k  U_s^k$ for $s\in[t]$, AdaHedge, predicting $(w_s)_{s\in[t]}$, has regret
\begin{align*}
%R_T &\le 2\sqrt{(\sigma L_T - \frac{L_T^2}{T})\log(K)} + \sigma(2+\log(K)16/3) \: ,\\
R_t&:= \max_{w\in\Sigma_K}\sum_{s=1}^t g_s(w) -g_s(w_s) \\
&\le 2\sigma\sqrt{t\log(K)} + 16\sigma(2+\log(K)/3) \: ,\\
\text{where }
\sigma &= \max_{s\le t}  (\max_{k\in[K]}U_s^{k}- \min_{k\in[K]}U_s^{k}) \:.\\
%L_T &= \sum_{t=1}^T (\max_{k\in[K]}U_t^{k} - U_t^{\star}) \le T \sigma \: .
\end{align*}
\end{lemma}
Other combinations of algorithms are possible, as long as the sum of their regrets is sufficiently small. At each stage $t\in\N$, both algorithms advance only by one iteration and as time progresses, the quality of the saddle point approximation improves. This is in contrast with \Track \cite{garivier2016tracknstop}, in which an exact saddle point is computed at each stage, at a potentially much greater computational cost.

\paragraph{Optimism.} The above saddle point argument would be correct for a known game, while our algorithm is confronted to a game depending on the unknown parameter $\theta$. Following a long tradition of stochastic bandit algorithms, we use the principle of Optimism in Face of Uncertainty. Given an estimate $\hat{\theta}_{t-1}$, we compute upper bounds for the gain of the agent at $\theta$, and feed these optimistic gains to the learner. Precisely, given the best response $\lambda_t^i \in \neg i$ we define,
\begin{align*}
U_t^{a,i} =\left\{
\begin{array}{ll}
\max_{\xi} \quad & \min\big(\Vert \xi - \lambda_t^i \Vert^2_{a a^\top},4L^2M^2\big)\\
\text{s.t.}\quad & \Vert \hat{\theta}_{t-1} - \xi \Vert^2_{V_{N_{t-1}}+\eta I_d} \le 2h(t)
\end{array}
\right. \: ,
\end{align*}
where $h(t)=\beta(t, 1/t^3)$ is some exploration function. We clipped the values, using that $\cM$ and $\cA$ are bounded to ensure bounded gains for the learners. Under the event that the true parameter verifies $\Vert \hat{\theta}_{t-1} - \theta \Vert^2 \le 2 h(t)$, this is indeed an optimistic estimate of $\Vert \theta - \lambda_t^i \Vert^2_{a a^\top}$. Note that $U_t^{a,i}$ has a closed form expression, see Appendix~\ref{app:proof}. The optimistic gain is then, for \algoWCvx (see Algorithm~\ref{alg:algoW} for the one of \algoW),
\[
g_t(\tw) = \frac{1}{2} \sum_{(a,i)\in\cA\times\cI}  \tw^{a,i} U_t^{a,i}\,.
\]

\paragraph{Tracking.} In both Algorithm~\ref{alg:algoW} and~\ref{alg:algoWCvx}, the agent plays weight vectors in a simplex. Since the bandit procedure allows only to pull one arm at each stage, our algorithm needs a procedure to transcribe weights into pulls. This is what we call tracking, following \citet{garivier2016tracknstop}. The choice of arm (or arm and answer) is
\begin{align*}
a_{t+1}          &\in \argmin_{a\in \mathcal A } N_{t}^{a} - W_{t+1}^{a} & \text{ for Algorithm~\ref{alg:algoW},}\\
(a_{t+1},i_{t+1})&\in \argmin_{(a,i)\in \mathcal A \times \mathcal I } N_{t}^{a,i} - \tW_{t+1}^{a,i} & \text{ for Algorithm~\ref{alg:algoWCvx}.}
\end{align*}
This procedure guarantees that for all $t\in\N, u \in \mathcal U$, with $\mathcal U = \mathcal A$ (resp. $\mathcal U =\cI\times\cA$) for Algo.~\ref{alg:algoW} (resp. Algo.~\ref{alg:algoWCvx}), $- \log (|\mathcal U|) \le N_t^{u} - W_t^{u} \le 1$. That result is due to~\citet{anon2020structure} and its proof is reproduced in Appendix~\ref{app:tracking}.

\begin{theorem}\label{thm:sample_complexity}
For a regularization parameter\footnote{This condition is a simple technical trick to simplify the analysis. An $\eta$ independent of $A$,$L$,$M$ will lead to the same results up to minor adaptations of the proof.} $\eta \geq 2(1+\log(A))AL^2+M^2$, for the threshold $\beta(t,\delta)$ given by~\eqref{eq:def_beta}, for an exploration function $h(t)=\beta(t,1/t^3)$, \algoW and \algoWCvx are $\delta$-correct and asymptotically optimal. That is, they verify for all $\theta\in \cM$,
\begin{align*}
\limsup_{\delta\to 0}\frac{\mathbb{E}_\theta[\tau_\delta]}{ \log 1/\delta} \le \Tstar(\theta) \: .
\end{align*}
\end{theorem}
The main ideas used in the proof are explained above. The full proof is in appendix~\ref{app:proof} with finite $\delta$ upper bounds.

%!TEX root = ../lin_bandit_explo.tex
\subsection{Bounded parameters}

We provide a bounded version of different examples (e.g. BAI) in Appendix~\ref{app:examples} where we add the assumption that the parameter set $\cM$ is bounded. In particular we show how it affects the lower bound of Theorem~\ref{th:lb_genral}: the characteristic time $\Tstar(\theta)$ is reduced (or equivalently $\Tstar(\theta)^{-1}$ increases). This is not surprising since we add a new constraint in the optimization problem. This means that the algorithm should stop earlier. The counterpart of this improvement is that it is often difficult to compute the best response for nature. Indeed, for example, in BAI, there is an explicit expression of the best response, see Appendix~\ref{app:bai}. When the constraint $\normm{\lambda}\leq M$ is added there is no explicit expression anymore and one needs to solve an uni-dimensional optimization problem, see Lemma~\ref{lem:lagrange_alternative}. To devise an asymptotically optimal algorithm without the boundedness assumption remains an open problem.

% , i.e for $i\in\cI$ and $w\in\interior{\Sigma_A}$ in the interior of the probability simplex of dimension $A-1$, find $\lambda^i$ such that
% \[
% \lambda^i \in \argmin_{\lambda\neg i}\normm{\theta - \lambda }_{V_{w}}\,.
% \]
% For example, in BAI, there is an explicit expression of the best response. Noting that, for $\astar\in\cA$, we have
% \[
% \min_{\lambda\neg \astar}\normm{\theta - \lambda }_{V_{w}} = \min_{a\neq \astar}\min_{\langle\lambda,a-\astar\rangle>0} \normm{\theta - \lambda }_{V_{w}}\,,
% \]
% the best response is then
% \begin{align}
% \lambda^{b} &= \theta - \frac{\max(\langle\theta,\astar-b \rangle,0) }{\normm{\astar-b}_{V_w^{-1}}^2} V_w^{-1}(\astar - b) \label{eq:best_response_bai}
% \\
% \text{for }b&\in\argmin_{a\neq \astar}\normm{\theta - \lambda^{a} }_{V_{w}}\nonumber.
% \end{align}
% When the constraint $\normm{\lambda}\leq M$ is added there is no explicit expression anymore and one needs to solve an uni-dimensional optimization problem, see Lemma~\ref{lem:lagrange_alternative}.

Note that in the proof of Theorem~\ref{thm:sample_complexity} we only use two times the boundedness assumption, first in the definition of the threshold $\beta(t,\delta)$ (see Theorem~\ref{th:confidence_beta}) to handle the bias induced by the regularization. Second, since the regret of AdaHedge is proportional to the maximum of the upper confidence bounds $U_s^{i,a}$, we need to ensure that they are bounded.

% And there exists pathological pure exploration problems where even the quantity uppper bounded by $U_s^{i,a}$, namely $\normm{\theta-\lambda}_{aa^\top}^2$ where $\lambda\in\argmin_{\lambda'\in\neg i}\normm{\theta -\lambda}$ is unbounded. See for example Appendix~G.3 by \citet{menard2019lma}.

%!TEX root = ../lin_bandit_explo.tex
\section{Related Work}\label{sec:related_work}

We survey previous work on linear BAI. The major focus is put on sampling rules in this section. We stress that all the stopping rules employed in the linear BAI literature are equivalent up to the choice of their exploration rate (More discussion given in Appendix~\ref{app:stopping}). As aforementioned, existing sampling rules are either based on \SE or \UGapE. Elimination-based sampling rules usually operate in phases and progressively discard sub-optimal directions. Gap-based sampling rules always play the most informative arm that reduces the uncertainty of the gaps between the empirical best arm and the others.

\paragraph{\XYS and \XYA.} \citet{soare2014linear} first propose a static allocation design \XYS that aims at reducing the uncertainty of the gaps of all arms. More precisely, it requires to either solve the $\xyopt$-complexity or use a \emph{greedy} version that pulls the arm $\argmin_{a\in\cA} \max_{b\in\cB_{\texttt{dir}}} \normm{b}_{V_w^{-1}}^2$ at the cost of having no guarantees. An elimination-like alternative called \XYA is proposed then to overcome that issue. We say elimination-like since \XYA does not discard arms once and for all, but reset the active arm set at each phase. \XYA and \XYS are the first algorithms being linked to $\gopt$-optimality, but are not asymptotically optimal.

\vspace{-0.2cm}
\paragraph{\ALBA.} \ALBA is also an eliminations-based algorithm designed by~\citet{tao2018alba} that improves over \XYA by a factor of $d$ in the sample complexity using a tighter elimination criterion.
\vspace{-0.2cm}

\paragraph{\RAGE.} \citet{fiez2019transductive} extend \XYS and \XYA to a more general transductive bandits setting. \RAGE is also elimination-based and requires the computation of $\xyopt$-complexity at each phase.
\vspace{-0.2cm}

\paragraph{\LGapE and variants.} \LGapE~\citep{xu2018linear} is the first gap-based sampling rule for linear BAI. \LGapE is inspired by \UGapE~\citep{gabillon2012ugape}. It is, however, not clear whether \LGapE is asymptotically optimal or not. Similar to \XYS, \LGapE either requires to solve a time-consuming optimization problem at each step, or can use a greedy version that pulls arm $\argmin_{a\in\cA} \normm{a_{i_t}-a_{j_t}}_{(V_w+aa^\top)^{-1}}^2$ instead, again at the cost of losing guarantees. Here $i_t = i^\star(\hat\theta_t)$ and $a_{j_t}$ is the most ambiguous arm w.r.t. $a_{i_t}$, i.e. $\argmax_{j\neq i_t}\langle\hat\theta_t,a_{j}-a^\star(\hat\theta_t)\rangle + \normm{a^\star(\hat\theta_t) - a_{j_t}}_{V_{N_t}^{-1}} \sqrt{2\beta(t,\delta)}$. On the other hand,~\citet{zaki2019maxoverlap} propose a new algorithm based on \LUCB. With a careful examination, we note that the sampling rule of \GLUCB is equivalent to that of the greedy \LGapE using the Sherman-Morrison formula. Later, \citet{kazerouni2019glb} provide a natural extension of \LGapE to the \emph{generalized linear bandits} setting, where the rewards depend on a strictly increasing \emph{inverse link function}. \GLGapE reduces to \LGapE when the inverse link function is the identity function.

Note that all the sampling rules presented here depend on $\delta$ (except \XYS), while our sampling rules have a $\delta$-free property which is appealing for applications as argued by~\citet{jun2016atlucb}.  Also all the guarantees in the literature are of the form $C\log(\delta) + O\big(\log(1/\delta)\big)$ for a constant $C$ that is strictly larger than $\Tstar(\theta)^{-1}$.

%\subfile{main/analysis}  % this part got absorbed into the "algorihtm" section.
%!TEX root = ../lin_bandit_explo.tex
\begin{figure*}[t!]
 \centering
 \includegraphics[clip, width= 0.3\textwidth]{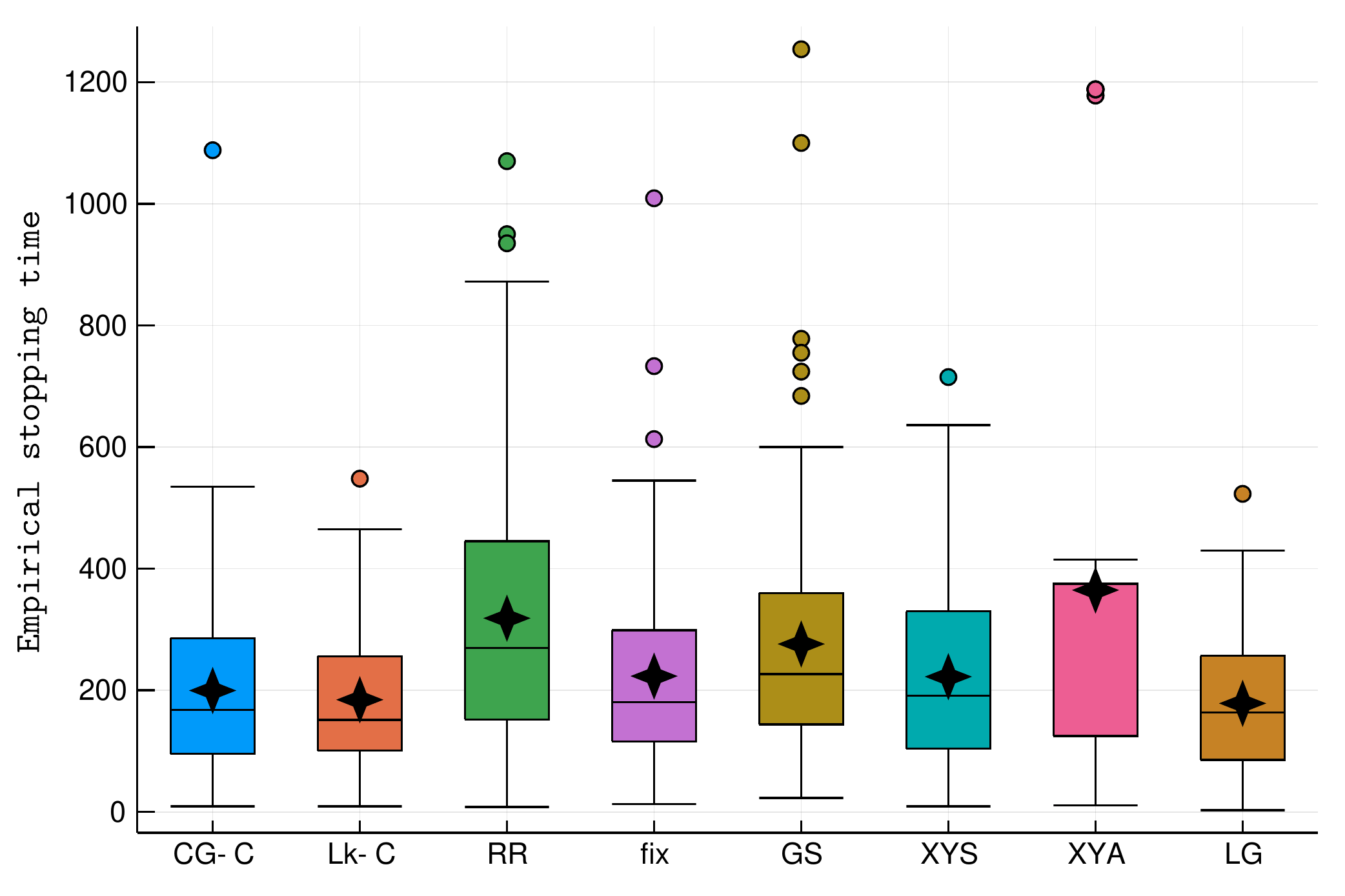}
 \includegraphics[clip, width= 0.3\textwidth]{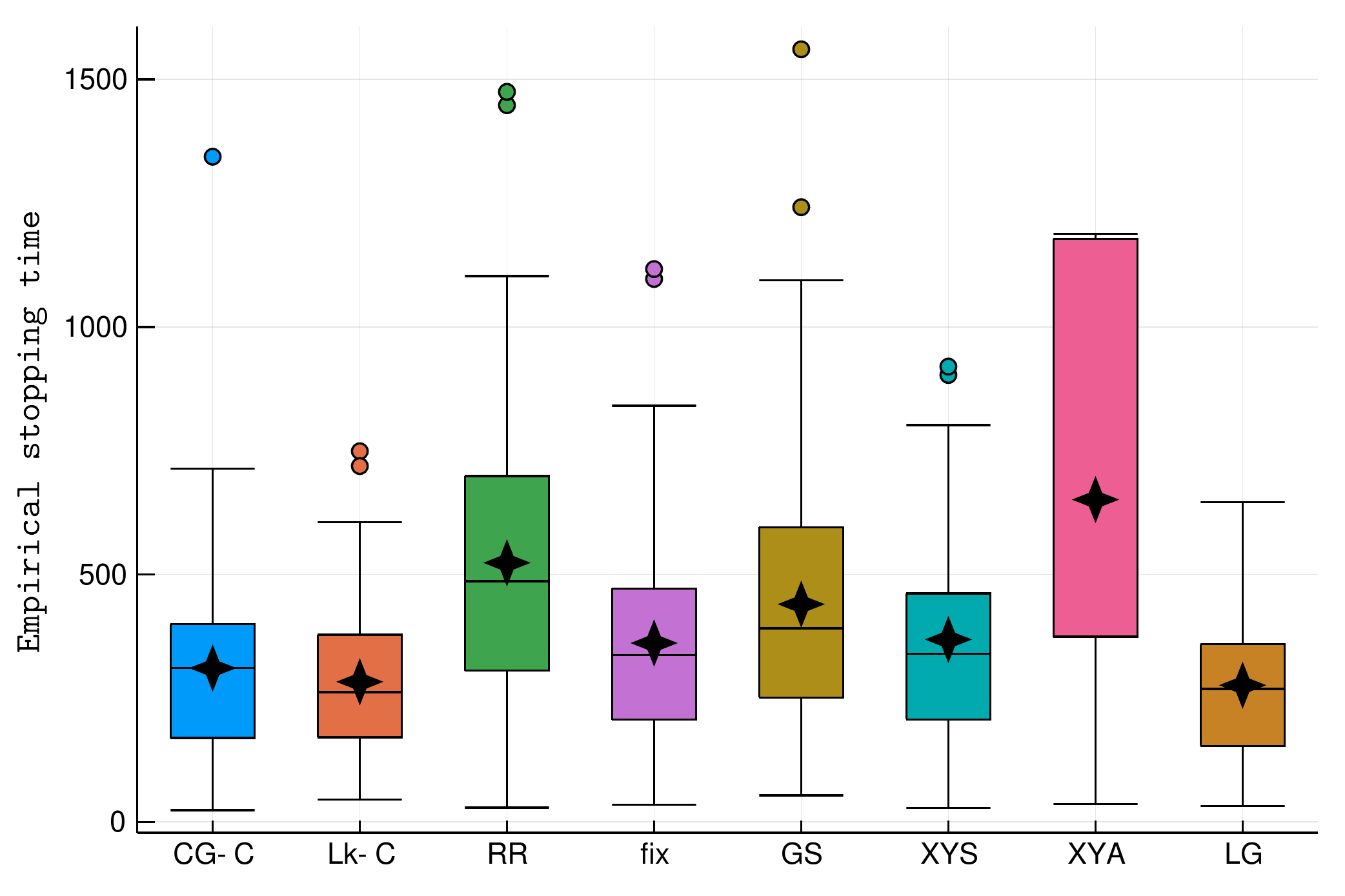}
 \includegraphics[clip, width= 0.3\textwidth]{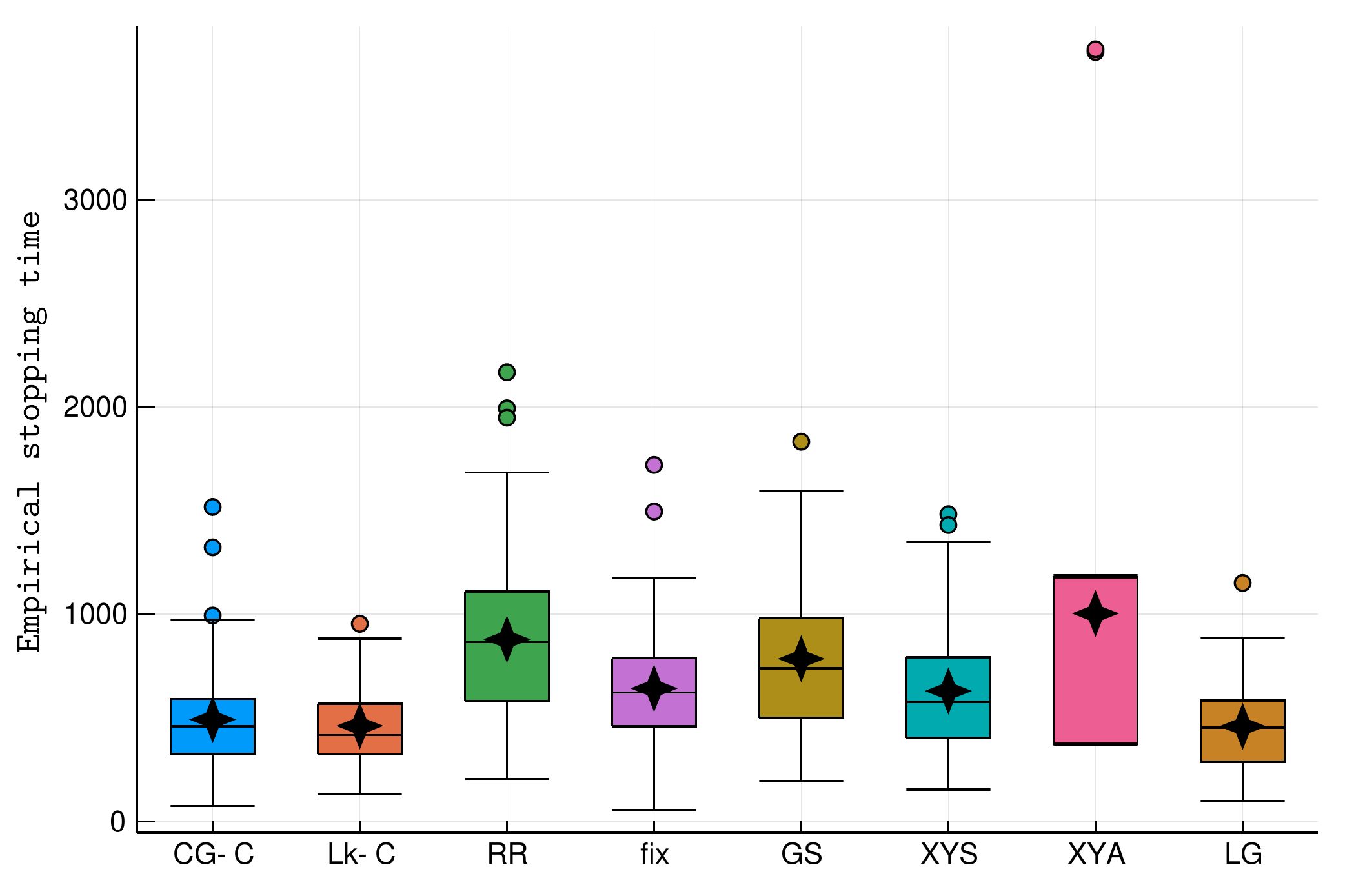}
 \caption{Sample complexity over the usual counter-example with $\delta=0.1, 0.01, 0.0001$ respectively. CG = \algoWCvx,  Lk = \algoW, RR = uniform sampling, fix = tracking the fixed weights, GS = \XYS with $\gopt$-allocation, XYS = \XYS with $\xyopt$-allocation, LG = \LGapE. The mean stopping time is represented by a black cross.}
 \label{fig:sample_complexity_1}
\end{figure*}

\begin{figure*}[t!]
 \centering
 \includegraphics[clip, width= 0.24\textwidth]{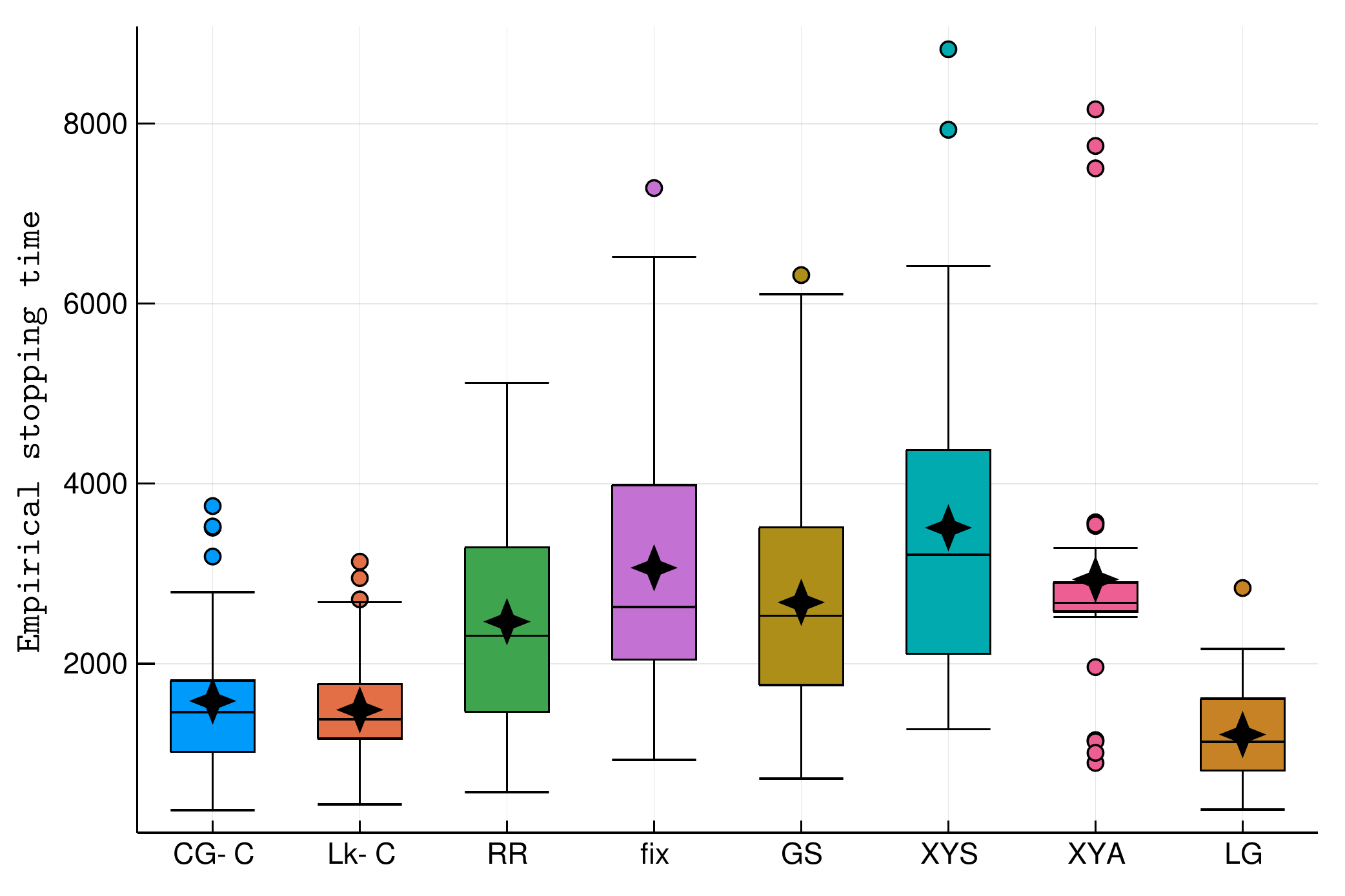}
 \includegraphics[clip, width= 0.24\textwidth]{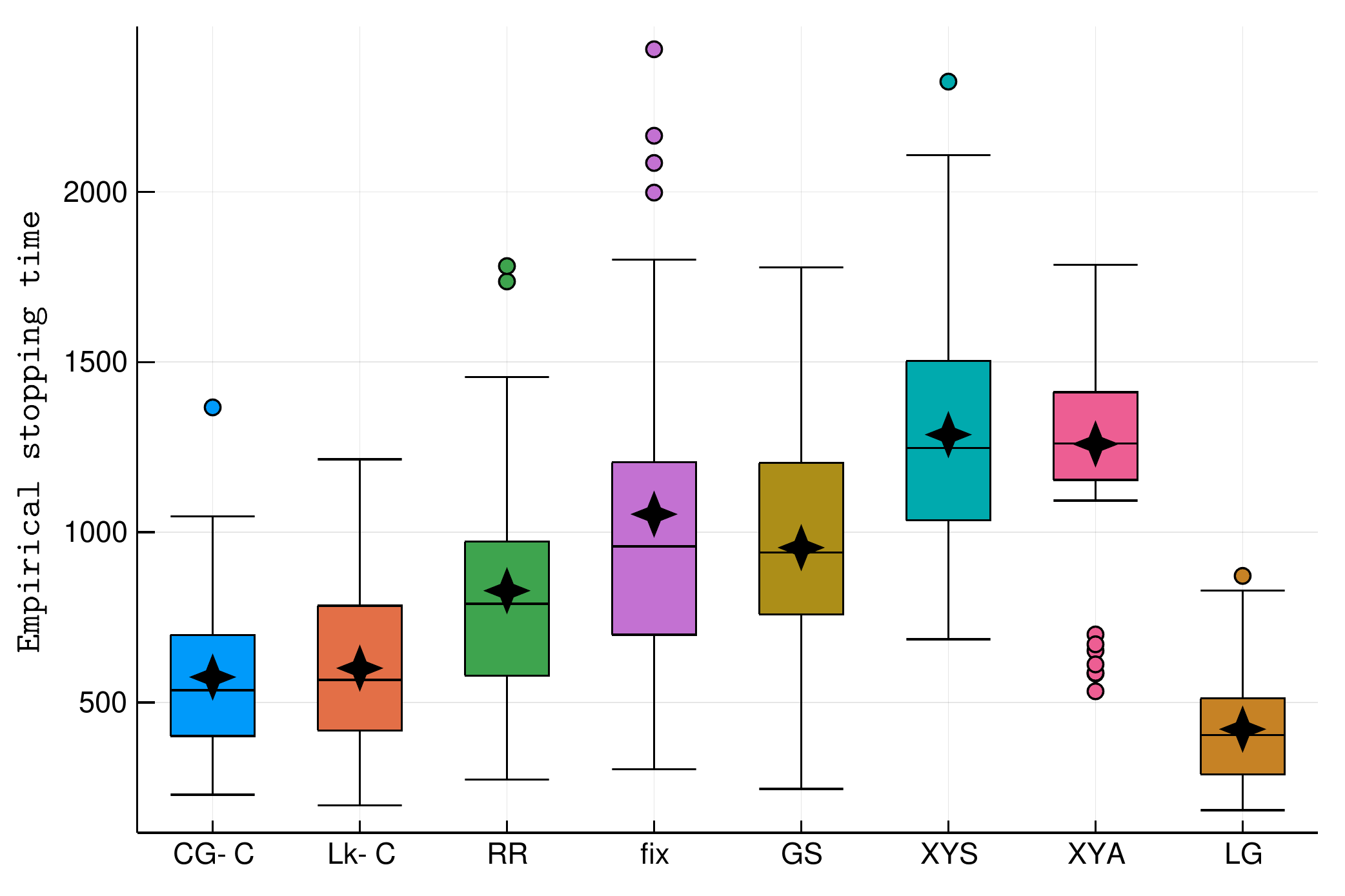}
 \includegraphics[clip, width= 0.24\textwidth]{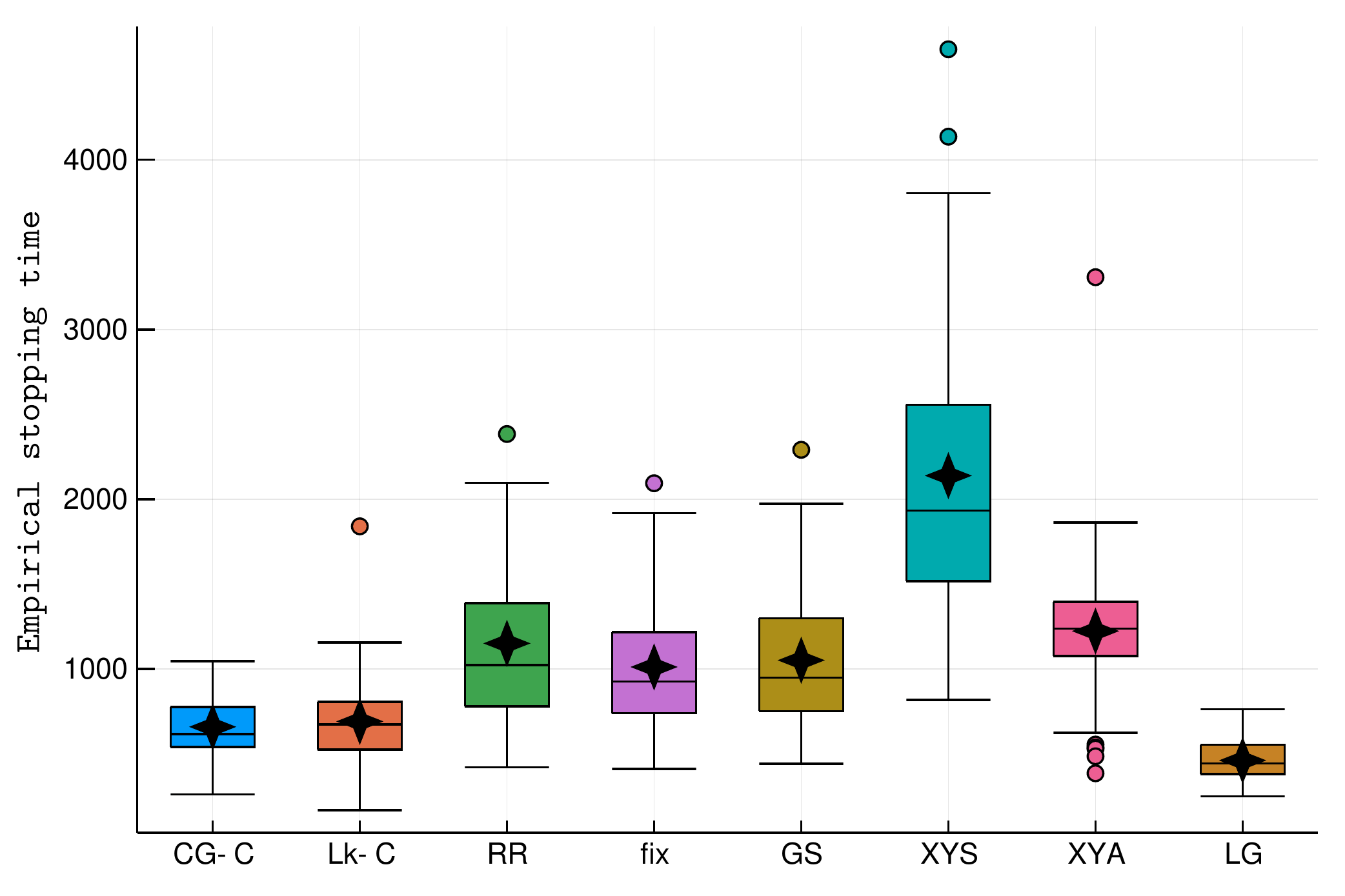}
 \includegraphics[clip, width= 0.24\textwidth]{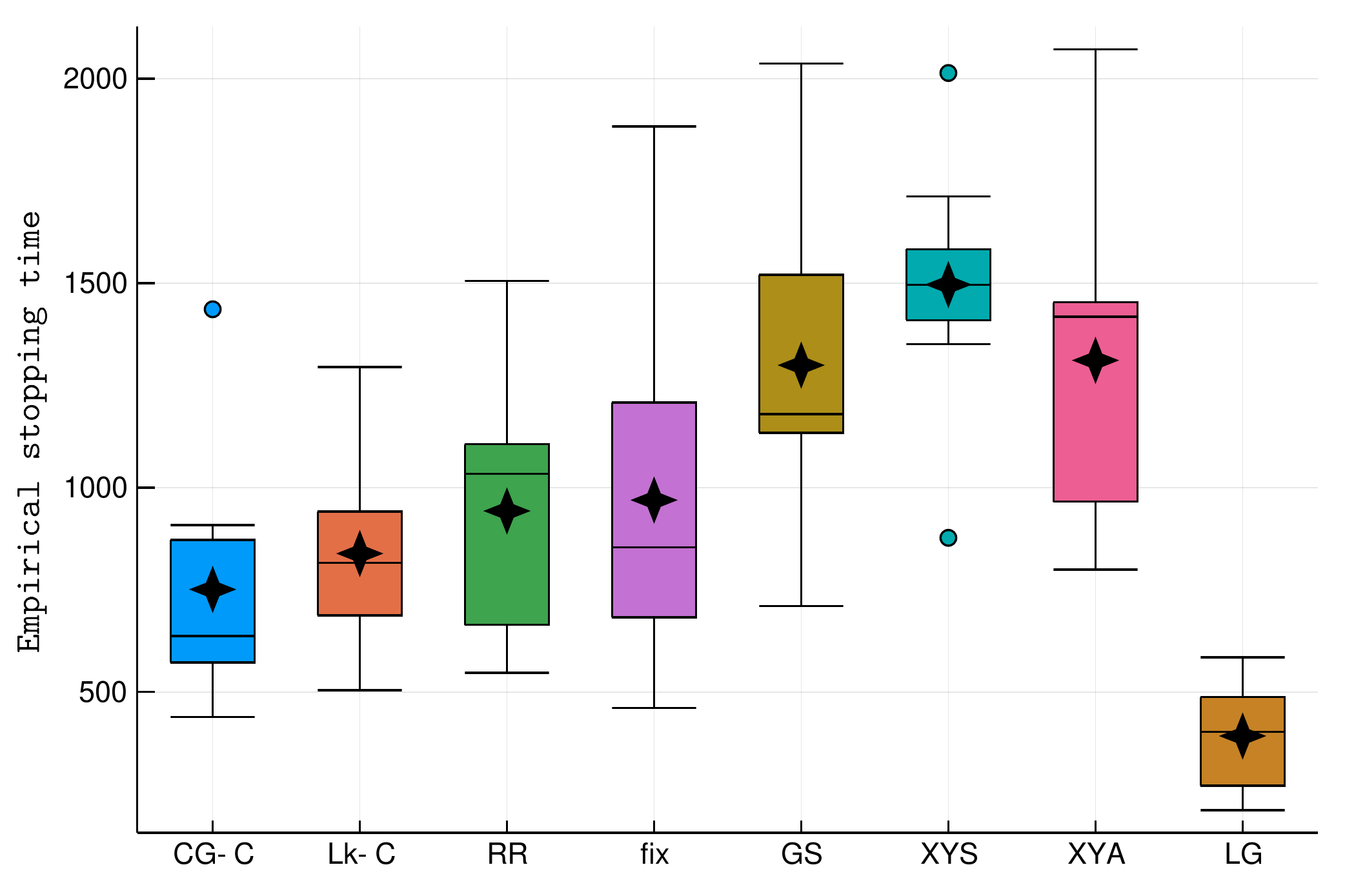}
 \caption{Sample complexity over random unit sphere vectors with $d=6, 8, 10, 12$ from left to right.}
 \label{fig:sample_complexity_2}
\end{figure*}

\vspace{-0.3cm}
\section{Experiments}\label{sec:experiments}
Besides our algorithms, we implement the following algorithms, all using the same stopping rule (more discussion given in Appendix~\ref{app:stopping}): uniform sampling, the greedy version of \XYS (including $\gopt$-allocation and $\xyopt$-allocation), \XYA, and the greedy version of \LGapE. We skip \GLUCB/\GLGapE since they are more or less equivalent to \LGapE in the scope of this paper.
\vspace{-0.3cm}
\paragraph{The usual hard instance.}
Usual sampling rules for classical BAI may not work for the linear case. This can be understood on a well-studied instance already discussed by~\citet{soare2014linear,xu2018linear}, which encapsulates the difficulty of BAI in a linear bandit, and thus is the first instance on which we test our algorithms. In this instance, arms are the canonical basis  $a_1 = e_1, a_2 = e_2, a_d = e_d$, plus an additional disturbing arm $a_{d+1} = (\cos(\alpha), \sin(\alpha), 0, \ldots, 0)^\top$, and a true regression parameter $\theta$ equal to $e_1$. In this problem, the best arm is always $a_1$, but when the angle $\alpha$ is small, the disturbing arm $a_{d+1}$ is hard to discriminate from $a_1$. As already argued by~\citet{soare2014linear}, an efficient sampling rule for this problem instance would rather pull $a_2$ in order to reduce the uncertainty in the direction $a_1-a_{d+1}$. Naive adaptation of classical BAI algorithms cannot deal with that situation naturally. We further use a simple set of experiments to justify that intuition. We run our two algorithms and the one of~\citet{degenne2019game} that we call \texttt{DKM} over the problem instance whence $d=2$, $\delta=0.01$ and $\alpha=0.1$. We show the number of pulls for each arm averaged over 100 replications of experiments in Table~\ref{table:pulls}. We see that, indeed, \texttt{DKM} pulls too much $a_3$, while our algorithms focus mostly on $a_2$.

\begin{table}[th]\centering
\begin{tabular}{|c|c|c|c|}
 \hline
 & \algoW & \algoWCvx & \texttt{DKM} \\
 \hline
 \textbf{$a_1$} & $1912$ & $1959$ & $1943$ \\
 \hline
 \textbf{$a_2$} & $5119$ & $4818$ & $4987$ \\
 \hline
 \textbf{$a_3$} & $104$ & $77$ & $1775$ \\
 \hline
 \textbf{Total} & $7135$ & $\bf{6854}$ & $8705$ \\
 \hline
\end{tabular}
\caption{Average number of pulls of each arm.}
\label{table:pulls}
\end{table}

\vspace{-0.3cm}
\paragraph{Comparison of different complexities.}
We use the previous setting to illustrate various complexities differ in practice. In Table~\ref{table:optimal_weights} we compare the different complexities mentioned in this paper: the characteristic time $\Tstar(\theta)$ and its associated optimal weights $\wstar_{\cA\cBstar(\theta)}$, the $\xyopt$-complexity and its associated optimal design $\wstar_{\xyopt}$, the G-optimal complexity $\gopt$ and its associated optimal design $\wstar_{\cA\cA}$. For each weight vector $w \in\{\wstar_{\cA\cBstar(\theta)},w_{\xyopt}, w_{\gopt}\}$,
 we also provide the lower bound $T_w$ given by Therorem~\ref{th:lb_genral}, i.e.
 \[
 T_w = \max_{a\neq \astar(\theta)} \frac{\big\langle \theta, \astar(\theta)-a\big\rangle^2}{2\normm{\astar(\theta)-a}_{V_w^{-1}}^2} \log(1/\delta).
\]
In particular we notice that targeting the proportions of pulls $w_{\xyopt}, w_{\gopt}$ leads to a much larger lower bound than the one obtained with the optimal weights.
\begin{table}[th]
\centering
\begin{tabular}{|c|c|c|c|}
 \hline
   & $\wstar_{\cA\cBstar}$ & $\wstar_{\xyopt}$  & $\wstar_{\gopt}$   \\
 \hline
 \textbf{$a_1$} & $0.047599$ & $0.499983$ & $0.499983$ \\
 \hline
 \textbf{$a_2$} & $0.952354$ & $0.499983$ & $0.499983$ \\
 \hline
 \textbf{$a_3$} & $0.000047$ & $0.000033$ & $0.000033$ \\
 \hline
 \textbf{$T_w$} & $369$ & $2882$ & $2882$ \\
 \hline
   & $\Tstar(\theta)$ & $2\xyopt/\DeltaMin^2$ & $8\gopt/\DeltaMin^2$\\
 \hline
  \textbf{Complexity} & $0.124607$ & $32.0469$ & $64.0939$ \\
 \hline
\end{tabular}
\caption{Optimal $w$ for various complexities ($\DeltaMin= 0.0049958$).}
\label{table:optimal_weights}
\end{table}
\vspace{-0.3cm}
\paragraph{Comparison with other algorithms.}
Finally we benchmark our two sampling rules against others from the literature. %Note that the main purpose of this paper is to propose algorithms with asymptotic optimality while being practically usable, but we do not claim to have the best performing ones.
We test over two synthetic problem instances, with the first being the previous counter-example. We set $d=2$, $\alpha=\pi/6$. Fig.~\ref{fig:sample_complexity_1} shows the empirical stopping time of each algorithms averaged over 100 runs, with a confidence level $\delta=0.1, 0.01, 0.0001$ from left to right. Our two algorithms show competitive performance (the two leftmost boxes on each plot), and are only slightly worse than \LGapE.

For the second instance, we consider 20 arms randomly generated from the unit sphere $\mathbb{S}^{d-1}\eqdef\{a\in\R^d; \normm{a}_2=1\}$. We choose the two closest arms $a, a'$ and we set $\theta = a + 0.01(a'-a)$ so that a is the best arm. This setting has already been considered by~\citet{tao2018alba}. We report the same box plots over 100 replications as before with increasing dimension in Fig.~\ref{fig:sample_complexity_2}. More precisely, we set $d=6, 8, 10, 12$ respectively, and always keep a same $\delta = 0.01$. Our algorithms consistently show strong performances compared to other algorithms apart from \LGapE. Moreover, we can see that in these random examples, \algoWCvx works better than the non-confexified one, and is even competitive compared to \LGapE.

We stress that although the main focus of this paper is theoretical, with algorithms that are asymptotically optimal, our methods are also competitive with earlier algorithms experimentally.

%!TEX root = ../lin_bandit_explo.tex
\vspace{-0.4cm}
\section{Conclusion}\label{sec:conclusion}
In this paper, we designed the first practically usable asymptotically optimal sampling rules for the pure exploration game for finite-arm linear bandits. Should the boundedness assumption be necessary to have optimal algorithm remains an open question. 

Another concern about the current sampling rules could be their computational complexity. In BAI, the one step complexity of \algoWCvx (or \algoW) is dominated by the computation of the best response for nature, which requires a full matrix inversion. Alternatives that involve rank-1 updates should be considered.

More generally, however, the part of fixed-confidence pure exploration algorithms that needs an improvement the most is the stopping rule. While the one we used guarantees $\delta$-correctness, it is very conservative. Indeed, the experimental error rates of algorithms using that stopping rule are orders of magnitude below $\delta$. This means that the concentration inequality does not reflect the query we seek to answer. It quantifies deviations of the $d$-dimensional estimate in all directions (morally, along $2^d$ directions). However, for the usual BAI setting with $d$ arms in an orthogonal basis, it would be sufficient to control the deviation of that estimator in $d-1$ directions to make sure that $i^*(\theta) = i^*(\hat{\theta}_t)$.

Finally, the good performance of \LGapE raises the natural question of whether it could be proven to have similar asymptotic optimality.

\section*{Acknowledgements}
The research presented was supported by European CHIST-ERA project DELTA, French Ministry of
Higher Education and Research, Nord-Pas-de-Calais Regional Council,  and by French National Research Agency as part of the ``Investissements d'avenir'' program, reference ANR19-P3IA-0001 (PRAIRIE 3IA Institute) and the project BOLD, reference ANR19-CE23-0026-04.
%\textbf{Do not} include acknowledgements in the initial version of the paper submitted for blind review.

%If a paper is accepted, the final camera-ready version can (and probably should) include acknowledgements. In this case, please place such acknowledgements in an unnumbered section at the end of the paper. Typically, this will include thanks to reviewers who gave useful comments, to colleagues who contributed to the ideas, and to funding agencies and corporate sponsors that provided financial support.

\bibliographystyle{icml2020}
\bibliography{Major.bib,manual_bib}

\appendix
\onecolumn
%!TEX root = ../lin_bandit_explo.tex
\section{Outline}

The appendices are organized as follows:
\begin{itemize}[label=$\square$]
    \item Notation is given in Appendix~\ref{app:notations}.
    \item We give a full proof of Theorem~\ref{th:lb_genral} in Appendix~\ref{app:lower_bound}.
    \item We list and elaborate in detail several pure exploration problems in Appendix~\ref{app:examples}.
    \item Appendix~\ref{app:proof} is dedicated to the proof of sample complexity for \algoWCvx and \algoW.
    %\item Appendix~\ref{app:proof_nc} is dedicated to the proof of sample complexity for the non-convexified algorithm \algoW.
    \item Appendix~\ref{app:concentration} is dedicated to some important concentration results.
    \item We discuss the tracking procedure in Appendix~\ref{app:tracking}.
    \item We discuss the stopping rules in Appendix~\ref{app:stopping}.
    \item Finally, we provide some further details about the experiments in Appendix~\ref{app:implem}.
\end{itemize}

%!TEX root = ../lin_bandit_explo.tex
\section{Notation}\label{app:notations}

\begin{table}[h]
	\centering
	\caption{Table of notation}
	\begin{tabular}{@{}l|l@{}}
		\toprule
		\thead{Notation} & \thead{Meaning} \\ \midrule
		$\cM$ & set of parameters \\
        $M$ & upper bound on the norm of $\theta$\\
		$\cA$ & finite set or arms  \\
        $A$ & number of arms \\
        $\cB$ & transductive set \\
        $B$ & number of elements in the transductive set \\
        $\cI$ & finite set of answers \\
        $I$ & number of answers \\
        $L$ & upper bound on the norms of the arms\\
        $\theta$ & parameter in $\cM$ \\
        $a_t$ & arm pulled at time $t$ \\
				$i_t$ & answer chosen at time $t$\\
        $N_t^{a,i} = \sum_{s=1}^t \ind_{\{a_s = a, i_t = i\}}$ & number of draws of arm $a$ for a given answer $i$\\
        $\eta$ & regularization parameter\\
        $\htheta_t$ & regularized least square estimate\\
				$w_s$ & weights in $\Sigma_A$ played by the agent in \algoW\\
				$w_s^{a,i} = w_s \ind_{\{i_s=i\}}$ & weights for arm $a$ and answer $i$ in \algoW\\
				$W_t^{a,i} = \sum_{s=1}^t w_s^{a,i}$& cumulative sum of weights in \algoW\\
				$w_s^{a} = \sum_{i\in\cI} w_s^{a,i}\text{ or } (w_s^{a,i})_{i\in\cI}$ & partial weights or vector of partial weights in \algoW \\
				$w_s^{i} = \sum_{a\in\cA} w_s^{a,i}\text{ or } (w_s^{a,i})_{a\in\cA}$ & partial weights or vector of partial weights in \algoW \\
				$\tw_s$ & weights in $\Sigma_{AI}$ played by the agent in \algoWCvx\\
				$\tW_t^{a,i} = \sum_{s=1}^t \tw_s^{a,i}$ & cumulative sum of weights in \algoWCvx\\
				$\tw_s^{a} = \sum_{i\in\cI} \tw_s^{a,i}\text{ or } (\tw_s^{a,i})_{i\in\cI}$ & partial weights or vector of partial weights in \algoWCvx \\
				$\tw_s^{i} = \sum_{a\in\cA} \tw_s^{a,i}\text{ or } (\tw_s^{a,i})_{a\in\cA}$ & partial weights or vector of partial weights in \algoWCvx \\
				$U_s^{a,i}$ & upper confidence bounds to build the optimistic gain, see~\eqref{eq:def_ucb_br_general}\\
		\bottomrule
	\end{tabular}
\end{table}

%!TEX root = ../lin_bandit_explo.tex
\section{Proofs of the Lower Bounds and Equivalent Formulations}\label{app:lower_bound}

We start this section by proving the lower bound.

\begin{proof}[Proof of Theorem~\ref{th:lb_genral}]
Fix $\lambda\in\neg\istar(\theta)$ in the alternative of $\theta$. Thanks to the contraction of the entropy and the chain rule (see \citealt{garivier2018explore}), we have

\[
\kl\!\big(\P_\theta(\hi = \istar),\P_\lambda(\hi = \istar)\big) \leq \sum_{a\in\cA} \E_\theta\big[N_a(\tau_\delta)\big]\frac{\normm{\theta -\lambda^i}_{a a^\top}^2}{2}\,,
\]
where we denote by $\kl(p,q)$ the Kullback-Leibler divergence between two Bernoulli distributions $\Ber(p)$ and $\Ber(q)$
\[
\kl(p,q) = p \log\!\left(\frac{p}{q}\right) + (1-p) \log\!\left(\frac{1-p}{1-q}\right)\,.
\]
Since the algorithm is $\delta$-correct we know that
\[
\P_\lambda\big(\hi = \istar(\theta)\big) \leq \delta \leq \frac{1}{2} \leq  1-\delta \leq \P_\theta\big(\hi = \istar(\theta)\big)\,.
\]
Thanks to monotonic properties of the function $\kl(\cdot,\cdot)$ and the inequality $\kl(1-p,p) \geq -\log(2.4p)$ (see \citealt{garivier2018explore}), it yields
\[
\kl\big(\P_\theta(\hi = \istar(\theta)),\P_\lambda(\hi = \istar(\theta))\big) \geq \kl(1-\delta,\delta) \geq \log\left(\frac{1}{2.4\delta}\right)\,,
\]
thus
\[
\log\left(\frac{1}{2.4\delta}\right) \leq \sum_{a\in\cA} \E_\theta\big[N_a(\tau_\delta)\big]\frac{\normm{\theta -\lambda^i}_{a a^\top}^2}{2}\,.
\]
Using the fact that the previous inequality is true for all $\lambda\in\neg\istar(\theta)$ and that the vector of components $\E_\theta\big[N_a(\tau_\delta)\big]/\E_\theta\big[\tau_\delta\big]$ belongs to the probability simplex $\Sigma_A$ we get
\begin{align*}
\log\left(\frac{1}{2.4\delta}\right) &\leq \E_\theta[\tau_\delta] \inf_{\lambda\in\neg \istar(\theta)} \sum_{a=1}^K \frac{\E_\theta\big[N_a(\tau_\delta)\big]}{\E_\theta[\tau_\delta]}\frac{\normm{\theta -\lambda^i}_{a a^\top}^2}{2}\\ &\leq \E_\theta[\tau_\delta] \sup_{w\in\Sigma_K}
\inf_{\lambda\in\neg \istar(\theta)} \sum_{a=1}^K w_a\frac{\normm{\theta -\lambda^i}_{a a^\top}^2}{2}\,.
\end{align*}
Dividing the previous inequality by $\log(1/\delta)$ and taking the limit inferior when $\delta$ goes to zero allows us to conclude.
\end{proof}

We restate and prove below Lemma~\ref{lem:sion_convexify}.
\begin{lemma}
  \label{lem:sion_convexify_extended} For all $\theta \in\cM$,
\begin{align}
  \Tstar(\theta)^{-1} &= \max_{i\in \cI} \max_{w \in \Sigma_A} \inf_{\lambda\in \neg i} \frac{\normm{\theta - \lambda}_{V_w}^2}{2}\label{eq:sion_base}\\
  &=\max_{\tw \in \Sigma_{AI}} \inf_{\tlambda\in \prod_i (\neg i) }\frac{1}{2} \sum_{(a,i)\in\cA\times\cI}\tw^{a,i}\normm{\theta - \lambda^i}_{aa^\top}^2\label{eq:sion_nature_first}\\
  & = \max_{\tw \in \Sigma_{AI}}\inf_{\tq\in \prod_i\cP(\neg i) }\!\sum_{(a,i)\in\cA\times\cI}\!\!\!\tw^{a,i}\E_{\lambda^i\sim \tq^i}\normm{\theta - \lambda^i}_{aa^\top}^2 \label{eq:sion_both}\\
  &= \inf_{\tq\in \prod_i\cP(\neg i) } \frac{1}{2}\max_{(a,i)\in\cA\times\cB}\E_{\lambda^i\sim \tq^i}\normm{\theta - \lambda^i}_{aa^\top}^2\label{eq:sion_player_fisrt}\,.
\end{align}

\end{lemma}
\begin{proof}
  The transition from \eqref{eq:sion_nature_first} to \eqref{eq:sion_both} comes from the fact that the second player can use indifferently mixed or pure strategy. The equality between \eqref{eq:sion_both} and \eqref{eq:sion_player_fisrt} is just an application of the Sion lemma (see \citealt{degenne2019pure}).
  It remains to prove \eqref{eq:sion_base} = \eqref{eq:sion_nature_first}. First note that we can replace the first maximum in \eqref{eq:sion_base} over $\istar(\theta)$ by a maximum over $\cI$ because when $i\notin \istar(\theta)$ we know that $\theta\in \neg i$. Since we can express the maximum over the answers as a maximum over the probability simplex $\Sigma_I$ we have
\begin{align*}
    \max_{i\in \cI} \sup_{w \in \Sigma_A} \inf_{\lambda\in \neg i} \frac{1}{2}\sum_{a\in\cA} w_a \normm{\theta - \lambda}_{aa^\top}^2 &= \max_{\rho \in \Sigma_I} \sum_{i\in\cI}\sup_{w \in \Sigma_A} \inf_{\lambda\in \neg i} \frac{1}{2}\sum_{a\in\cA} \rho_i w_a\normm{\theta - \lambda}_{aa^\top}^2\\
     &= \max_{\rho\in \Sigma_I} \sum_{i\in \cI} \sup_{w^i \in \Sigma_A} \inf_{\lambda^i\in \neg i}\frac{1}{2} \sum_{a\in\cA} \rho_i w_a^i\normm{\theta - \lambda^i}_{aa^\top}^2\\
     &= \sup_{\tw \in \Sigma_{A I}} \inf_{\tlambda \in \prod_{i\in\cI}\neg i} \frac{1}{2}\sum_{(a,i)\in \cA\times\cI}  \tw_a^i \normm{\theta - \tlambda^i}_{aa^\top}^2\,,
\end{align*}
where for the last line we use the fact that all $\tw \in \Sigma_{A I}$ can be written as $\tw_a^i = \rho_i w_a^i $ with $\rho \in \Sigma_I$ and $w^i \in \Sigma_A$ for all $i \in \cI$.
\end{proof}

%!TEX root = ../lin_bandit_explo.tex
\section{Examples}\label{app:examples}

We gather in this appendix several pure exploration problems for linear bandits. We first state a useful lemma.

\begin{lemma}\label{lem:lagrange_alternative}
For $\theta, \lambda \in \R^d\,$, $w$ in the interior of the probability simplex $\interior{\Sigma_A}$, $y\in\R^d\,$, $x\in\R$, we have
\begin{align*}
\inf_{\lambda:\ \langle \lambda,y\rangle \geq x} \frac{\normm{\theta-\lambda}^2_{V_w}}{2} = \begin{cases}
\dfrac{(x - \langle\theta,y\rangle)^2}{2 \normm{y}_{V_w^{-1}}^2} &\text{if } x \geq \langle\theta,y\rangle \\
0 &\text{otherwise}
\end{cases}\,.
\end{align*}
\end{lemma}

\begin{proof}
We consider the Lagrangian of the problem, and we obtain
\begin{align*}
  \inf_{\lambda:\ \langle \lambda,y \rangle \geq x} \frac{\normm{\theta-\lambda}^2_{V_w}}{2}
  &= \sup_{\alpha \geq 0}\inf_{\lambda \in \R^d} \frac{\normm{\theta-\lambda}^2_{V_w}}{2}+ \alpha (x-\langle\lambda,y\rangle)\\
  &=  \sup_{\alpha \geq 0} \alpha (x-\langle\theta,y\rangle) - \alpha^2 \frac{\normm{y}^2_{V_w^{-1}}}{2}\\
  &= \begin{cases}
  \dfrac{(x - \langle\lambda,y\rangle)^2}{2 \normm{y}_{V_w^{-1}}^2} &\text{if } x \geq \langle\theta,y\rangle \\
  0 & \text{otherwise}
  \end{cases}\,,
\end{align*}
where the infimum in the first equality is reached at $\lambda = \theta + \alpha V_w^{-1} y$ and the supremum in the last equality is reached at $\alpha = (x- \langle\theta,y\rangle)/\normm{y}_{V_w^{-1}}^2$ if $x \geq \langle\theta,y\rangle$ and at $\alpha = 0$ else.
\end{proof}

\subsection{Best-arm identification}
\label{app:bai}
For BAI the goal is to identify the arm with the largest mean. Thus, the set of parameters is $\cM=\cR^d/\{\theta\in\R^d:\  |\argmax_{a\in\cA} \langle\theta,a\rangle|>1\}$, the set of possible answers is $\cI = \cA$ and the correct answer is given by $\istar(\theta)=\astar(\theta)\eqdef \argmax_{a\in\cA} \langle\theta,a\rangle$.
\begin{lemma}
\label{lem:complexity_bai}
For all $\theta\in \cM$,
\[
\Tstar(\theta)^{-1} = \max_{w\in\Sigma_A} \min_{a\neq \astar(\theta)} \frac{\big\langle \theta, \astar(\theta)-a\big\rangle^2}{2 \normm{\astar(\theta)-a}_{V_w^{-1}}^2}\,,
\]
and
\[
\Tstar(\theta) = \min_{w\in\Sigma_A} \max_{a\neq \astar(\theta)} \frac{2\normm{\astar(\theta)-a}_{V_w^{-1}}^2}{\big\langle \theta, \astar(\theta)-a\big\rangle^2}\,.
\]
\end{lemma}
\begin{proof}
Recall that the characteristic time is given by
\[
\Tstar(\theta)^{-1} = \max_{w \in \Delta_A} \inf_{\lambda\in \neg \astar(\theta)} \frac{\normm{\theta - \lambda}_{V_w}^2}{2}\,.
\]
We just express the set $\neg \astar(\theta)$ as a union of convex sets, and then compute the infimum for each one of them. Using Lemma~\ref{lem:lagrange_alternative}, it yields
\begin{align*}
  \Tstar(\theta)^{-1} &= \max_{w \in \Delta_A} \min_{a \neq \astar(\theta)} \inf_{\lambda: \langle\lambda,a\rangle > \langle\lambda,\astar(\theta)\rangle} \frac{\normm{\theta - \lambda}_{V_w}^2}{2}\\
  &= \max_{w \in \Delta_A} \min_{a \neq \astar(\theta)} \frac{\big\langle \theta, \astar(\theta)-a\big\rangle^2}{2 \normm{\astar(\theta)-a}_{V_w^{-1}}^2}\,.
\end{align*}
The formula for $\Tstar(\theta)$ is then straightforward given the one for  $\Tstar(\theta)^{-1}$.

\end{proof}
In fact the characteristic time is just a particular case of the optimal transductive design. Indeed if we set
\[
\cBstar(\theta) \eqdef \left\{ \frac{1}{\left|\big\langle \theta, \astar(\theta)-a\big\rangle\right|}\big(\astar(\theta)- a\big): a\in\cA/\big\{\astar(\theta)\big\}  \right\}\,,
\]
then we have $\Tstar(\theta) = 2 \cA\cBstar(\theta)$ where
\[
\cA\cBstar(\theta) \eqdef  \min_{w\in\Sigma_A} \max_{b\in \cBstar} \normm{b}_{V_w^{-1}}^2\,.
\]

\paragraph{Best response.} There is an explicit formula for the best response in BAI. Indeed if we inspect the proof of Lemma~\ref{lem:complexity_bai} we have
\[
\inf_{\lambda\in \neg \astar(\theta)} \frac{\normm{\theta - \lambda}_{V_w}^2}{2} = \min_{a \neq \astar(\theta)} \normm{\theta-\lambda^\star_a}_{V_w^{-1}}\,.
\]
where $\lambda^\star_a$ is defined in Lemma~\ref{lem:best_response_BAI}.

\begin{lemma}
\label{lem:best_response_BAI}
For $\theta \in\R^d\,$, $w$ in the interior of the probability simplex $\interior{\Sigma_A}$, we have
\[
\min_{\substack{\langle \lambda,a-\astar(\theta)\rangle\geq 0}} \normm{\theta -\lambda }_{V_w}^2 = \frac{\big\langle \theta, \astar(\theta)-a\big\rangle^2}{2 \normm{\astar(\theta)-a}_{V_w^{-1}}^2}\,,
\]
and $\lambda^\star_a$ defined below attains the infimum of the left hand term above
\[
\lambda^\star_a = \theta - \frac{\max\left(\langle \theta, \astar(\theta)-a\rangle,0\right)}{\normm{\astar(\theta)-a}^2_{(V_w+\gamma I_d)^{-1}}} V_w^{-1}(a^\star - a)\,.
\]
\end{lemma}
\begin{proof}
See proof of Lemma~\ref{lem:lagrange_alternative}.
\end{proof}

\subsubsection{Bounded BAI}
\label{app:bounded_bai}
One straightforward extension of this setting is to consider the \emph{bounded} BAI. In this case, the set of parameters is $\cM=\{\theta \in \R^d:\ |\argmax_{a\in\cA} \langle\theta,a\rangle|=1 \text{ and } \normm{\theta}\leq M\}$ for some $M>0$. The set of possible answers is $\cI = \cA$ and the correct answer is given by $\istar(\theta)=\astar(\theta)\eqdef \argmax_{a\in\cA} \langle\theta,a\rangle$.
This additional assumption reduces the characteristic time to
\[
\Tstar(\theta)^{-1} = \max_{w\in\Sigma_A} \min_{a\neq \astar(\theta)} \inf_{\substack{\langle \lambda,a-\astar(\theta)\rangle>0\\ \normm{\lambda}\leq M}} \normm{\theta -\lambda }_{V_w}^2 \,.
\]
But the best response is less trivial to compute, in particular there is no closed formula for $\lambda^\star_a$ as in BAI, see Lemma~\ref{lem:lagrange_bounded_BAI}.
\begin{lemma}
  \label{lem:lagrange_bounded_BAI}
For $\theta, \lambda \in \R^d\,$, $w$ in the interior of the probability simplex $\interior{\Sigma_A}$,
\begin{align}\label{eq:bounded_bai}
\min_{\substack{\langle \lambda,a-\astar(\theta)\rangle\geq 0\\ \normm{\lambda}\leq M}} \normm{\theta -\lambda }_{V_w}^2 = \sup_{\gamma\geq 0} \frac{\max\left(\langle \theta, (V_w+\gamma I_d)^{-1} V_w (\astar(\theta)-a)\rangle,0\right)^2 }{2\normm{\astar(\theta)-a}^2_{(V_w+\gamma I_d)^{-1}}}- \frac{\gamma}{2}\left(\normm{\theta}^2-M^2\right)\,.
\end{align}
And if $\gamma$ attains the supremum in the right hand term of~\eqref{eq:bounded_bai}, then
\[
\lambda = \theta - \frac{\max\left(\langle \theta, (V_w+\gamma I_d)^{-1} V_w (\astar(\theta)-a)\rangle,0\right)}{\normm{\astar(\theta)-a}^2_{(V_w+\gamma I_d)^{-1}}} (V_w+\gamma I_d)^{-1}(a^\star - a)\,,
\]
attains the minimum of the left hand term of~\eqref{eq:bounded_bai}.
\end{lemma}
\begin{proof}
We set $\astar(\theta) = \astar$, and introduce the Lagrangian
\[
 \inf_{\substack{\langle \lambda,a-\astar \rangle>0\\ \normm{\lambda}\leq M}} \normm{\theta -\lambda }_{V_w}^2 = \sup_{\gamma\geq 0, \alpha\geq 0} \inf_{\substack{\langle \lambda,a-\astar\rangle>0\\ \normm{\lambda}\leq M}} \normm{\theta -\lambda }_{V_w}^2 +\alpha \langle \theta, \astar-a\rangle + \frac{\gamma}{2}\left(\normm{\lambda}^2-M^2 \right)\,.
\]
The infimum above is attained for
\[
\lambda = \theta - \alpha (V_w + \gamma I_d)^{-1}(\astar-a)\,.
\]
Thus the Lagrangian reduces to
\[
\inf_{\substack{\langle \lambda,a-\astar \rangle>0\\ \normm{\lambda}\leq M}} \normm{\theta -\lambda }_{V_w}^2 = \sup_{\gamma\geq 0, \alpha\geq 0}
-\frac{\alpha^2}{2} \normm{\astar-a}^2_{V_w+\gamma I_d} + \alpha \langle \theta, (V_w+\gamma I_d)^{-1}V_w (\astar-a)\rangle +\frac{\gamma}{2}\left(\normm{\theta}^2-M^2 \right)\,.
\]
The supremum in $\alpha$ is reached for
\[
\alpha =\frac{\max\left(\langle \theta, (V_w+\gamma I_d)^{-1} V_w (\astar-a)\rangle,0\right)}{\normm{\astar-a}^2_{(V_w+\gamma I_d)^{-1}}}\,.
\]
Using this particular $\alpha$ in the definition of $\lambda$ and in the Lagrangian allows us to conclude.
\end{proof}

\subsubsection{Transudctive BAI}

We can also consider the transductive BAI~\citep{fiez2019transductive} where the agent wants to find the best arm of a different set $\cB$ that the one he is allow to pull. Precisely the set of parameters is $\cM=\cR^d/\{\theta\in\R^d:\  |\argmax_{b\in\cB} \langle\theta,b\rangle|>1\}$, the set of possible answers is $\cI = \cB$ and the correct answer is given by $\istar(\theta)=\bstar(\theta)\eqdef \argmax_{b\in\cB} \langle\theta,b\rangle$.

The characteristic time in this case is
\[
\Tstar(\theta)^{-1} = \max_{w\in\Sigma_A} \min_{b\neq \bstar(\theta)} \frac{\big\langle \theta, \bstar(\theta)-b\big\rangle^2}{2 \normm{\bstar(\theta)-b}_{V_w^{-1}}^2}\,.
\]
Note that the dependency on the arm set $\cA$ here only appears through the matrix $V_w$.
\subsection{Threshold bandits}
\label{app:threshold_bandits}
In this example the goal is to identify the set of arms whose mean is above a threshold $\iota\in \R$ known by the agent. Thus, the set of parameters is $\cM=\cR^d/\{\theta\in\R^d:\ \exists a\in \cA,\, \langle \theta,a\rangle = \iota\}$, the set of possible answers is $\cI = \cP(\cA)$, the power set of the set of arms and the correct answer is given by $\istar(\theta)=\{a\in\cA:\ \langle \theta,a\rangle \geq \iota\}$.
We can also express in this example the characteristic time in a more explicit way.
\begin{lemma}
\label{lem:complexity_threshold_bandits}
For all $\theta\in \cM$,
\[
\Tstar(\theta)^{-1} =  \max_{w\in\Sigma_A} \min_{a\in\cA} \frac{\big(\iota -\langle \theta,a\rangle\big)^2}{2 \normm{a}_{V_w^{-1}}^2}\,,
\]
and $\Tstar(\theta)= 2\cA\cA(\iota)$, where we define $\cA(\iota)\eqdef \{ |\iota- \langle\theta,a\rangle|^{-1} a:\ a\in\cA\}$ and
\[
\cA\cA(\iota) \eqdef  \min_{w\in\Sigma_A} \max_{a\in\cA(\iota)}\normm{a}_{V_w^{-1}}^2\,.
\]
\end{lemma}
\begin{proof}
We proceed as the proof of Lemma~\ref{lem:complexity_bai}. We have, using Lemma~\ref{lem:lagrange_alternative},
\begin{align*}
  \Tstar(\theta)^{-1} &= \max_{w \in \Delta_A} \min_{a\in\cA} \inf_{\lambda:\ \text{sign}(\iota-\langle\lambda,a\rangle)\langle\lambda,a\rangle > \iota} \frac{\normm{\theta - \lambda}_{V_w}^2}{2}\\
  &= \max_{w \in \Delta_A} \min_{a\in\cA}  \frac{\big(\iota -\langle \theta,a\rangle\big)^2}{2 \normm{a}_{V_w^{-1}}^2}\,.
\end{align*}
\end{proof}
Note that we recover in this example a weighted version of the G-complexity ($\gopt$-complexity) defined in Section~\ref{sec:lower_bound}. In particular if $\theta=0$ and $\iota=1$ then
\[
\Tstar(\theta) =2\gopt = 2d\,.
\]
That makes sense since in this case, one shall estimate \emph{uniformly} the mean of each arms.

\subsubsection{Transductive threshold bandits}
\label{app:transductive_threshold_bandits}
We can generalize the previous example to any set of arms. Indeed if we fix a finite set of vector $\cB\in\R^d$ the goal is then to identify all the elements $b$ of this set such that $\langle \theta, b \rangle \geq \iota$ for a known threshold $\tau \in \R$. Thus, the set of parameters is $\cM=\cR^d/\{\theta\in\R^d:\ \exists b\in \cB,\, \langle \theta,b\rangle = \iota\}$, the set of possible answers is $\cI = \cP(\cB)$ and the correct answer is given by
$\istar(\theta)=\{b\in\cB:\ \langle \theta,b\rangle \geq \iota\}$. The characteristic time makes appear, unsurprisingly, in this case, the transductive optimal design \citep{yu2006active}.
\begin{lemma} For all $\theta \in\cM$,
  \label{lem:complexity_transductive_threshold_bandits}
  \[
  \Tstar(\theta)^{-1} =  \max_{w\in\Sigma_A} \min_{b\in\cB} \frac{\big(\iota -\langle \theta,b\rangle\big)^2}{2 \normm{b}_{V_w^{-1}}^2}\,,
  \]
  and $\Tstar(\theta)= 2\cA\cB(\iota)$, where we defined $\cB(\iota)\eqdef \{ |\iota- \langle\theta,b\rangle|^{-1} b:\ b\in\cB\}$ and
  \[
  \cA\cB(\iota) \eqdef  \min_{w\in\Sigma_A} \max_{b\in\cB(\iota)}\normm{b}_{V_w^{-1}}^2\,.
  \]
\end{lemma}
\begin{proof}
  Simple adaptation of the proof of Lemma~\ref{lem:complexity_threshold_bandits}.
\end{proof}
Again, in particular, if $\theta=0$ and $\tau=1$ we recover the complexity of the optimal transductive design
\[
\Tstar(\theta)^{-1} = 2 \cA\cB\,.
\]

%\subfile{appendix/oracle_computations}
%!TEX root = ../lin_bandit_explo.tex
\section{Proof for the Sample Complexity}\label{app:proof}

In this section we prove the asymptotic optimality of \algoWCvx and \algoW.
\subsection{Events}\label{app:proof.events}
We fix a constant $\alpha>2$ and define the event where the least square estimator is concentrated around the true parameter,
\[
\cE_t = \left\{\forall s \leq  t:\ \frac{1}{2}\normm{\htheta_s-\theta}^2_{V_{N_s}+\eta I_d} \leq h(t)\eqdef \beta(t,1/t^\alpha)\right\}\,.
\]
This event holds with high probability.
\begin{lemma}
\label{lem:prb_Et}
For all $t \geq 1$
\[
\P_\theta\left(\cE_t^c\right)\leq \frac{1}{t^{\alpha-1}}\,.
\]
\end{lemma}

\paragraph{Optimistic loss.} We need to build an upper confidence bound on the true gain for \algoWCvx and \algoW respectively at time $s$, for all $\tw\in\Sigma_{AI}$ or all $w\in\Sigma_{A}$,
\[
g_{s}^{\theta}(\tw) = \frac{1}{2} \sum_{(a,i)\in\cA\times\cI} \tw^{a,i} \normm{\theta - \tlambda_t^i}_{V_{\tw}}^2\quad
g_{s}^{\theta}(w) = \frac{1}{2} \sum_{a\in\cA\times\cI} w^{a} \normm{\theta - \lambda_t^{i_s}}_{V_{\tw}}^2\,,
\]
where $\lambda_s^i \in\argmin_{\lambda\in\neg i} \normm{\htheta_{s-1}-\lambda}_{V_{w_s}}$.
For that, we just build a confidence bound for each term that appears in the right hand sum.
\begin{lemma}
\label{lem:confidence_bound_general}
On the event $\cE_t$, for all $a\in\cA$ and $\lambda\in\cM$, for all $s\leq t$,
\[
\normm{\theta-\lambda}^2_{aa^\top} \leq \min\left(\max_\pm \left( \langle \htheta_{s} - \lambda,a\rangle \pm \sqrt{2 h(t)} \normm{a}_{(V_{N_s}+ \eta I_d)^{-1}} \right)^2,4L^2 M^2\right)\,.
\]
\end{lemma}
\begin{proof}
First, note that since $\theta,\lambda\in\cM$, their norms are bounded by $M$, thus it holds
\[
\normm{\theta-\lambda}^2_{aa^\top} = \langle\theta-\lambda,a \rangle^2 \leq \normm{\theta-\lambda}^2 \normm{\lambda}^2 \leq 4M^2L^2\,.
\]
Furthermore on $\cE_t$ we have
\begin{align*}
\normm{\theta-\lambda}^2_{aa^\top} = \langle\theta-\lambda,a \rangle^2 &\leq \sup_{\big\{\theta':\ \normm{\htheta_s-\theta'}^2_{(V_{N_s} + \eta I_d)^{-1} } \leq 2h(t)\big\}} \langle\theta'-\lambda,a \rangle^2\\
&=\max_\pm \left( \langle \htheta_{s} - \lambda,a\rangle \pm \sqrt{2 h(t)} \normm{a}_{(V_{N_s}+ \eta I_d)^{-1}} \right)^2\,.
\end{align*}
Combining the two inequalities above allows us to conclude.
\end{proof}
Thus we define the upper confidence $U_s^{a,i}$ on the coordinate $(a,i)$ of the loss at time $s\leq t$ by
\begin{equation}
\label{eq:def_ucb_br_general}
U_s^{a,i} = \min\left(\max_\pm \left( \langle \htheta_{s-1} - \lambda,a\rangle \pm \sqrt{2 h(t)} \normm{a}_{(V_{N_{s-1}}+ \eta I_d)^{-1}} \right)^2,4L^2 M^2\right)\,,
\end{equation}
where $\lambda = \tlambda_s^i$ for \algoWCvx and $\lambda = \lambda_s^i$ for \algoW. The optimistic gains are: $g_s(\tw) = \sum_{(a,i)\in\cA\times\cI} \tw^{a,i} U_s^{a,i}/2$ for \algoWCvx and  $g_s(w) = \sum_{a\in\cA} w^{a} U_s^{a,i_s}/2$ for \algoW.

\paragraph{Analysis.} The first step of our analysis is to restrict it to the event $\mathcal E_t$, as done by \citet{garivier2016tracknstop,degenne2019game}.
\begin{lemma}
\label{lem:start_analysis}
Let $\mathcal E_t$ be an event and $T_0(\delta) \in \N$ be such that for $t\ge T_0(\delta)$, $\cE_t \subseteq \{\tau_\delta \le t\}$. Then
\begin{align*}
\mathbb{E}[\tau_\delta] \leq T_0(\delta) + 1 + \frac{2^{\alpha-2}}{\alpha-2}\,.
\end{align*}
\end{lemma}
\begin{proof}
We have, using Lemma~\ref{lem:prb_Et},
\begin{align*}
\E_\theta[\tau_\delta] &= \sum_{t=0}^{+\infty} \P(\tau_\delta \leq t) = T_0(\delta) + \sum_{t=T_0(\delta)}^{+\infty} \P(\cE_t^c)\\
&\leq T_0(\delta) + \sum_{t=1}^{+\infty} \frac{1}{t^{\alpha-1}} \leq  T_0(\delta) + 1 + \frac{2^{\alpha-2}}{\alpha-2}\,,
\end{align*}
where we use an integral-sum comparison for the last inequality.
\end{proof}

We need to prove that if $\mathcal E_t$ holds, there exists such a time $T_0(\delta)$ of order $\Tstar(\theta)^{-1}\log(1/\delta)+o\big(\log(1/\delta)\big)$. The proof is given in the next section.

\subsection{Analysis under concentration of \algoWCvx}
\label{app:analysis_under_concentration_cvx}

% \hrule
% Check this:
% \begin{align*}
% \sum_{i\in \mathcal I} \inf_{\lambda_i\in \neg i} \frac{1}{2}\Vert \hat{\theta}_t - \lambda_i \Vert^2_{V_{N_t^i}}
% &\le \inf_{\lambda\in \neg i^*} \frac{1}{2}\Vert \hat{\theta}_t - \lambda \Vert^2_{V_{N_t^{i^*}}}
% 	 + \sum_{i\neq i^*} \frac{1}{2}\Vert \hat{\theta}_t - \theta \Vert^2_{V_{N_t^i}}
% \\
% &\le \inf_{\lambda\in \neg i^*} \frac{1}{2}\Vert \hat{\theta}_t - \lambda \Vert^2_{V_{N_t}}
% 	 + \frac{1}{2}\Vert \hat{\theta}_t - \theta \Vert^2_{V_{N_t}}
% \\
% &\le \max_i \inf_{\lambda\in \neg i} \frac{1}{2}\Vert \hat{\theta}_t - \lambda \Vert^2_{V_{N_t}}
% 	 + \frac{1}{2}\Vert \hat{\theta}_t - \theta \Vert^2_{V_{N_t}}
% \\
% &\le \max_i \inf_{\lambda\in \neg i} \frac{1}{2}\Vert \hat{\theta}_t - \lambda \Vert^2_{V_{N_t}}
% 	 + h(t)
% \: .
% \end{align*}
% \todo[inline]{Todo: insert that into the flow of the proof.}
% \hrule
We assume in this section that the event $\cE_t$ holds. If the algorithm does not stop at stage $t$, then it holds
\begin{align*}
\beta(t,\delta)
\ge\max_{i\in\cI} \inf_{\lambda_i \in \neg i}\frac{1}{2} \Vert \hat{\theta}_t - \lambda_i \Vert_{V_{N_t}}^2
%= \frac{1}{2} \inf_{\lambda \in \neg i_t}\Vert \hat{\theta}_t - \lambda \Vert_{V_{N_t}}^2
%\ge \frac{1}{2}\sum_{i\in\cI} \inf_{\lambda_i \in \neg i}\Vert \hat{\theta}_t - \lambda_i \Vert_{V_{N_t^i}}^2
\,.
\end{align*}
However, by definition of the event $\cE_t$ we have
\begin{align*}
\sum_{i\in \mathcal I} \inf_{\lambda_i\in \neg i} \frac{1}{2}\Vert \hat{\theta}_t - \lambda_i \Vert^2_{V_{N_t^i}}
&\le \inf_{\lambda\in \neg \istar(\theta)} \frac{1}{2}\Vert \hat{\theta}_t - \lambda \Vert^2_{V_{N_t^{\istar(\theta)}}}
	 + \sum_{i\neq \istar(\theta)} \frac{1}{2}\Vert \hat{\theta}_t - \theta \Vert^2_{V_{N_t^i}}
\\
&\le \inf_{\lambda\in \neg i^*} \frac{1}{2}\Vert \hat{\theta}_t - \lambda \Vert^2_{V_{N_t}}
	 + \frac{1}{2}\Vert \hat{\theta}_t - \theta \Vert^2_{V_{N_t}}
\\
&\le \max_i \inf_{\lambda\in \neg i} \frac{1}{2}\Vert \hat{\theta}_t - \lambda \Vert^2_{V_{N_t}}
	 + \frac{1}{2}\Vert \hat{\theta}_t - \theta \Vert^2_{V_{N_t}}
\\
&\le \max_i \inf_{\lambda\in \neg i} \frac{1}{2}\Vert \hat{\theta}_t - \lambda \Vert^2_{V_{N_t}}
	 + h(t)
\:,
\end{align*}
thus one obtains
\[
\beta(t,\delta) + h(t) \ge \sum_{i\in \mathcal I} \inf_{\lambda_i\in \neg i} \frac{1}{2}\Vert \hat{\theta}_t - \lambda_i \Vert^2_{V_{N_t^i}}\,.
\]
% where the equality is due to the fact that all the terms of the sum are 0 ($\lambda_i = \hat{\theta}_t$) except for one $i\in \mathcal I$. The last inequality is simply $N_t^a \geq N_t^{i,a}$ for all $a\in \mathcal A$.
Hence we need to find a lower bound for the right hand sum.
Let $\lambda_{i,w}(\theta) \in \argmin_{\lambda \in \neg i} \Vert \theta - \lambda \Vert_{V_{w}}$ . %First remark that $\Vert \theta - \lambda_{i,N_t^i}(\hat{\theta}_t) \Vert_{V_{N_t^i}} \ge \Vert \theta - \lambda_{i,N_t^i}(\theta) \Vert_{V_{N_t^i}}$ by definition.
\begin{lemma}On $\cE_t$, if the algorithm does not stop at $t$,
\begin{equation}
	\label{eq:to_theta_cvx}
	\beta(t,\delta) + \sqrt{4 h(t) \beta(t,\delta)} +4h(t) \geq \frac{1}{2}\sum_{i\in\cI}  \Vert \theta - \lambda_{i,N_t^i}(\hat{\theta}_t) \Vert_{V_{N_t^i}}^2 \,.
\end{equation}
\end{lemma}

\begin{proof}
Using the triangular inequality,
\begin{align*}
\Vert \theta - \hat{\theta}_t \Vert_{V_{N_t^i}} + \Vert \hat{\theta}_t - \lambda_{i,N_t^i}(\hat{\theta}_t) \Vert_{V_{N_t^i}}
\ge \Vert \theta - \lambda_{i,N_t^i}(\hat{\theta}_t) \Vert_{V_{N_t^i}}
%\ge \Vert \theta - \lambda_{i,N_t^i}(\theta) \Vert_{V_{N_t^i}}
\:,
\end{align*}
and the inequality of Cauchy-Schwarz, we obtain
\begin{align*}
\sum_{i\in\cI}\frac{1}{2}\Vert \hat{\theta}_t - \lambda_{i,N_t^i}(\hat{\theta}_t) \Vert_{V_{N_t^i}}^2
&\ge \sum_{i\in\cI}\frac{1}{2} \left(\Vert \theta - \lambda_{i,N_t^i}(\hat{\theta}_t) \Vert_{V_{{N_t}^i}} -\Vert \hat{\theta}_t -\theta \Vert_{V_{N_t^i}} \right)^2\\
&\ge \sum_{i\in\cI}\frac{1}{2}\Vert \theta - \lambda_{i,N_t^i}(\hat{\theta}_t) \Vert_{V_{N_t^i}}^2
	-  \sum_{i\in\cI} \Vert \hat{\theta}_t -\theta \Vert_{V_{N_t^i}} \Vert \theta - \lambda_{i,N_t^i}(\hat{\theta}_t) \Vert_{V_{N_t^i}}\\
&\ge \sum_{i\in\cI}\frac{1}{2}\Vert \theta - \lambda_{i,N_t^i}(\hat{\theta}_t) \Vert_{V_{N_t^i}}^2
	-  \sqrt{\sum_{i\in\cI} \Vert \hat{\theta}_t -\theta \Vert_{V_{N_t^i}}^2}\sqrt{\sum_{i\in\cI} \Vert \theta - \lambda_{i,N_t^i}(\hat{\theta}_t) \Vert_{V_{N_t^i}}^2}
\,.
\end{align*}

Using again that $\frac{1}{2}\Vert \hat{\theta}_t -\theta \Vert_{V_{N_t}}^2 \leq h(t)$ on $\cE_t$, we get
\begin{align*}
\beta(t,\delta) &\ge \sum_{i\in\cI}\frac{1}{2} \Vert \theta - \lambda_{i,N_t^i}(\hat{\theta}_t) \Vert_{V_{N_t^i}}^2 - \sqrt{4 h(t) \sum_{i\in\cI}\frac{1}{2}\Vert \theta - \lambda_{i,N_t^i}(\hat{\theta}_t) \Vert_{V_{N_t^i}}^2}\,,
\end{align*}
which leads to, using Lemma~\ref{lem:inq_revert_sqrt},
\begin{equation*}
%\label{eq:to_theta_cvx}
\beta(t,\delta) + \sqrt{4 h(t) \beta(t,\delta)} +4h(t) \geq \frac{1}{2}\sum_{i\in\cI}  \Vert \theta - \lambda_{i,N_t^i}(\hat{\theta}_t) \Vert_{V_{N_t^i}}^2 \,.
\end{equation*}
\end{proof}

We now continue the proof by replacing the counts by the weights in~\eqref{eq:to_theta_cvx}.
\begin{lemma}On $\cE_t$, if the algorithm does not stop at $t$,
	\begin{equation}
	  \label{eq:to_W_cvx}
	  \beta(t,\delta)+5AI\left( \sqrt{h(t)\beta(t,\delta)}+2h(t)\right)
	  \geq  \frac{1}{2} \sum_{i\in\cI}\Vert \theta - \lambda_{i,N_t^i}(\hat{\theta}_t) \Vert_{V_{\tW_t^i}}^2\,.
	\end{equation}
\end{lemma}

\begin{proof}
 Using the tracking property, see Lemma~\ref{lem:tracking}, to state that for all $(a,i)\in\cA\times\cI$, $- \log(AI) \le N_t^{a,i} -\tW_t^{a,i}\le 1$, which implies
\begin{align*}
\frac{1}{2}\sum_{i\in\cI} \Vert \theta - \lambda_{i,N_t^i}(\htheta_t) \Vert_{V_{N_t^i}}^2
&\ge \frac{1}{2}\sum_{i\in\cI} \Vert \theta - \lambda_{i,N_t^i}(\hat{\theta}_t) \Vert_{V_{\tW_t}}^2
	- \frac{\log(AI)}{2} \sum_{(a,i)\in\cA\times\cI} \Vert \theta - \lambda_{i,N_t^i}(\hat{\theta}_t) \Vert_{a a^\top}^2
\\
&\ge \frac{1}{2}\sum_{i\in\cI} \Vert \theta - \lambda_{i,N_t^i}(\hat{\theta}_t) \Vert_{V_{\tW_t^i}}^2
	- \frac{\log(AI)}{2} \sqrt{\sum_{(a,i)\in\cA\times\cI} N_t^{a,i}\Vert \theta - \lambda_{i_t,N_t}(\hat{\theta}_t) \Vert_{a a^\top}^2\sum_{(a,i):N_t^{a,i}\ge 1} \frac{1}{N_t^a}   }
\\
% &=   \sum_{i\in\cI}\frac{1}{2} \Vert \theta - \lambda_{i,N_t^i}(\hat{\theta}_t) \Vert_{V_{W_t^i}}^2
% 	- \frac{\log(AI)}{2} \sqrt{\sum_{i\in\cI}\Vert \theta - \lambda_{i,N_t^i}(\hat{\theta}_t) \Vert_{V_{N_t^i}}^2\sum_{a:N_t^a\ge 1} \frac{1}{N_t^a}   }
% \\
&\ge \frac{1}{2}\sum_{i\in\cI} \Vert \theta - \lambda_{i,N_t^i}(\hat{\theta}_t) \Vert_{V_{\tW_t}}^2
	-\frac{\log(AI)}{2} \sqrt{ \sum_{i\in\cI} \Vert \theta - \lambda_{i,N_t^i}(\hat{\theta}_t) \Vert_{V_{N_t^i}}^2  AI}\,.
\end{align*}
% \todo[inline]{TODO: it seems that we introduce a $\sqrt{t}$ factor that should not be there when the $N$ are introduced and then lower bounded by 1.}
Combining the last inequality with \eqref{eq:to_theta_cvx} yields
\[
\beta(t,\delta) + \sqrt{4 h(t) \beta(t,\delta)} +4h(t)+\frac{\log(AI)}{2} \sqrt{2 AI}\sqrt{\beta(t,\delta) + \sqrt{4 h(t) \beta(t,\delta)} +4h(t)}  \geq \frac{1}{2} \sum_{i\in\cI}\Vert \theta - \lambda_{i,N_t^i}(\hat{\theta}_t) \Vert_{V_{\tW_t^i}}^2\,.
\]
Some simplifications, using the fact that $h(t)\geq 1 $ and $\beta(t,\delta)\geq 1$ (thanks to the choice of $\eta$), give us
\begin{equation*}
  %\label{eq:to_W_cvx}
  \beta(t,\delta)+5AI\left( \sqrt{h(t)\beta(t,\delta)}+2h(t)\right)
  \geq  \frac{1}{2} \sum_{i\in\cI}\Vert \theta - \lambda_{i,N_t^i}(\hat{\theta}_t) \Vert_{V_{\tW_t^i}}^2\,.
\end{equation*}
\end{proof}

We now transit from $\theta$ to each $\hat{\theta}_s$ for $s \le t$ in the right hand term of~\eqref{eq:to_W_cvx}.
\begin{lemma}
On $\cE_t$, if the algorithm does not stop at $t$,
\begin{align}
\label{eq:to_hteta_cvx}
\beta(t,\delta) + 30 AI \left(h(t)\sqrt{ \beta(t,\delta)}+ 2h(t)^2\right)   \geq
\frac{1}{2}\sum_{i\in\cI}\sum_{s=1}^t \Vert \hat{\theta}_{s-1} - \lambda_{i,N_t^i}(\hat{\theta}_t)\Vert_{V_{\tw_s^i}}^2\,.
\end{align}
\end{lemma}
\begin{proof}
Thanks to the inequality of Cauchy-Schwarz, we have,
\begin{align}
\frac{1}{2} \sum_{i\in\cI}\Vert \theta - \lambda_{i,N_t^i}(\hat{\theta}_t) \Vert_{V_{\tW_t^i}}^2
&=   \frac{1}{2}\sum_{i\in\cI} \sum_{s=1}^t  \Vert \theta - \lambda_{i,N_t^i}(\hat{\theta}_t) \Vert_{V_{\tw^i_s}}^2
\nonumber\\
&\ge \frac{1}{2} \sum_{i\in\cI} \sum_{s=1}^t \left(\Vert \hat{\theta}_{s-1} - \lambda_{i,N_t^i}(\hat{\theta}_t)\Vert_{V_{\tw^i_s}}^2 - 2\Vert \theta - \hat{\theta}_{s-1} \Vert_{V_{\tw^i_s}} \Vert \hat{\theta}_{s-1} - \lambda_{i,N_t^i}(\hat{\theta}_t) \Vert_{V_{\tw^i_s}}\right)
\nonumber\\
&\ge \frac{1}{2} \sum_{i\in\cI}\sum_{s=1}^t \Vert \hat{\theta}_{s-1} - \lambda_{i,N_t^i}(\htheta_t) \Vert_{V_{\tw^i_s}}^2
\nonumber\\
&\ - \sqrt{ \sum_{i\in\cI}\sum_{s=1}^t \Vert \theta - \htheta_{s-1} \Vert_{V_{\tw^i_s}}^2  \sum_{i\in\cI}\sum_{s=1}^t\Vert \hat{\theta}_{s-1} -\lambda_{i,N_t^i}(\hat{\theta}_t) \Vert_{V_{\tw_s^i}}^2}
\: .\label{eq:tempW_cvx}
\end{align}
We need to upper bound the quantity $ \sum_{i\in\cI}\sum_{s=1}^t \Vert \theta - \htheta_{s-1} \Vert_{\tw_s^i}^2$. By definition of the event $\cE_t$ we have
\begin{align*}
  \Vert \theta - \hat{\theta}_{s-1} \Vert_{a a^\top}^2 &=  \langle \theta - \hat{\theta}_{s-1} ,a\rangle^2\\
  &\leq\normm{ \theta - \hat{\theta}_{s-1}}_{V_{N_{s-1}}+\eta I_d}^2 \normm{ a}_{(V_{N_{s-1}}+\eta I_d)^{-1}}^2\\
  &\leq 2 h(t) \normm{ a }_{(V_{N_{s-1}}+\eta I_d)^{-1}}^2\,.
\end{align*}
Thus thanks to Lemma~\ref{lem:computation_sum_w_a_N}, we get
\begin{align*}
  \sum_{i\in\cI}\sum_{s=1}^t \Vert \theta - \htheta_{s-1} \Vert_{\tw_s^i}^2 = \sum_{s=1}^t \sum_{(a,i)\in\cA\times\cI} \tw_s^{a,i}\Vert \theta - \hat{\theta}_{s-1} \Vert_{a a^\top}^2 \leq 2h(t)\sum_{s=1}^t \sum_{a\in\cA} \tw_s^{a} \normm{ a }_{(V_{N_{s-1}}+\eta I_d)^{-1}}^2 \leq 4h(t)^2\,.
\end{align*}
% But thanks to the tracking we know that $N_{s-1}^a \geq W_{s-1}^a -\log(A)$. Thus, in combination with the choice of $\eta$  we can exchange counts and weights
% \begin{align*}
%   V_{N_{s-1}} + \eta I_d \geq V_{W_{s}} - V_{w_s} - \log(A) V_{\bOne_A} +\eta I_d \geq  V_{W_{s}} - (\log(A)+1) V_{\bOne_A} +\eta I_d \geq V_{W_{s}}+ \frac{\eta}{2} I_d\,.
% \end{align*}
% Hence we obtain
% \[
% \Vert \theta - \hat{\theta}_{s-1} \Vert_{a a^\top}^2 \leq 2 h(t) \normm{ a }_{(V_{W_{s}}+(\eta/2) I_d)^{-1}}^2\,,
% \]
% and applying Lemma~\ref{lem:sum_w_norm_a} we get
% \[
% \sum_{s=1}^t \sum_{a\in\cA} w_s^{a}\Vert \theta - \hat{\theta}_{s-1} \Vert_{a a^\top}^2 \leq 2 h(t) g(t,\eta/2)\,.
%\]
Now going back to \eqref{eq:tempW_cvx} in combination with \eqref{eq:to_W_cvx} and Lemma~\ref{lem:inq_revert_sqrt}, it follows
\begin{align*}
%\label{eq:to_hteta_cvx}
\beta(t,\delta) + 30 AI \left(h(t)\sqrt{ \beta(t,\delta)}+ 2h(t)^2\right)   \geq
\frac{1}{2}\sum_{i\in\cI}\sum_{s=1}^t \Vert \hat{\theta}_{s-1} - \lambda_{i,N_t^i}(\hat{\theta}_t)\Vert_{V_{\tw_s^i}}^2\,.
\end{align*}
\end{proof}
% TODO: prove that there exists a quantity $C'_t$ of adequate size such that $\sum_s \sum_{i,a} w_s^{i,a}\Vert \theta - \hat{\theta}_{s-1} \Vert_{a a^\top}^2 \le C'_t$. With that fact, we get
% \begin{align*}
% &\frac{1}{2} \sum_s \sum_{i,a} w_s^{i,a}\Vert \theta - \lambda_{i,N_t^i}(\theta) \Vert_{a a^\top}^2
% \\
% &\ge \frac{1}{2}\sum_s \sum_{i,a} w_s^{i,a}\Vert \hat{\theta}_{s-1} - \lambda_{i,N_t^{i}}(\theta)\Vert_{a a^\top}^2
% - \sqrt{C'_t \sum_s \sum_{i,a} w_s^{i,a}\Vert \hat{\theta}_{s-1} - \lambda_{i,N_t^{i}}(\theta)\Vert_{a a^\top}^2}
% \end{align*}
%
% Let $f(x) = \frac{1}{2}x - \sqrt{Cx}$. $f'(x) = \frac{1}{2} - \frac{\sqrt{C}}{2\sqrt{x}}$. this is greater than 0 iff $x \ge C$.
%
% Either $\sum_s \sum_{i,a} w_s^{i,a}\Vert \hat{\theta}_{s-1} - \lambda_{i,N_t^{i}}(\theta)\Vert_{a a^\top}^2 \le C'_t$, or any lower bound for that quantity gives us a lower bound on the value of $f$ at that point.
We now introduce the upper confidence bounds.
\begin{lemma}On $\cE_t$, if the algorithm does not stop at $t$,
	\begin{equation}
	\beta(t,\delta) + 50 AI \left(h(t)\sqrt{\beta(t,\delta)}+ 2h(t)^2\right)  \geq \frac{1}{2}\sum_{s=1}^t\sum_{(a,i)\in\cA\times\cI} \tw_s^{a,i}U_s^{a,i}\,.\label{eq:to_UCB_cvx}
	\end{equation}
	\label{lem:penultimate_lemma_sample_complexity_cvx}
\end{lemma}
\begin{proof}
By definition of the best response $\tlambda_s^i = \argmin_{\lambda \in \neg i} \Vert \hat{\theta}_{s-1} - \lambda \Vert_{\tw_s^i}^2$, we have
\begin{align}
\frac{1}{2}\sum_{i\in\cI} \sum_{s=1}^t  \Vert \hat{\theta}_{s-1} - \lambda_{i,N_t^i}(\hat{\theta}_t)\Vert_{V_{\tw_s^i}}^2
&\ge \frac{1}{2}\sum_{i\in\cI} \inf_{\lambda_i \in \neg i} \sum_{s=1}^t \Vert \hat{\theta}_{s-1} - \lambda_i\Vert_{V_{\tw_s^i}}^2
\nonumber\\
&\geq \frac{1}{2} \sum_{s=1}^t \sum_{(a,i)\in\cA\times\cI} \tw_s^{i,a}\Vert \hat{\theta}_{s-1} - \tlambda_s^i\Vert_{a a^\top}^2\,.\label{eq:to_br_cvx}
\end{align}

We recall the upper confidence bounds~\eqref{eq:def_ucb_br_general},
\[U_s^{i,a} = \min\left(\max_\pm \left( \langle \htheta_{s-1} - \tlambda^{i}_s), a\rangle \pm \sqrt{2 h(t)} \norm{a}_{(V_{N_{s-1}}+\eta I_d)^{-1}} \right)^2,4L^2M^2\right)\,,\]

% \begin{align*}
% \sum_{s=1}^t \sum_{(a,i)\in\cA\times\cI} w_s^{i,a} \Vert \hat{\theta}_{s-1} - \tlambda_s^i\Vert_{a a^\top}^2
% &= \sum_{s=1}^t \sum_{(a,i)\in\cA\times\cI} w_s^{i,a} U_s^{i,a} - w_s^{i,a}\left(\Vert \hat{\theta}_{s-1} - \tlambda_s^i \Vert_{a a^\top}^2 - U_s^{i,a}\right)\,.
% \end{align*}
% We deal with the last term in the sum,

it then holds
\begin{align*}
U_s^{i,a} -\Vert \htheta_{s-1} - \tlambda_s^i\Vert_{a a^\top}^2
&\leq   \max_\pm \left( \langle \htheta_{s-1} - \tlambda^i_s,a\rangle \pm \sqrt{2 h(t)} \normm{a}_{(V_{N_{s-1}}+ \eta I_d)^{-1}} \right)^2-\Vert \htheta_{s-1} - \tlambda_s^i\Vert_{a a^\top}^2\\
&\leq  2h(t)\normm{a}^2_{(V_{N_{s-1}}+ \eta I_d)^{-1}} + 2\sqrt{2 h(t)} \normm{a}_{(V_{N_{s-1}}+ \eta I_d)^{-1}}| \langle \htheta_{s-1} - \tlambda_s^i,a\rangle|\,.\\
\end{align*}
Hence summing over $t$ and using the inequality of Cauchy-Schwarz, we obtain
\begin{align*}
&\frac{1}{2}\sum_{s=1}^t\sum_{(a,i)\in\cA\times\cI} \tw_s^{a,i}\left(U_s^{i,a} -\Vert \hat{\theta}_{s-1} - \tlambda_s^i\Vert_{a a^\top}^2\right)
\\
&\le \sum_{s=1}^t \sum_{(a,i)\in\cA\times\cI} \tw_s^{a,i}h(t)\normm{a}^2_{(V_{N_{s-1}}+ \eta I_d)^{-1}} +\tw_s^{a,i} \sqrt{2 h(t)} \normm{a}_{(V_{N_{s-1}}+ \eta I_d)^{-1}}| \langle \htheta_{s-1} - \tlambda_s^i,a\rangle|
\\
&\le  h(t) \sum_{(a,i)\in\cA\times\cI} \sum_{s=1}^t \tw_s^{a,i}\normm{a}^2_{(V_{N_{s-1}}+ \eta I_d)^{-1}}\\
&+ \sqrt{2 h(t)} \sqrt{\sum_{(a,i)\in\cA\times\cI} \sum_{s=1}^t \tw_s^{a,i}\normm{a}^2_{(V_{N_{s-1}}+ \eta I_d)^{-1}}}  \sqrt{\sum_{s=1}^t \sum_{(a,i)\in\cA\times\cI} \tw_s^{a,i}\Vert \htheta_{s-1} - \tlambda_s^i \Vert_{a a^\top}^2} \\
&\leq 2h(t)^2 +2\sqrt{2}h(t)  \sqrt{\frac{1}{2}\sum_{s=1}^t \sum_{(a,i)\in\cA\times\cI} \tw_s^{a,i}\Vert \htheta_{s-1} - \tlambda_s^i \Vert_{a a^\top}^2}\,,
% &\le 2 C_t \sum_s \sum_{i,a} w_s^{i,a} a^\top V_{N_s^i}^{-1}a + 2\sqrt{2 C_t}\sqrt{\sum_s \sum_{i,a} w_s^{i,a}a^\top V_{N_s^i}^{-1}a }\sqrt{\sum_s \sum_{i,a} w_s^{i,a}\Vert \hat{\theta}_{s-1} - \lambda_{s,i}\Vert_{a a^\top}^2}
\end{align*}
where the last inequality is derived from Lemma~\ref{lem:computation_sum_w_a_N}. Thus combining the previous inequality with \eqref{eq:to_hteta_cvx} and \eqref{eq:to_br_cvx} with some simplifications leads to
\begin{equation*}
\beta(t,\delta) + 50 AI \left(h(t)\sqrt{\beta(t,\delta)}+ 2h(t)^2\right)  \geq \frac{1}{2}\sum_{s=1}^t\sum_{(a,i)\in\cA\times\cI} \tw_s^{a,i}U_s^{a,i}\,.%\label{eq:to_UCB_cvx}
\end{equation*}
\end{proof}

%  we have
% \[
%  \sum_{s=1}^t\sum_{(a,i)\in\cA\times\cI} w_s^{i,a}\Vert \hat{\theta}_{s-1} - \lambda_s^i\Vert_{a a^\top}^2 \geq \sum_{s=1}^t\sum_{(a,i)\in\cA\times\cI} w_s^{i,a}U_s^{i,a} + 2h(t)g(t)+2\sqrt{g(t)h(t)}  \sqrt{\sum_{s=1}^t \sum_{(a,i)\in\cA\times\cI} w_s^{i,a}\Vert \htheta_{s-1} - \tlambda_s^i \Vert_{a a^\top}^2}
% \]
% Let $C''_t$ be an upper bound for $\sum_s \sum_{i,a} w_s^{i,a} a^\top V_{N_s^i}^{-1}a$ . Then
% \begin{align*}
% \sum_s \sum_{i,a} w_s^{i,a}\Vert \hat{\theta}_{s-1} - \lambda_{s,i}\Vert_{a a^\top}^2
% &\ge \sum_s \sum_{i,a} w_s^{i,a}U_s^{i,a}
% - 2 C_t C''_t - 2\sqrt{2 C_t C''_t }\sqrt{\sum_s \sum_{i,a} w_s^{i,a}\Vert \hat{\theta}_{s-1} - \lambda_{s,i}\Vert_{a a^\top}^2}
% \end{align*}
% That is, $x=\sum_s \sum_{i,a} w_s^{i,a}\Vert \hat{\theta}_{s-1} - \lambda_{s,i}\Vert_{a a^\top}^2$ verifies the inequality $(\sqrt{x} + \sqrt{2C_tC''_t})^2 \ge \sum_{i,a} w_s^{i,a}U_s^{i,a}$ . We get $x \ge (\sqrt{\sum_{i,a} w_s^{i,a}U_s^{i,a}} - \sqrt{2C_tC''_t})^2$ .
%
It remains to bound the regret of the learner.
\begin{lemma}
	On $\cE_t$, if the algorithm does not stop at $t$,
\begin{equation}
	\label{eq:to_t_1}
	\beta(t,\delta) + 50 AI \left(h(t)\sqrt{\beta(t,\delta)}+ 2h(t)^2\right) +C_1\sqrt{t} +C_2  \geq t \Tstar(\theta)^{-1}\,,
\end{equation}
where
\[
C_1 = 4L^2M^2\sqrt{\log(AI)}\qquad \qquad C_2 = 64L^2M^2\left( 2+\log(AI)\right) \,.
\]
\end{lemma}
\begin{proof}
Thanks to Proposition~\ref{lem:adahedge} for the algorithm AdaHedge, we have the following regret bound for the $\cL_{\tw}$-learner
\[
\max_{\tw\in \Sigma_{AI}}\frac{1}{2}\sum_{s=1}^t\sum_{(a,i)\in\cA\times\cI} \tw^{a,i}U_s^{a,i}- \frac{1}{2}\sum_{s=1}^t\sum_{(a,i)\in\cA\times\cI} \tw_s^{a,i}U_s^{a,i}\leq \underbrace{4L^2M^2\sqrt{\log(AI)}}_{\eqdef C_1}\sqrt{t}+\underbrace{64L^2M^2\left( 2+\log(AI)\right)}_{\eqdef C_2}\,.
\]
Finally, using this inequality in combination with \eqref{eq:to_UCB_cvx} and the fact that the gains are optimistic (see Lemma~\ref{lem:confidence_bound_general}), we obtain
\begin{align*}
  \beta(t,\delta) + 50 AI \left(h(t)\sqrt{\beta(t,\delta)}+ 2h(t)^2\right) +C_1\sqrt{t} +C_2 &\geq \sup_{\tw\in\Sigma_{AI}}\frac{1}{2}\sum_{s=1}^t\sum_{(a,i)\in\cA\times\cI} \tw^{a,i}U_s^{a,i}\\
  &\geq \sup_{\tw\in\Sigma_{AI}}\frac{1}{2}\sum_{s=1}^t\sum_{(a,i)\in\cA\times\cI} \tw^{a,i} \normm{\theta -\tlambda_s^{i}}^2_{aa^\top}\\
  &= t \sup_{w\in\Sigma_ {AI}}\sum_{i\in\cI}\frac{1}{t}\sum_{s=1}^t\frac{1}{2}\normm{\theta -\tlambda_s^{i}}^2_{V_{\tw^i}}\\
  &\geq t \sup_{\tw\in\Sigma_{AI}}\sum_{i\in\cI} \inf_{q^i\in\cP(\neg i)} \E_{\lambda\sim q^i}\frac{1}{2}\normm{\theta -\lambda}^2_{V_{\tw^i}}\\
 &= t \Tstar(\theta)^{-1}\,,
\end{align*}
where for the last line we use the Sion's minimax theorem, see Lemma~\ref{lem:sion_convexify}.
\end{proof}

Now let $T(\delta)$ be the maximum of the $t\in\NN$ such that
\begin{equation}
	\label{eq:def_T_delta}
\beta(t,\delta) + 50 AI \left(h(t)\sqrt{\beta(t,\delta)}+ 2h(t)^2\right) +C_1\sqrt{t} +C_2  \geq  t \Tstar(\theta)^{-1}\,.
\end{equation}

And let $\delta_{\min}$ be the largest $\delta\in(0,1)$ such that
{\small
\begin{equation}
\label{eq:def_delta_min}
\beta(\log(1/\delta)^{3/2},\delta) + 50 AI \left(h\big(\log(1/\delta)^{3/2}\big)\sqrt{\beta\big(\log(1/\delta)^{3/2},\delta\big)}+ 2h\big(\log(1/\delta)^{3/2}\big)^2\right) +C_1\log(1/\delta)^{3/4} +C_2  < \log(1/\delta)^{3/2} \Tstar(\theta)^{-1}\,.
\end{equation}
}%
Such $\delta$ always exists since
{\small
\[
\beta(\log(1/\delta)^{3/2},\delta) + 50 AI \left(h\big(\log(1/\delta)^{3/2}\big)\sqrt{\beta\big(\log(1/\delta)^{3/2},\delta\big)}+ 2h\big(\log(1/\delta)^{3/2}\big)^2\right) +C_1\log(1/\delta)^{3/4} +C_2 = o\big(\log(1/\delta)\big)
\]
}%
and it depends only on the parameter of the problem: $\Tstar(\theta),L,M,\alpha,\eta, A,I$. Hence for $\delta\leq \delta_{\min}$ we know that $T(\delta) \leq \log(1/\delta)^{3/2}$, which implies
{\small
\begin{align}
T(\delta)&< T_0(\delta)\nonumber\\
&\eqdef\Tstar(\theta) \left(\beta(\log(1/\delta)^{3/2},\delta) + 50 AI \left(h\big(\log(1/\delta)^{3/2}\big)\sqrt{\beta\big(\log(1/\delta)^{3/2},\delta\big)}+ 2h\big(\log(1/\delta)^{3/2}\big)^2\right) +C_1\log(1/\delta)^{3/4} +C_2+1\right)\,.\label{eq:def_T_0}
\end{align}
}%
Actually, if we do not stop on $\cE_t$, it implies that $t\leq T(\delta)$ thanks to Lemma~\ref{lem:penultimate_lemma_sample_complexity_cvx}. Thus for $t\geq T_0(\delta)> T(\delta)$ we know that $\tau_\delta \leq t$. We just proved the following lemma.
\begin{lemma}
\label{lem:T_delta_cvx}
If $\delta\leq \delta_{\min}$, defined in \eqref{eq:def_delta_min} then for $t\geq T_0(\delta)$ where $T_0(\delta)$ is defined in~\eqref{eq:def_T_0}\,, $\cE_t \subset \{\tau_\delta \leq t\}$\,.
\end{lemma}

\subsection{Proof of Theorem~\ref{thm:sample_complexity}}
Combining Lemma~\ref{lem:start_analysis} and Lemma~\ref{lem:T_delta_cvx} for \algoWCvx or Lemma~\ref{lem:T_delta_nc} in Appendix~\ref{app:analysis_under_concentration_nc} for \algoW, we have for $\delta\leq \delta_{\min}$,
\begin{equation}
	\label{eq:tmp_prf_th}
\E_{\theta}[\tau_\delta] \leq T_0(\delta)+1+\frac{2^{\alpha-2}}{\alpha-2}\,,
\end{equation}
with $ T_0(\delta) = \Tstar(\theta) \log(1/\delta) + o(\log(1/\delta))$, see \eqref{eq:def_T_0} for an exact expression for \algoWCvx (or equation~\eqref{eq:def_T_0_nc} for \algoW). That implies the asymptotic optimality of \algoWCvx and \algoW,
\[
\limsup_{\delta\to 0}\frac{\E_{\theta}[\tau_\delta]}{\log(1/\delta)} = \Tstar(\theta)\,.
\]
It remains to prove the $\delta$-correctness. Thanks to the inequality~\eqref{eq:tmp_prf_th} above, we have\footnote{If $\delta\leq \delta_{\min}$ the inequality holds replacing $T_0(\delta)$ by $T(\delta)+1$ defined in~\eqref{eq:def_T_delta} which is also finite.} that $\tau_\delta <+\infty$ almost surely. And using
Lemma~\ref{lem:chernoff_stopping rule_pac} we know that $\P_\theta\big(\hi \neq \istar(\theta) \big)\leq \delta$, which allows us to conclude.

\subsection{Analysis under concentration of \algoW}
\label{app:analysis_under_concentration_nc}

In this section we assume that the event $\cE_t$ holds and set $\istar=\istar(\theta)$. We prove that $\beta(t,\delta)  \gtrsim N_t^{\istar} \Tstar(\theta)^{-1}$ and that $N_t^i = O(\sqrt{t})$ , for all $i\neq \istar$. We define the weights $w_t^i = \ind_{\{i_t = i\}}w_t$ and the best response $\lambda_s^t = \argmin_{\lambda \in \neg i} \Vert \hat{\theta}_{t-1} - \lambda \Vert_{w_t}^2$.
 Note that our algorithm computes $\lambda_t^{i}$ only when $i_t = i$.

\paragraph{When $i_s = \istar$.} Here we proceed as in Appendix~\ref{app:analysis_under_concentration_cvx}. If the algorithm does not stop at stage $t$ we have
\begin{align*}
\beta(t,\delta)
\ge \frac{1}{2}\max_{i\in\cI} \inf_{\lambda_i \in \neg i}\Vert \hat{\theta}_t - \lambda_i \Vert_{V_{N_t}}^2
\geq  \frac{1}{2} \inf_{\lambda \in \neg \istar}\Vert \hat{\theta}_t - \lambda \Vert_{V_{N_t}}^2\,.
%\ge \frac{1}{2}\sum_{i\in\cI} \inf_{\lambda_i \in \neg i}\Vert \hat{\theta}_t - \lambda_i \Vert_{V_{N_t^i}}^2
\end{align*}
Hence we need to find a lower bound for the right hand sum. Let $\lambda_{i,w}(\theta) \in \argmin_{\lambda \in \neg i} \Vert \theta - \lambda \Vert_{V_{w}}$. We first replace in this sum $\htheta_t$ by $\theta$.
\begin{lemma}On $\cE_t$, if the algorithm does not stop at $t$,
\begin{equation}
	\label{eq:to_theta_nc}
	\beta(t,\delta) + \sqrt{4 h(t) \beta(t,\delta)} +4h(t) \geq \frac{1}{2}\Vert \theta - \lambda_{\istar,N_t}(\hat{\theta}_t) \Vert_{V_{N_t}}^2 \,.
\end{equation}
\end{lemma}

\begin{proof}
	Using the triangular inequality,
	\begin{align*}
	\Vert \theta - \hat{\theta}_t \Vert_{V_{N_t}} + \Vert \hat{\theta}_t - \lambda_{\istar,N_t}(\htheta_t) \Vert_{V_{N_t}}
	\ge \Vert \theta - \lambda_{\istar,N_t}(\hat{\theta}_t) \Vert_{V_{N_t}}\,,
	\end{align*}
	we obtain
	\begin{align*}
	\frac{1}{2}\Vert \hat{\theta}_t - \lambda_{\istar,N_t}(\hat{\theta}_t) \Vert_{V_{N_t}}^2
	&\ge \frac{1}{2} \left(\Vert \theta - \lambda_{\istar,N_t}(\hat{\theta}_t) \Vert_{V_{N_t}} -\Vert \hat{\theta}_t -\theta \Vert_{V_{N_t}} \right)^2
	\\
	&\ge \frac{1}{2}\Vert \theta - \lambda_{\istar,N_t}(\hat{\theta}_t) \Vert_{V_{N_t}}^2
		-  \Vert \hat{\theta}_t -\theta \Vert_{V_{N_t}} \Vert \theta - \lambda_{\istar,N_t}(\hat{\theta}_t) \Vert_{V_{N_t}}
	\,.
	\end{align*}
	By definition of the event $\cE_t$ we know that $\frac{1}{2}\Vert \hat{\theta}_t -\theta \Vert_{V_{N_t}}^2 \leq h(t)$. Thus we get
	\begin{align*}
	\frac{1}{2}\Vert \hat{\theta}_t - \lambda_{\istar,N_t}(\hat{\theta}_t) \Vert_{V_{N_t}}^2 &\ge \frac{1}{2} \Vert \theta - \lambda_{\istar,N_t}(\hat{\theta}_t) \Vert_{V_{N_t}}^2 - \sqrt{4 h(t) \frac{1}{2}\Vert \theta - \lambda_{\istar,N_t}(\hat{\theta}_t) \Vert_{V_{N_t}}^2}\,,
	\end{align*}
	which leads to, using Lemma~\ref{lem:inq_revert_sqrt},
	\begin{equation*}
	\beta(t,\delta) + \sqrt{4 h(t) \beta(t,\delta)} +4h(t) \geq \frac{1}{2} \Vert \theta - \lambda_{\istar,N_t}(\hat{\theta}_t) \Vert_{V_{N_t}}^2 \,.
\end{equation*}
\end{proof}

We now continue the proof by replacing the counts by the weights in~\eqref{eq:to_theta_nc}.
\begin{lemma}On $\cE_t$, if the algorithm does not stop at $t$,
	\begin{equation}
	  \label{eq:to_W_nc}
	  \beta(t,\delta)+5A\left( \sqrt{h(t)\beta(t,\delta)}+2h(t)\right)
	  \geq  \frac{1}{2} \Vert \theta - \lambda_{\istar,N_t}(\hat{\theta}_t) \Vert_{V_{W_t}}^2\,.
	\end{equation}
\end{lemma}

\begin{proof}
	Using the tracking property, see Lemma~\ref{lem:tracking}, to state that for all $a$, $- \log(A) \le N_t^{a} - W_t^{a}\le 1$, we get
	\begin{align*}
	\frac{1}{2} \Vert \theta - \lambda_{\istar,N_t}(\hat{\theta}_t) \Vert_{V_{N_t}}^2
	&\ge \frac{1}{2} \Vert \theta - \lambda_{\istar,N_t}(\hat{\theta}_t) \Vert_{V_{W_t}}^2
		- \frac{\log(A)}{2} \sum_{a\in\cA} \Vert \theta - \lambda_{\istar,N_t}(\hat{\theta}_t) \Vert_{a a^\top}^2
	\\
	&\ge \frac{1}{2} \Vert \theta - \lambda_{\istar,N_t}(\hat{\theta}_t) \Vert_{V_{W_t}}^2
		- \frac{\log(A)}{2} \sqrt{\sum_{a\in\cA} N_t^{a}\Vert \theta - \lambda_{\istar,N_t}(\hat{\theta}_t) \Vert_{a a^\top}^2\sum_{a:N_t^a\ge 1} \frac{1}{N_t^a}   }
	\\
	&=   \frac{1}{2} \Vert \theta - \lambda_{\istar,N_t}(\hat{\theta}_t) \Vert_{V_{W_t}}^2
		- \frac{\log(A)}{2} \sqrt{\Vert \theta - \lambda_{\istar,N_t}(\hat{\theta}_t) \Vert_{V_{N_t}}^2\sum_{a:N_t^a\ge 1} \frac{1}{N_t^a}   }
	\\
	&\ge \frac{1}{2} \Vert \theta - \lambda_{\istar,N_t}(\hat{\theta}_t) \Vert_{V_{W_t}}^2
		-\frac{\log(A)}{2} \sqrt{ A \Vert \theta - \lambda_{\istar,N_t}(\hat{\theta}_t) \Vert_{V_{N_t}}^2  }\,.
	\end{align*}
	Combining the last inequality with \eqref{eq:to_theta_nc} yields
	\[
	\beta(t,\delta) + \sqrt{4 h(t) \beta(t,\delta)} +4h(t)+\frac{\log(A)}{2} \sqrt{2 A}\sqrt{\beta(t,\delta) + \sqrt{4 h(t) \beta(t,\delta)} +4h(t)}  \geq \frac{1}{2} \Vert \theta - \lambda_{\istar,N_t}(\hat{\theta}_t) \Vert_{V_{W_t}}^2\,.
	\]
	Some simplifications, using the fact that $\beta(t,\delta)\geq 1 $ and  $h(t)\geq 1 $, give us
	\begin{equation*}
	  %\label{eq:to_tW_nc}
	  \beta(t,\delta)+5A\left( \sqrt{h(t)\beta(t,\delta)}+2h(t)\right)
	  \geq  \frac{1}{2} \Vert \theta - \lambda_{\istar,N_t}(\hat{\theta}_t) \Vert_{V_{W_t}}^2\,.
	\end{equation*}
\end{proof}

We now go from $\theta$ to each $\hat{\theta}_s$ for $s \le t$ in the right hand term of~\eqref{eq:to_W_nc}.
\begin{lemma}
On $\cE_t$, if the algorithm does not stop at $t$,
\begin{align}
\label{eq:to_hteta_nc}
\beta(t,\delta) + 30 A \left(h(t)\sqrt{ \beta(t,\delta)}+ 2h(t)^2\right)   \geq
\frac{1}{2}\sum_{s=1}^t \Vert \hat{\theta}_{s-1} - \lambda_{\istar,N_t}(\hat{\theta}_t)\Vert_{V_{w_s}}^2\,.
\end{align}
\end{lemma}
\begin{proof}
    Using the inequality of Cauchy-Schwarz, we have,
	\begin{align}
	\frac{1}{2}\Vert \theta - \lambda_{\istar,N_t}(\hat{\theta}_t) \Vert_{V_{W_t}}^2
	&=   \frac{1}{2} \sum_{s=1}^t  \Vert \theta - \lambda_{\istar,N_t}(\hat{\theta}_t) \Vert_{V_{w_s}}^2
	\nonumber\\
	&\ge \frac{1}{2} \sum_{s=1}^t \left(\Vert \hat{\theta}_{s-1} - \lambda_{\istar,N_t}(\hat{\theta}_t)\Vert_{V_{w_s}}^2 - 2\Vert \theta - \hat{\theta}_{s-1} \Vert_{V_{w_s}} \Vert \hat{\theta}_{s-1} - \lambda_{\istar,N_t}(\hat{\theta}_t) \Vert_{V_{w_s}}\right)
	\nonumber\\
	&\ge \frac{1}{2} \sum_{s=1}^t \Vert \hat{\theta}_{s-1} - \lambda_{\istar,N_t}(\hat{\theta}_t) \Vert_{V_{w_s}}^2
	\nonumber\\
	&\ - \sqrt{\sum_{s=1}^t \Vert \theta - \hat{\theta}_{s-1} \Vert_{V_{w_s}}^2 \sum_{s=1}^t\Vert \hat{\theta}_{s-1} -\lambda_{\istar,N_t}(\hat{\theta}_t) \Vert_{V_{w_s}}^2}
	\: .\label{eq:tempW_nc}
	\end{align}
	We need to upper bound the quantity $\sum_{s=1}^t  w_s^{a}\Vert \theta - \hat{\theta}_{s-1} \Vert_{a a^\top}^2$. By definition of the event $\cE_t$ we have
	\begin{align*}
	  \Vert \theta - \hat{\theta}_{s-1} \Vert_{a a^\top}^2 &=  \langle \theta - \hat{\theta}_{s-1} ,a\rangle^2\\
	  &\leq\normm{ \theta - \hat{\theta}_{s-1}}_{V_{N_{s-1}}+\eta I_d}^2 \normm{ a}_{(V_{N_{s-1}}+\eta I_d)^{-1}}^2\\
	  &\leq 2 h(t) \normm{ a }_{(V_{N_{s-1}}+\eta I_d)^{-1}}^2\,.
	\end{align*}
	Thus thanks to Lemma~\ref{lem:computation_sum_w_a_N}, we get
	\begin{align}
    \label{eq:sum_dist_theta_nc}
	  \sum_{s=1}^t \sum_{a\in\cA} w_s^{a}\Vert \theta - \hat{\theta}_{s-1} \Vert_{a a^\top}^2 \leq 2h(t)\sum_{s=1}^t \sum_{a\in\cA} w_s^{a} \normm{ a }_{(V_{N_{s-1}}+\eta I_d)^{-1}}^2 \leq 4h(t)^2\,.
	\end{align}
	Now going back to \eqref{eq:tempW_nc} in combination with \eqref{eq:to_W_nc} and Lemma~\ref{lem:inq_revert_sqrt} leads to
	\begin{align}
	%\label{eq:to_hteta_nc}
	\beta(t,\delta) + 30 A\left(h(t)\sqrt{h(t) \beta(t,\delta)}+ 2h(t)^2\right)   \geq
	\frac{1}{2}\sum_{s=1}^t \Vert \hat{\theta}_{s-1} - \lambda_{\istar,N_t}(\hat{\theta}_t)\Vert_{V_{w_s}}^2\,.
	\end{align}

\end{proof}

We now introduce the upper confidence bounds.
\begin{lemma}On $\cE_t$, if the algorithm does not stop at $t$,
	\begin{equation}
	\beta(t,\delta) + 50 A \left(h(t)\sqrt{\beta(t,\delta)}+ 2h(t)^2\right)  \geq \frac{1}{2}\sum_{s=1}^t w_s^{a}U_s^{a,\istar}\,.\label{eq:to_UCB_nc}
	\end{equation}
	\label{lem:penultimate_lemma_sample_complexity_nc}
\end{lemma}

\begin{proof}

	By definition of the best response $\lambda_s^i = \argmin_{\lambda \in \neg i} \Vert \hat{\theta}_{s-1} - \lambda \Vert_{w_s}^2$, we have
	\begin{align}
	\frac{1}{2} \sum_{s=1}^t  \Vert \hat{\theta}_{s-1} - \lambda_{\istar,N_t}(\hat{\theta}_t)\Vert_{V_{w_s}}^2
	&\ge \frac{1}{2} \inf_{\lambda \in \neg \istar} \sum_{s=1}^t \Vert \hat{\theta}_{s-1} - \lambda\Vert_{V_{w_s}}^2
	\nonumber\\
	&\geq\frac{1}{2} \sum_{s=1}^t \sum_{a\in\cA} w_s^{a}\Vert \hat{\theta}_{s-1} - \lambda_s^{\istar}\Vert_{a a^\top}^2\,.\label{eq:to_br_nc}
	\end{align}

	We recall the upper confidence bounds~\eqref{eq:def_ucb_br_general}
	\[U_s^{a,i} = \min\left(\max_\pm \left( \langle \htheta_{s-1} - \lambda^{i}_s), a\rangle \pm \sqrt{2 h(t)} \norm{a}_{(V_{N_{s-1}}+\eta I_d)^{-1}} \right)^2,4L^2M^2\right)\,,\]

	% \begin{align*}
	% \sum_{s=1}^t \sum_{(a,i)\in\cA\times\cI} w_s^{i,a} \Vert \hat{\theta}_{s-1} - \tlambda_s^i\Vert_{a a^\top}^2
	% &= \sum_{s=1}^t \sum_{(a,i)\in\cA\times\cI} w_s^{i,a} U_s^{i,a} - w_s^{i,a}\left(\Vert \hat{\theta}_{s-1} - \tlambda_s^i \Vert_{a a^\top}^2 - U_s^{i,a}\right)\,.
	% \end{align*}
	% We deal with the last term in the sum,

	it then holds
	\begin{align*}
	U_s^{a,i} -\Vert \htheta_{s-1} - \lambda_s^i\Vert_{a a^\top}^2
	&\leq   \max_\pm \left( \langle \htheta_{s-1} - \lambda^i_s,a\rangle \pm \sqrt{2 h(t)} \normm{a}_{(V_{N_{s-1}}+ \eta I_d)^{-1}} \right)^2-\Vert \htheta_{s-1} - \lambda_s^i\Vert_{a a^\top}^2\\
	&\leq  2h(t)\normm{a}^2_{(V_{N_{s-1}}+ \eta I_d)^{-1}} + 2\sqrt{2 h(t)} \normm{a}_{(V_{N_{s-1}}+ \eta I_d)^{-1}}| \langle \htheta_{s-1} - \lambda_s^i,a\rangle|\,.\\
	\end{align*}
	Hence summing over $t$ and using the inequality of Cauchy-Schwarz we obtain
	\begin{align*}
	&\frac{1}{2}\sum_{s=1}^t\sum_{a\in\cA} w_s^{a}\left(U_s^{a,\istar} -\Vert \hat{\theta}_{s-1} - \lambda_s^{\istar}\Vert_{a a^\top}^2\right)
	\\
	&\le \sum_{s=1}^t \sum_{a\in\cA} w_s^{a}h(t)\normm{a}^2_{(V_{N_{s-1}}+ \eta I_d)^{-1}} +w_s^{a} \sqrt{2 h(t)} \normm{a}_{(V_{N_{s-1}}+ \eta I_d)^{-1}}| \langle \htheta_{s-1} - \lambda_s^{\istar},a\rangle|
	\\
	&\le  h(t) \sum_{a\in\cA} \sum_{s=1}^t w_s^{a}\normm{a}^2_{(V_{N_{s-1}}+ \eta I_d)^{-1}}\\
	&+ \sqrt{2 h(t)} \sqrt{\sum_{a\in\cA} \sum_{s=1}^t w_s^{a}\normm{a}^2_{(V_{N_{s-1}}+ \eta I_d)^{-1}}}  \sqrt{\sum_{s=1}^t \sum_{a\in\cA} w_s^{a}\Vert \htheta_{s-1} - \lambda_s^{\istar} \Vert_{a a^\top}^2} \\
	&\leq 2h(t)^2 +2\sqrt{2}h(t)  \sqrt{\frac{1}{2}\sum_{s=1}^t \sum_{a\in\cA} w_s^{a}\Vert \htheta_{s-1} - \lambda_s^{\istar} \Vert_{a a^\top}^2}\,,
	% &\le 2 C_t \sum_s \sum_{i,a} w_s^{i,a} a^\top V_{N_s^i}^{-1}a + 2\sqrt{2 C_t}\sqrt{\sum_s \sum_{i,a} w_s^{i,a}a^\top V_{N_s^i}^{-1}a }\sqrt{\sum_s \sum_{i,a} w_s^{i,a}\Vert \hat{\theta}_{s-1} - \lambda_{s,i}\Vert_{a a^\top}^2}
	\end{align*}
	where the last inequality is derived from Lemma~\ref{lem:computation_sum_w_a_N}. Thus combining the previous inequality with \eqref{eq:to_hteta_nc} and \eqref{eq:to_br_nc} with some simplifications leads to
	\begin{equation*}
	\beta(t,\delta) + 50 A \left(h(t)\sqrt{\beta(t,\delta)}+ 2h(t)^2\right)  \geq \frac{1}{2}\sum_{s=1}^t\sum_{a\in\cA} w_s^{a}U_s^{a,\istar}\,.%\label{eq:to_UCB_cvx}
	\end{equation*}

\end{proof}

We then bound the regret of the learner.
\begin{lemma}
  \label{lem:upper_bound_nstar}
	On $\cE_t$, if the algorithm does not stop at $t$,
\begin{equation}
	\label{eq:to_t_2}
	\beta(t,\delta) + 50 A \left(h(t)\sqrt{\beta(t,\delta)}+ 2h(t)^2\right) +C_1\sqrt{t} +C_2  \geq N^{\istar}(t) \Tstar(\theta)^{-1}\,,
\end{equation}
where
\[
C_1 = 4L^2M^2\sqrt{\log(A)}\qquad \text{and} \qquad C_2 = 64L^2M^2\left( 2+\log(A)\right)\,.
\]
\end{lemma}
\begin{proof}
Thanks to Proposition~\ref{lem:adahedge} for the algorithm AdaHedge we have the following regret bound for the learner $\cL_w^{\istar}$
\[
\max_{w\in \Sigma_{A}}\frac{1}{2}\sum_{s=1}^t \ind_{\{i_s=\istar\}}\sum_{a\in\cA} w^{a}U_s^{a,\istar}- \frac{1}{2}\sum_{s=1}^t\sum_{a\in\cA} w_s^{a,\istar}U_s^{a,\istar}\leq \underbrace{4L^2M^2\sqrt{\log(A)}}_{\eqdef C_1}\sqrt{t}+\underbrace{64L^2M^2\left( 2+\log(A)\right)}_{\eqdef C_2}\,.
\]
Note that the learner $\cL_w^{\istar}$ is updated only when $i_s=\istar$ that is why the indicator function appears in the regret above. Finally using this inequality in combination with \eqref{eq:to_UCB_nc} and the fact that the gains are optimistic (see Lemma~\ref{lem:confidence_bound_general}) we obtain
\begin{align*}
  \beta(t,\delta) + 50 A \left(h(t)\sqrt{\beta(t,\delta)}+ 2h(t)^2\right) +C_1\sqrt{t} +C_2 &\geq  \frac{1}{2}\sum_{s=1}^t\sum_{a\in\cA} w_s^{a,\istar}U_s^{a,\istar}\\
	&\geq \sup_{w\in\Sigma_{A}}\frac{1}{2}\sum_{s=1}^t \ind_{\{i_s=\istar\}}\sum_{a\in\cA} w^{a}U_s^{a,\istar}\\
  &\geq \sup_{w\in\Sigma_{A}}\frac{1}{2}\sum_{s=1}^t\ind_{\{i_s=\istar\}}\sum_{a\in\cA} w^{a} \normm{\theta -\lambda_s^{\istar}}^2_{aa^\top}\\
  &= N^{\istar}_t \sup_{w\in\Sigma_ {A}}\frac{1}{t}\sum_{s=1}^t\ind_{\{i_s=\istar\}}\frac{1}{2}\normm{\theta -\lambda_s^{\istar}}^2_{V_{w}}\\
  &\geq N^{\istar}_t \sup_{w\in\Sigma_A} \inf_{q\in\cP(\neg \istar)} \E_{\lambda\sim q}\frac{1}{2}\normm{\theta -\lambda}^2_{V_{w}}\\
 &= N^{\istar}_t \Tstar(\theta)^{-1}\,,
\end{align*}
where in the last line we use the Sion's minimax theorem, see~\eqref{eq:sion}.
\end{proof}

\paragraph{When $i_s \neq \istar$.} It remains to show that $N^{\istar}_t = t- O(\sqrt{t})$.

\begin{lemma}
\label{lem:lower_bound_nstar}
Under event $\cE_t$,
\begin{align*}
   N_t^{\istar} \geq t - \frac{2 A I}{\Delta_{\min}^2}\big(8h(t)^2 + C_1\sqrt{t} + C_2\big )\,,
\end{align*}
where
\[
C_1 = 4L^2M^2\sqrt{\log(A)}\qquad \text{and} \qquad C_2 = 64L^2M^2\left( 2+\log(A)\right)\,.
\]
\end{lemma}
\begin{proof}
 Under $\mathcal E_t$, using successively \eqref{eq:sum_dist_theta_nc} and the fact that $\theta\in \neg i$ for $i\neq \istar$, we get
\begin{align*}
2h(t)^2
\ge \sum_{s=1}^t \sum_{a\in \mathcal A} w_s^a \frac{1}{2}\Vert \theta - \hat{\theta}_{s-1} \Vert_{aa^\top}^2
&\ge \sum_{s=1}^t\ind_{\{i\neq \istar\}}  \frac{1}{2}\sum_{a \in \mathcal A} w_s^a\Vert \theta - \hat{\theta}_{s-1} \Vert_{aa^\top}^2
\\
&\ge \sum_{i \neq \istar} \inf_{\lambda\in \neg i} \sum_{s = 1}^t \frac{1}{2}\sum_{a \in \mathcal A} w_s^{a,i} \Vert \lambda - \hat{\theta}_{s-1} \Vert_{aa^\top}^2\\
&\geq \sum_{i \neq \istar} \sum_{s = 1}^t \frac{1}{2}\sum_{a \in \mathcal A} w_s^{a,i} \Vert \lambda_s^i - \hat{\theta}_{s-1} \Vert_{aa^\top}^2\\
\end{align*}
where for the last inequality we used the fact that $w_s^{a,i} = 0$ if $i_s\neq i$ and $w_s^{a,i} = w_s^{a}$ otherwise. Now proceeding exactly as in the proof of Lemma~\ref{lem:penultimate_lemma_sample_complexity_nc} we have
\begin{align*}
  \frac{1}{2}\sum_{s=1}^t\sum_{a\in\cA} w_s^{a,i} U_s^{a,i} \leq \frac{1}{2}&\sum_{s=1}^t\sum_{a\in\cA} w_s^{a,i} \Vert   \hat{\theta}_{s-1} - \lambda_s^i\Vert_{a a^\top}^2\\
  &+2h(t)^2 +2\sqrt{2}h(t)  \sqrt{\frac{1}{2}\sum_{s=1}^t \sum_{a\in\cA} w_s^{a}\Vert \htheta_{s-1} - \lambda_s^{i_s} \Vert_{a a^\top}^2}\,.
\end{align*}
Summing over $i\neq \istar$ in combination with the previous inequality then the inequality of Cauchy-Schwarz, we obtain
\[
\sum_{i\neq \istar}\frac{1}{2}\sum_{s=1}^t\sum_{a\in\cA} w_s^{a,i} U_s^{a,i} \leq 2h(t)^2 + 2h(t)^2 I + 4h(t)^2\sqrt{I} \leq 8Ih(t)^2\,.
\]
However, thanks to Proposition~\ref{lem:adahedge} for the algorithm AdaHedge we have the following regret bound for the learner $\cL_w^{i}$
\[
\max_{a\in\cA}\frac{1}{2}\sum_{s=1}^t \ind_{\{i_s=i\}} U_s^{a,i}- \frac{1}{2}\sum_{s=1}^t\sum_{a\in\cA} w_s^{a,i}U_s^{a,i}\leq \underbrace{4L^2M^2\sqrt{\log(A)}}_{\eqdef C_1}\sqrt{t}+\underbrace{64L^2M^2\left( 2+\log(A)\right)}_{\eqdef C_2}\,.
\]
Therefore combining the two previous inequalities leads to
\begin{equation}
I\big(8h(t)^2 + C_1\sqrt{t} + C_2\big ) \geq \sum_{i\neq \istar} \frac{1}{2}\max_{a\in\cA}\sum_{s=1}^t \ind_{\{i_s=i\}} U_s^{a,i} \geq
\sum_{i\neq \istar} \frac{1}{2A}\sum_{a\in\cA}\sum_{s=1}^t \ind_{\{i_s=i\}} U_s^{a,i}\,.
\label{eq:lb_mean_U_a}
\end{equation}

We now use the lemma stated below to lower bound one of the $U_s^{a,i}$. Lemma~\ref{lem:lower_bound_U} and \eqref{eq:lb_mean_U_a} then allow us to conclude
\begin{align*}
  I\big(8h(t)^2 + C_1\sqrt{t} + C_2\big ) \geq \sum_{i\neq \istar} \frac{1}{2A}\sum_{s=1}^t \ind_{\{i_s=i\}}  \Delta_{\min}^2 =  \frac{\Delta_{\min}^2}{2A} (t- N_t^{\istar})\,.
\end{align*}
\end{proof}
\begin{lemma}
\label{lem:lower_bound_U}
On $\mathcal E_t$, for all $s\le t$, if $i_s \neq \istar$ then there exists an arm $a\in\cA$ such that
\begin{align*}
U_s^{a,i_s} \ge \Delta_{\min}^2 \,,
\end{align*}
where $\Delta_{\min}^2 = \inf_{\lambda\in\neg \istar} \max_{a\in\cA} \langle\theta-\lambda ,a\rangle^2 >0\,.$
\end{lemma}
\begin{proof}
Consider the projection $\htheta_{s-1}^{\cM} \in \argmin_{\theta'\in\cM}\normm{\htheta_{s-1}-\theta'}_{V_{N_{s-1}}+\eta I_d}$. By definition of $i_s$ we know that $\istar(\htheta_{s-1}^{\cM}) = i_s \neq \istar$. Indeed if $\istar(\htheta_{s-1}^{\cM})=i \neq i_s$
then
\begin{align*}
\inf_{\lambda \in\neg i} \normm{\htheta_{s-1}-\lambda}_{V_{N_{s-1}}+\eta I_d} &< \inf_{\lambda \in\neg i_s }\normm{\htheta_{s-1}-\lambda}_{V_{N_{s-1}}+\eta I_d} \\
&\leq \inf_{\lambda\in\cM:\ \istar(\lambda) =i}\normm{\htheta_{s-1}-\lambda}_{V_{N_{s-1}}+\eta I_d}
= \normm{\htheta_{s-1} - \htheta_{s-1}^{\cM} }_{V_{N_{s-1}}+\eta I_d}\,,
\end{align*}
which leads to a contradiction since $\neg i\in\cM$.
In particular $\htheta_{s-1}^{\cM} \in \neg \istar$,
 thus there exists $a\in\cA$ such that
\[
\langle\theta- \htheta_{s-1}^{\cM},a\rangle^2 \geq \Delta_{\min}^2\,.
\]
Then by construction of the upper confidence bound, the definition of the projection  and the inequality of Cauchy-Schwarz,
\begin{align*} U_s^{a,i_s} &\geq 2h(t) \normm{a}^2_{(V_{N_{s-1}}+\eta I_d)^{-1}}\\
  &\geq   \normm{\theta - \htheta_{s-1}}^2_{V_{N_{s-1}}+\eta I_d} \normm{a}^2_{(V_{N_{s-1}}+\eta I_d)^{-1}}\\
  &\geq   \normm{\theta - \htheta_{s-1}^{\cM} }^2_{V_{N_{s-1}}+\eta I_d} \normm{a}^2_{(V_{N_{s-1}}+\eta I_d)^{-1}}\\
  &\geq \langle\theta- \htheta_{s-1}^{\cM},a\rangle^2 \geq \Delta_{\min}^2\,.
\end{align*}
\end{proof}

\paragraph{Conclusion.} Hence combining Lemma~\ref{lem:upper_bound_nstar} and Lemma~\ref{lem:lower_bound_nstar}, on the event $\cE_t$, if the algorithm does not stop
\begin{equation}\label{eq:conclusion_nc}
\beta(t,\delta) + 50 A \left(h(t)\sqrt{\beta(t,\delta)}+ 2h(t)^2\right) +C_1\sqrt{t} +C_2 +\frac{2 A I \Tstar(\theta)^{-1}}{\Delta_{\min}^2}\big(8h(t)^2 + C_1\sqrt{t} + C_2\big ) \geq \Tstar(\theta)^{-1} t\,.
\end{equation}
We can then conclude as in Appendix~\ref{app:analysis_under_concentration_cvx}. Let $\delta_{\min}$ be the largest $\delta\in(0,1)$ such that
\begin{align*}
\beta(\log(1/\delta)^{3/2},\delta) + 50 A \left(h\big(\log(1/\delta)\big)\sqrt{\beta(\log(1/\delta)^{3/2},\delta)}+ 2h(t)^2\right) +C_1\log(1/\delta)^{3/4} +C_2\\ +\frac{2 A I \Tstar(\theta)^{-1}}{\Delta_{\min}^2}\Big(8h\big(\log(1/\delta)^{3/2}\big)^2 + C_1\log(1/\delta)^{3/4} + C_2\Big)
< \Tstar(\theta)^{-1} \log(1/\delta)^{3/2}\,.
\end{align*}
Then for
\begin{align}
  \label{eq:def_T_0_nc}
  T_0(\delta) \eqdef \Tstar(\theta) \Bigg( \beta(\log(1/\delta)^{3/2},\delta) + 50 A \left(h\big(\log(1/\delta)\big)\sqrt{\beta(\log(1/\delta)^{3/2},\delta)}+ 2h(t)^2\right) +C_1\log(1/\delta)^{3/4} +C_2 \nonumber\\+\frac{2 A I \Tstar(\theta)^{-1}}{\Delta_{\min}^2}I\Big(8h\big(\log(1/\delta)^{3/2}\big)^2 + C_1\log(1/\delta)^{3/4} + C_2\Big)\Bigg)
\end{align}
we have the following lemma. Note that we have $T_0(\delta) = \Tstar(\theta)\log(1/\delta) +o\big(\log(1/\delta)\big)$.
\begin{lemma}
\label{lem:T_delta_nc}
If $\delta\leq \delta_{\min}$ then for $t\geq T_0(\delta)$ where $T_0(\delta)$ is defined in~\eqref{eq:def_T_0_nc}\,, $\cE_t \subset \{\tau_\delta \leq t\}$\,.
\end{lemma}

\subsection{Technical lemmas}
We regroup in this section  some technical lemmas.
\begin{lemma}
\label{lem:inq_revert_sqrt}
For all $\alpha,y\geq 0$, if for some $x\geq 0$ if holds $y \geq x-\alpha\sqrt{x}$ then
\[
x \leq y + \alpha \sqrt{y} + \alpha^2\,.
\]
\end{lemma}
\begin{proof}
Just note that for $z=\sqrt{x}$ we have
\[
z^2-\alpha z -y \leq 0\,,
\]
thus
\begin{align*}
  x \leq \frac{1}{4}\left(\alpha +\sqrt{\alpha^2+4y}\right)^2
  \leq y +\frac{\alpha^2}{2}+\frac{\alpha}{2}\sqrt{\alpha^2+4y}
  \leq y +\alpha\sqrt{y}+\alpha^2\,.
\end{align*}
\end{proof}

We then state a result derived from the concavity of $V\mapsto \log\det(V)$.
\begin{lemma}
\label{lem:sum_w_norm_a}
Let $(w_t)_{t\geq 1}$ be a sequence in $\Sigma_A$ and $\eta>0$ then
\[
\sum_{s=1}^t \sum_{a\in\cA} w_a^s \normm{a}^2_{W_s +\eta I_d} \leq d \log\left(1 +\frac{t L^2}{d \eta}\right)\,.
\]
where $W_t = \sum_{s=1}^t w_s$.
\end{lemma}
\begin{proof}
Define the function $f(W)= \log\det(V_W+\eta I_d)$ for any $W\in(\R^+)^A$. It is a concave function since the function
$V\mapsto \log\det(V)$ is a concave function over the set of positive definite matrices (see Exercise 21.2 of \citealt{lattimore2018}). And its partial derivative with respect to the coordinate $a$ at $W$ is
\[
\nabla_a f(W) = \normm{a}^2_{(W+\eta I_d)^{-1}}\,.
\]
Hence using the concavity of $f$ we have
\begin{align*}
  \sum_{a\in\cA} w_a^s \normm{a}^2_{(V_{W_s} +\eta I_d)^{-1}} = \langle W_s - W_{s-1}, \nabla_a f(W_s) \rangle \leq f(W_s) - f(W_{s-1})\,.
\end{align*}
Which implies that
\begin{align*}
  \sum_{s=1}^t \sum_{a\in\cA} w_a^s \normm{a}^2_{V_{W_s} +\eta I_d} \leq f(W_t)-f(W_0) = \log\left(\frac{\det(V_{W_t} +\eta I_d) }{\det(\eta I_d)}\right) \leq d \log\left(1 +\frac{t L^2}{d \eta}\right)\,,
\end{align*}
where for the last inequality we use the inequality of arithmetic and geometric means in combination with $\Tr(W_t) \leq t L^2$\,.
\end{proof}

A simple consequence of the previous lemma follows.
\begin{lemma} For all $t$,
\label{lem:computation_sum_w_a_N}
\begin{align*}
\sum_{s=1}^t \sum_{a\in\cA} \tw_s^{a} \normm{ a }_{(V_{N_{s-1}}+\eta I_d)^{-1}}^2 &\leq 2h(t) = 2\beta(t,1/t^\alpha) \\
\sum_{s=1}^t \sum_{a\in\cA} w_s^{a} \normm{ a }_{(V_{N_{s-1}}+\eta I_d)^{-1}}^2 &\leq 2h(t)\,.
\end{align*}
\end{lemma}

\begin{proof}
According to the tracking procedure (Lemma~\ref{lem:tracking}), we know that $N_{s-1}^a \geq \tW_{s-1}^a -\log(AI)$. Thus, in combination with the choice of $\eta$  we can replace counts by weights
\begin{align*}
  V_{N_{s-1}} + \eta I_d \geq V_{\tW^a_{s}} - V_{\tw_s^a} - \log(AI) V_{\bOne_A} +\eta I_d \geq  V_{\tW^a_{s}} - (\log(A)+1) V_{\bOne_A} +\eta I_d \geq V_{W_{s}}+ \frac{\eta}{2} I_d\,,
\end{align*}
where $\bOne_A = (1,\ldots,1)\in\R^A$.
Hence we obtain
\[
\normm{ a }_{(V_{N_{s-1}}+\eta I_d)^{-1}}^2 \leq \normm{ a }_{(V_{\tW^a_{s}}+(\eta/2) I_d)^{-1}}^2\,,
\]
and applying Lemma~\ref{lem:sum_w_norm_a} leads to
\[
\sum_{s=1}^t \sum_{a\in\cA} \tw_s^{a} \normm{ a }_{(V_{N_{s-1}}+\eta I_d)^{-1}}^2 \leq d \log\!\left(1 +\frac{t L^2}{d \eta} \right)\leq 2h(t)\,.
\]
The exact same proof holds for $w^a_s$ instead of $\tw_s^a$ since thanks to the tracking we have also in this case $N_{s-1}^a \geq W_{s-1}^a -\log(A) \geq W_{s-1}^a -\log(AI)$.

\end{proof}

%\subfile{appendix/nc_sample_complexity_proof}
%!TEX root = ../lin_bandit_explo.tex
\section{Concentration Results}\label{app:concentration}

We restate here the Theorem~20.4 (in combination with the Equation~20.10) by \citet{lattimore2018}.
\begin{theorem}
\label{th:confidence_beta}
For all $\eta >0$ and $\delta\in(0,1)$,
\[
\P_{\theta}\!\!\left(\exists t\in\NN,\, \frac{1}{2}\normm{\htheta_t-\theta}^2_{V_{N_t}+\eta I_d} \geq \beta(t,\delta)\right) \leq \delta\,,
 \]
 where
 \begin{align*}
\beta(t,\delta) &\eqdef  \left( \sqrt{\log\!\left( \frac{1}{\delta}\right)+\frac{d}{2}\log\!\left(1+\frac{t L^2}{\eta d} \right)} +\sqrt{\frac{\eta}{2}}M\right)^2\\
&=\log\!\left( \frac{1}{\delta}\right)+\frac{d}{2}\log\!\left(1+\frac{t L^2}{\eta d} \right) +  M\sqrt{\eta}\sqrt{2\log\!\left( \frac{1}{\delta}\right)+d\log\!\left(1+\frac{t L^2}{\eta d} \right)}+\frac{\eta M^2}{2}\,.
%\left(\sqrt{\log\!\left(\frac{1}{\delta}\right) + \frac{d}{2}\log\!\left(1+ \frac{tL^2}{\eta d}\right)}   +\sqrt{\frac{\eta}{2}}M\right)^2\,.
\end{align*}
\end{theorem}

The Lemma~\ref{lem:chernoff_stopping rule_pac} is a simple consequence of this theorem.

\begin{proof}[Proof of Lemma~\ref{lem:chernoff_stopping rule_pac}]
Using the fact that $\theta \in  \neg \istar_t$ when $\istar_t \neq \istar(\theta)$ and Theorem~\ref{th:confidence_beta}, it follows
\begin{align*}
\P_{\theta}\big(\tau_{\delta} < \infty \wedge \istar_{\tau_\delta} \neq \istar(\theta)\big) &\leq
\P_{\theta}\!\left(\exists t\in\N, \inf_{\lambda \in \neg \istar_t}\frac{1}{2}\Vert \htheta_t - \lambda \Vert^2_{V_{N_t}} > \beta(t,\delta),\, \istar_t\neq \istar(\theta)\right)\\
&\leq \P_{\theta}\!\left(\exists t\in\N, \frac{1}{2}\Vert \htheta_t - \theta \Vert^2_{V_{N_t}+\eta I_d} \geq \beta(t,\delta)\right)\\
&\leq \delta\,.
\end{align*}

\end{proof}

\DeclareBoldMathCommand{\zeros}{0}
\DeclareBoldMathCommand{\v}{v}
\DeclareBoldMathCommand{\w}{w}
\DeclareBoldMathCommand{\e}{e}
\DeclareBoldMathCommand{\u}{u}

\section{Tracking}\label{app:tracking}

We call tracking the following interaction. Starting from vectors $W_0 = N_0 = (0, \ldots, 0) \in \mathbb{R}^K$, for each stage $t=1,2,\ldots$
\begin{itemize}
	\item Nature reveals a vector $w_t$ in the simplex $\Sigma_K$ and updates $W_t = W_{t-1} + w_t$.
	\item A tracking rule selects $k_t \in [K]$ based on $(w_1, \ldots, w_t)$ and forms $N_t = N_{t-1} + e_{k_t}$, where $(e_i)_{i\in [K]}$ is the canonical basis.
\end{itemize}
Note that $w_t$ is known to the tracking rule when choosing $k_t$.

\begin{definition}
We call C-Tracking any rule which for all stages $t\ge 1$ selects $k_t \in \argmin_{k\in [K]} N_{t-1}^k - W_t^k$.
\end{definition}
This defines C-tracking up to the choice of $k_t$ when the $\argmin$ is not unique. The name stands for cumulative tracking and is introduced by~\citet{garivier2016tracknstop}.

\begin{lemma}\label{lem:tracking}
The C-Tracking procedure described above ensures that for all $t\in \N$, for all $k\in [K]$,
\begin{align*}
-\sum_{j=2}^K \frac{1}{j} \le N_t^k - W_t^k \le 1 \: .
\end{align*}
\end{lemma}
The upper bound is given by~\citet{garivier2016tracknstop}. The lower bound is due to~\citet{anon2020structure} and is reproduced below.
\begin{proof}
Let $\mathcal S_0 = \{v \in \mathbb{R}^K \: : \: \sum_{k=1}^K v^k = 0\}$. The tracking procedure is such that for all stages $t\in\N$, $N_t-W_t \in S_0$. Our proof strategy is to characterize the subset of $\mathcal S_0$ that can be reached during the tracking procedure, starting from $\v_0 = N_0-W_0 = \zeros$ .

We define a move $\rightarrow_\w$ as function from $\mathcal{S}_0$ to itself parametrized by $\w$ that maps $\v$ to $\v - \w + \e_k$, where $k=\argmin_{j\in[K]} \v - \w$. If the value of that function at $\v$ is $\u$, we write $\v \rightarrow_\w \u$ .
A vector $\u \in \mathcal{S}_0$ is said to be reachable in one move from $\v \in \mathcal{S}_0$ if there exists $\w \in \triangle_K$ such that $\v \rightarrow_\w \u$. We denote it by $\v \rightarrow \u$. It is said to be reachable from $\v$ if there is a finite sequence of such moves such that $\v \rightarrow \ldots \rightarrow \u$.

A reverse move $\leftarrow_{k,\w}$ is a function from $\mathcal{S}_0$ to itself parametrized by $k$ and $\w$ that maps $\v$ to $\v + \w - \e_k$. A reverse move is said to be valid at $\v$ if $v^k \leq \min_j v^j + 1$. If the value of that function at $\v$ is $\u$, we write $\u \leftarrow_{k,\w} \v$ .
A vector $\u \in \mathcal{S}_0$ is said to be reverse-reachable in one move from $\v \in \mathcal{S}_0$ if there exists $k\in[K]$ and $\w \in \triangle_K$ such that $\u \leftarrow_{k,\w} \v$ and such that this is a valid reverse move at $\v$. We denote it by $\u \leftarrow \v$.

We now prove that $(\u \rightarrow \v) \Leftrightarrow (\u \leftarrow \v)$ . First, if $\u \rightarrow \v$ then let $\w$ be the parameter of a move $\u \rightarrow_\w \v$ and let $k = \argmin_{j\in[K]} \u - \w$. Then $\u \leftarrow_{k,\w} \v$ is a valid reverse move. Second, if $\u \leftarrow_{k,\w} \v$ is a valid reverse move, then $k = \argmin_{j\in[K]} \u - \w$ and we have $\u \rightarrow_\w \v$ .

We characterize the elements $\v$ of $\mathcal S_0$ such that $\zeros \leftarrow \ldots \leftarrow \v$ .

Let $\u,\v \in \mathcal{S}_0$ be such that $\u \leftarrow \v$. Let $M_\v = \{k\in[K]: v^k \le \min_j v^j + 1\}$. Then for any set $S\subseteq[K]$ such that $M_\v\subseteq S$, $\sum_{i\in S} u^i \le \sum_{i\in S} v^i$. Indeed, for the reverse move to be valid, one of the coordinates in $M_\v$ was decreased by 1, and they were added coordinates of a $\w\in \triangle_K$, that sum at most to 1.

Let $S\subseteq [K]$ and $A_S = \{u\in\mathcal{S}_0 : \forall k \notin S, \: u^k > \frac{1}{|S|}\sum_{i\in S}u^i + 1\}$. We now prove that if $\u \leftarrow \v$ and $\v \in A_S$, then $\u \in A_S$ and as a consequence, that if $\u \leftarrow \ldots \leftarrow \v$ and $\v \in A_S$, then $\u \in A_S$. Indeed,
\begin{itemize}[nosep]
	\item Since $\v \in A_S$, we have $M_\v \subseteq S$, hence the previous remark proves $\frac{1}{|S|}\sum_{i\in S}u^i \leq \frac{1}{|S|}\sum_{i\in S}v^i$.
	\item For $k\notin S$, then $k\notin M_\v$ and $u^k \geq v^k > \frac{1}{|S|}\sum_{i\in S}v^i + 1 \geq \frac{1}{|S|}\sum_{i\in S}u^i + 1$ .
\end{itemize}

Since $\zeros \notin \bigcup_{S \in \mathcal{P}([K])\setminus\{[K]\}} A_S$, we can now state that if $\zeros \leftarrow \ldots \leftarrow \v$, then $\v \notin \bigcup_{S \in \mathcal{P}([K])\setminus\{[K]\}} A_S$ .

Let $j\in[2:K]$ and let $\v_{(j)} \in \mathcal{S}_0$ be such that $v_{(j)}^1 \ge \ldots \ge v_{(j)}^{j-1} > v_{(j)}^j = \ldots = v_{(j)}^K$. Then we will prove that one of the two following statements is true:
\begin{enumerate}
\item $v_{(j)}^{j-1} > v_{(j)}^j + 1$ and $\v_{(j)}$ is not reachable from $\zeros$,
\item $v_{(j)}^{j-1} \le v_{(j)}^j + 1$. Then let $v_{j-1:K}$ be the mean of $v_{(j)}^{j-1},\ldots,v_{(j)}^K$ and let $\v_{(j-1)}$ be the vector with $v_{(j-1)}^1 = v_{(j)}^1$, ..., $v_{(j-1)}^{j-2} = v_{(j)}^{j-2}$ and $v_{(j-1)}^{j-1} = \ldots = v_{(j-1)}^K = v_{j-1:K}$. Then $\v_{(j-1)} \leftarrow \v_{(j)}$ .
\end{enumerate}
Case 1: $v_{(j)}^{j-1} > v_{(j)}^j + 1$. Let $S = [j:K]$. then $S\neq[K]$ and $\v\in A_S$. Hence $\v$ is not reachable from $\zeros$.

Case 2: $v_{(j)}^{j-1} \le v_{(j)}^j + 1$. Let $\w$ be defined by $w^1 = \ldots = w^{j-2} = 0$, $w^{j-1} = 1 - (K-j+1)(v_{j-1:K} - v_{(j)}^j)$ and $w^j = \ldots = w^K = v_{j-1:K} - v_{(j)}^j$. Since $v_{(j)}^{j-1} \le v_{(j)}^j + 1$ and $\w\in\triangle_K$, the reverse move $\leftarrow_{j-1,\w}$ is valid at $\v_{(j)}$. Then the image of $\v_{(j)}$ by that reverse move is $\v_{(j-1)}$. Note that $w^{j-1} \ge w^j = \ldots = w^K$.

We have now all the tools to state the characterization of the the elements $\v$ of $\mathcal S_0$ such that $\zeros \leftarrow \ldots \leftarrow \v$ . By a simple induction using the last case distinction, we have the following:
let $\v \in \mathcal S_0$ and let $i_1, \ldots, i_K\in[K]$ be such that $v^{i_1} \ge \ldots \ge v^{i_K}$. If $\zeros \leftarrow \ldots \leftarrow \v$, then there exists $\u_1, \ldots, \u_{K-2}$ and $\w_1, \ldots, \w_{K-1}$ such that
\begin{enumerate}
\item $\zeros \leftarrow _{i_1,\w_1} \u_1 \leftarrow_{i_2,\w_2} \ldots \leftarrow_{i_{K-2},\w_{K-2}} \u_{K-2} \leftarrow_{i_{K-1}, \w_{K-1}} \v$,
\item for all $j \in [K-2]$, $u_j^{i_1} \ge \ldots u_j^{i_j} \ge u_j^{i_{j+1}} = \ldots = u_j^{i_K} = \frac{1}{K-j}\sum_{k=j+1}^K v^{i_k}$ ,
\item for all $j \in [K-1]$, $w_j^{i_1} = \ldots w_j^{i_{j-1}} = 0$ and $w_j^j \ge w_j^{i_{j+1}} = \ldots = w_j^{i_K}$. 
\end{enumerate}
In order to prove the theorem, we then only need a bound on $v^{i_K} = - \sum_{j=1}^{K-1} w_j^K$. The characterization of $\w_j$ implies that $w_j^K \leq 1/(K-j+1)$ . Hence $v^{i_K} \ge - \sum_{j=2}^{K} \frac{1}{j}$ .

\end{proof}

%\subfile{appendix/adaptation_BAI} merged with example
%!TEX root = ../lin_bandit_explo.tex
\section{A Fair Comparison of Stopping Rules}\label{app:stopping}

We investigate closely the stopping rules employed in existing linear BAI algorithm. We first make a synthesized table that resembles stopping rules and decision rules of all existing algorithms, including ours, in Table~\ref{tab:stopping_rules}. We denote by $\hat\cA_t$ the active arm set for elimination-based algorithms, and by $i_{\hat\cA_t}$ the only arm left in $\hat\cA_t$ when $|\hat\cA_t|=1$.

We show that they are all the same up to the choice of the exploration rate. Note that in Table~\ref{tab:stopping_rules}, we have replaced all the exploration term by $\beta(t,\delta)$, and we have also listed the original terms (with their original notation, thus may be in conflict with notation of the current paper). In the following, we always use the same exploration rate $\beta(t,\delta)$ for all stopping rules.

\begin{table}[ht]
    \centering
	\small
	\caption{Stopping rules for different algorithms}
	\label{tab:stopping_rules}
	\begin{tabular}{@{}llll@{}}
		\toprule
		\thead{Algorithm} & \thead{Stopping rule} & \thead{Original term} & \thead{Decision rule} \\
		\midrule
		\XYS & \makecell{$\exists a\in\cA, \forall a'\neq a, \normm{a-a'}_{V_{N_t}^{-1}}\sqrt{2\beta(t,\delta)}\leq \langle\hat\theta_t,a - a'\rangle$} & $\log_t(A^2/\delta)$ & $\hi = i^\star(\hat{\theta}_{t})$ \\
		\XYA & \makecell{$|\hat\cA_t|=1$, where all arms  $a\in\cA$ s.t. \\ $\exists a'\in\cA, \normm{a-a'}_{V_{N_t}^{-1}}\sqrt{2\beta(t,\delta)}\leq \langle\hat\theta_t,a' - a\rangle$ are discarded} & $\log_t(A^2/\delta)$ & $\hi = i_{\hat\cA_t}$ \\
		\ALBA & \makecell{$|\hat\cA_t|=1$, where all arms  $a\in\cA$ s.t. \\ $\dfrac{\normm{a^\star(\hat\theta_t)-a}_{V_{N_t}^{-1}}}{\sqrt{1/2\beta(t,\delta)}}\leq\langle\hat\theta_t,a^\star(\hat\theta_t)-a\rangle$ are discarded} & $1/\ell_t$ & $\hi = i_{\hat\cA_t}$ \\
		\RAGE & \makecell{$|\hat\cA_t|=1$, where all arms  $a\in\cA$ s.t. \\ $\exists a'\in\cA, 2^{-t-2}\leq\langle\hat\theta_t,a' - a\rangle$ are discarded} & - & $\hi = i_{\hat\cA_t}$ \\
		\LGapE & \makecell{$\langle\hat\theta_t,a_{j_t}-a^\star(\hat\theta_t)\rangle + \normm{a^\star(\hat\theta_t) - a_{j_t}}_{V_{N_t}^{-1}} \sqrt{2\beta(t,\delta)} < 0$ \\ with $j_t = \argmax_{j\in\cI} \langle\hat\theta_t,a_{j}-a^\star(\hat\theta_t)\rangle + \normm{a^\star(\hat\theta_t) - a_{j}}_{V_{N_t}^{-1}} \sqrt{2\beta(t,\delta)} $} & $C_t$ & $\hi = i^\star(\hat{\theta}_{t})$ \\
		\GLGapE & \makecell{$\langle\hat\theta_t,a_{j_t}-a^\star(\hat\theta_t)\rangle + \normm{a^\star(\hat\theta_t) - a_{j_t}}_{V_{N_t}^{-1}} \sqrt{2\beta(t,\delta)} < 0$ \\ with $j_t = \argmax_{j\in\cI}\langle\hat\theta_t,a_{j}-a^\star(\hat\theta_t)\rangle + \normm{a^\star(\hat\theta_t) - a_{j}}_{V_{N_t}^{-1}} \sqrt{2\beta(t,\delta)} $} & $C_t$ & $\hi = i^\star(\hat{\theta}_{t})$ \\
		\GLUCB & \makecell{$\max_{i\in \cI} \inf_{\lambda\in\neg i} \dfrac{\normm{\htheta_t-\lambda}^2_{V_{N_t}}}{2}\geq \beta(t,\delta)$} & - & $\hi = i^\star(\hat{\theta}_{t})$ \\
		\midrule
		\algoW & \makecell{$\max_{i\in \cI} \inf_{\lambda\in\neg i} \dfrac{\normm{\htheta_t-\lambda}^2_{V_{N_t}}}{2}\geq \beta(t,\delta)$} & - & $\hi = i_{t+1}$ \\
		\algoWCvx & \makecell{$\max_{i\in \cI} \inf_{\lambda\in\neg i} \dfrac{\normm{\htheta_t-\lambda}^2_{V_{N_t}}}{2}\geq \beta(t,\delta)$} & - & $\hi = i^\star(\hat{\theta}_{t})$ \\
		\bottomrule
	\end{tabular}
\end{table}

\paragraph{\algoW and \algoWCvx.}
We first notice that using the same argument as the proves of Lemma~\ref{lem:lagrange_alternative} and Lemma~\ref{lem:complexity_bai}, the stopping rule of \algoW and \algoWCvx (and also the one of \GLUCB) can be rewritten as
\[
	\min_{i \neq i^\star(\hat{\theta}_{t})} \dfrac{\langle\hat\theta_t,a_{i}-a^\star(\hat\theta_t)\rangle^2}{2\normm{a_i - a^\star(\hat\theta_t)}_{{V_{N_t}^{-1}}}^2} \1\left\{a^\star(\hat\theta_t)^\top\hat\theta_t \geq a_i^\top\hat\theta_t\right\} > \beta(t,\delta)\,.
\]
Now we compare it with other stopping rules.

\paragraph{\algoWCvx and \algoW $\Rightarrow$ \XYS.}
If \algoWCvx stops at time $t$, then for $a = a^\star(\hat\theta_t)$, we have
\[
    \forall a'\neq a, \normm{a-a'}_{V_{N_t}^{-1}}\sqrt{2\beta(t,\delta)}\leq \langle\hat\theta_t,a - a'\rangle\,,
\]
and \XYS stops as well.

\paragraph{\XYS $\Rightarrow$ \XYA.}
Suppose that \XYS stops at time $t$ under its stopping rule, then
\[
\exists a\in\cA, \forall a'\neq a, \normm{a-a'}_{V_{N_t}^{-1}}\sqrt{2\beta(t,\delta)}\leq \langle\hat\theta_t,a - a'\rangle\,.
\]
It is clear that if such $a$ exists, then it can only be the empirical best arm $a^\star(\hat\theta_t)$. Thus,
\[
    \forall a'\neq a^\star(\hat\theta_t), \normm{a^\star(\hat\theta_t)-a'}_{V_{N_t}^{-1}}\sqrt{2\beta(t,\delta)}\leq\langle\hat\theta_t,a^\star(\hat\theta_t)-a'\rangle\,,
\]
and all arms different from $a^\star(\hat\theta_t)$ would be discarded under $\XYA$. Furthermore, $a^\star(\hat\theta_t)$ would never be discarded since
\[
    \forall a'\neq a^\star(\hat\theta_t), \langle\hat\theta_t,a'-a^\star(\hat\theta_t)\rangle < 0 \le \normm{a^\star(\hat\theta_t)-a'}_{V_{N_t}^{-1}}\sqrt{2\beta(t,\delta)}\,,
\]
and \XYA stops.

\paragraph{\XYA $\Rightarrow$ \ALBA}
Now if \XYA stops at time $t$, then all arms but $a^\star(\hat\theta_t)$ are discarded, and
\[
    \forall a \neq a^\star(\hat\theta_t), \normm{a^\star(\hat\theta_t)-a'}_{V_{N_t}^{-1}}\sqrt{2\beta(t,\delta)} = \dfrac{\normm{a^\star(\hat\theta_t)-a}_{V_{N_t}^{-1}}}{\sqrt{1/2\beta(t,\delta)}} \le \langle\hat\theta_t,a^\star(\hat\theta_t)-a\rangle\,.
\]
Therefore, those arms would also be discarded under \ALBA, and \ALBA stops.

\paragraph{\ALBA $\Rightarrow$ \LGapE and \GLGapE.}
Next, suppose that \ALBA stops at time $t$ under its stopping rule, then the only arm left would be $a^\star(\hat\theta_t)$, and
\[
    \forall a \neq a^\star(\hat\theta_t), \dfrac{\normm{a^\star(\hat\theta_t)-a}_{V_{N_t}^{-1}}}{\sqrt{1/2\beta(t,\delta)}} \le \langle\hat\theta_t,a^\star(\hat\theta_t)-a\rangle\,.
\]
And in particular, we get
\[
    \langle\hat\theta_t,a_{j_t}-a^\star(\hat\theta_t)\rangle + \normm{a^\star(\hat\theta_t) - a_{j_t}}_{V_{N_t}^{-1}} \sqrt{2\beta(t,\delta)} < 0\,.
\]
Thus \LGapE/\GLGapE stops under its stopping rule.

\paragraph{\LGapE $\Rightarrow$ \algoWCvx and \algoW}
Finally, we suppose that \LGapE stops at time $t$, then it comes
\begin{align*}
    j_t &= \argmax_{j\in\cI} \langle\hat\theta_t,a_{j}-a^\star(\hat\theta_t)\rangle + \normm{a^\star(\hat\theta_t) - a_{j}}_{V_{N_t}^{-1}} \sqrt{2\beta(t,\delta)} \\
    &= \argmin_{j\in\cI} \langle\hat\theta_t,a^\star(\hat\theta_t)-a_{j}\rangle - \normm{a^\star(\hat\theta_t) - a_{j}}_{V_{N_t}^{-1}} \sqrt{2\beta(t,\delta)}\,.
\end{align*}
By consequence,
\begin{align*}
	\min_{i \neq i^\star(\hat{\theta}_{t})} \dfrac{\langle\hat\theta_t,a_{i}-a^\star(\hat\theta_t)\rangle^2}{2\normm{a_i - a^\star(\hat\theta_t)}_{{V_{N_t}^{-1}}}^2} \1\left\{a^\star(\hat\theta_t)^\top\hat\theta_t \geq a_i^\top\hat\theta_t\right\} &= \dfrac{\langle\hat\theta_t,a_{j_t}-a^\star(\hat\theta_t)\rangle^2}{2\normm{a_{j_t} - a^\star(\hat\theta_t)}_{{V_{N_t}^{-1}}}^2}\\
	&\geq \beta(t,\delta)\,,
\end{align*}
and \algoWCvx stops as well.

In conclusion, all the stopping rules are equivalent if we set their exploration term to the same, though formulated in different manners.
%!TEX root = ../lin_bandit_explo.tex
\section{Implementation Details}\label{app:implem}

In this section, we provide some further experiment details. We also share a few insights over different aspects of implementations of different pure exploration linear bandits algorithms. In particular, we propose a new Frank-Wolfe-typed heuristic to solve generic $\cA\cB$-design.

\subsection{Experimental setting}

\paragraph{More details for algorithm implementations.} We give more clarifications on each individual algorithm implemented.

\begin{itemize}
	\item For our algorithms \algoW and \algoWCvx, we implemented the version with the boundedness assumption.
	\item For \LGapE We implemented the greedy version, that is, pull the arm $\argmin_{a\in\cA} \normm{a_{i_t}-a_{j_t}}_{(V_{N_t}+aa^\top)^{-1}}^2$ with $i_t = i^\star(\hat\theta_t)$ and $j_t = \argmax_{j\neq i_t}\langle\hat\theta_t,a_{j}-a^\star(\hat\theta_t)\rangle + \normm{a^\star(\hat\theta_t) - a_{j_t}}_{V_{N_t}^{-1}} \sqrt{2\beta(t,\delta)}$. Note that this version does not have a theoretical guarantee in the general case. However, as we stated in Section~\ref{sec:related_work}, the \GLUCB proposed by~\citet{zaki2019maxoverlap} is equivalent to this greedy version of \LGapE, and they provided an analysis for the 2-arm and 3-arm case. \LGapE is designed for $\epsilon$-best-arm identification, we set $\epsilon=0$ in our experiments to make sure that it outputs the optimal one.
	\item For \XYS, we implemented the greedy incremental version for both $\gopt$-allocation and $\xyopt$-allocation, that allows us to avoid the optimal design-computing step. To implement the non-greedy version, readers are invited to look at next Section~\ref{app:implem_complexity} where we discuss in detail the computation of $\cA\cB$-optimal design.
	\item For \XYA, it requires a hyper-parameter that characterizes the length of each phase. We set that hyper-parameter to $0.1$ as done by~\citet{soare2014linear}.
\end{itemize}

\paragraph{Technical details.} All the algorithms and experiments are implemented in \lstinline{Julia 1.3.1}, and plots are generated using the \lstinline{StatsPlots.jl} package. Other external dependencies are: \lstinline{JLD2.jl, Distributed.jl, IterTools.jl, CPUTime.jl, LaTeXStrings.jl}.

\paragraph{For reproducibility.} To rerun our code, your need to have Julia installed, then unzip \lstinline{code.zip} and do the following in your terminal.

\begin{lstlisting}
  $ cd PATH/TO/THE/FOLDER/code/linear
  $ julia
  julia> include("experiment_bai1.jl") # reproduce Fig.1
  julia> include("viz_bai1.jl") # visualization
  julia> include("experiment_bai2.jl") # reproduce Fig.2
  julia> include("viz_bai2.jl") # visualization
\end{lstlisting}

%\subsubsection{Arm generation.} In Section~\ref{sec:experiments}, we need to generate arms uniformlly from a unit sphere of arbitrary dimension. This can be done using Algorithm~\ref{alg:generation}, it can be trivially extended to higher dimension.
%
%\begin{algorithm}[ht]
%   \caption{Generating arms from the 2D unit sphere (Box-Mueller)}
%   \label{alg:generation}
%\begin{algorithmic}
%        \STATE $u \sim \cN(0,1)$
%        \STATE $v \sim \cN(0,1)$
%        \STATE $(x, y) = (u, v)/\normm{(x, y)}_2$
%        \RETURN $x, y$
%\end{algorithmic}
%\end{algorithm}

\subsection{Computation of different complexities}\label{app:implem_complexity}

As mentioned in Section~\ref{sec:lower_bound}, computing the solution to a specified optimization problem is required in many existing linear BAI algorithms. We survey some methods that can potentially be useful to handle that issue.

We recall that the three notions of complexity $\gopt, \xyopt, \cA\cBstar(\theta)$ can be written in a unified form,
\begin{align}\label{eq:complexity_general}
    \cA\cB = \min_{w\in\Sigma_A} \max_{b\in\cB}\normm{b}_{V_w^{-1}}^2,
\end{align}
where $\cB$ is the transductive set, i.e. a finite set of elements in $\R^d$. Transductive sets corresponding to different complexity types mentioned in this paper can be found in Table~\ref{tab:transductive_sets}.

\begin{table}[ht]
    \centering
	\caption{Different transductive sets}
	\label{tab:transductive_sets}
	\begin{tabular}{@{}lll@{}}
		\toprule
		\thead{Allocation type} & \thead{Arm set} & \thead{Transductive set} \\ \midrule
		(1) $\gopt$-allocation & $\cA$ & $\cA$\\
		(2) $\xyopt$-allocation & $\cA$ & $\cB_{\texttt{dir}} = \{a-a':\ (a,a')\in\cA\times\cA\}$ \\
		(3) $\cA\cBstar(\theta)$-allocation & $\cA$ & $\cB^\star(\theta) = \left\{ (\astar(\theta)- a)/\left|\big\langle \theta, \astar(\theta)-a\big\rangle\right|: a\in\cA/\big\{\astar(\theta)\big\}  \right\}$ \\
		\bottomrule
	\end{tabular}
\end{table}

\paragraph{Frank-Wolfe.} We can use a Frank-Wolfe heuristic to compute the optimizer of~\eqref{eq:complexity_general} shown in Algorithm~\ref{alg:algoFW}. This heuristic is used for example by~\citet{fiez2019transductive}. Note that this heuristic has been proved to have a linear convergence guarantee when $\cB = \cA$~\citep{ahipasaoglu2008fw}. It is not clear, however, that the same guarantee holds for other transductive sets.

A simple sanity check to test whether a solver works smoothly is  to solve $\cA\cBstar(\theta)$ for classical multi-armed bandits (i.e. when $\cA= \{e_1,e_2,\ldots,e_d\}$), for which a solver with guarantee exists (see \citealt{garivier2018explore}). In particular we found instances where Algorithm~\ref{alg:algoFW} does not converge toward the optimal weights, for example: $\cA= \{e_1,e_2,e_3\}, \theta = (0.9, 0.5, 0.5)$.

% \begin{algorithm}[ht]
%    \caption{Frank-Wolfe heuristic for computing $\gopt$-design}
%    \label{alg:algoFW}
% \begin{algorithmic}
%    \STATE {\bfseries Input:} arm set $\cA\subset\R^d$, transductive set $\cB\subset\R^d$, maximum iterations $n$
%    \STATE  {\bfseries Initialize:} $w \leftarrow{} (1/A, 1/A, \ldots, 1/A), V \leftarrow{} I_d, t \leftarrow{} 0$
%    \WHILE{$t<n$}
%         \STATE $i_{t}\in\argmin_{i\in\{1,\ldots,A\}}\max_{b\in\cB}\normm{b}^2_{(V+a_i a_i^\top)^{-1}}$
%         \STATE $V \leftarrow{} V + a_{i_{t}}a_{i_{t}}^\top$
%         \STATE $w \leftarrow{} \frac{t}{t+1}w+\frac{1}{t+1}e_{i_{t}}$
%         \STATE $t \leftarrow{} t+1$
%    \ENDWHILE
%    \RETURN $w$
% \end{algorithmic}
% \end{algorithm}

\begin{algorithm}[ht]
   \caption{Frank-Wolfe heuristic for computing generic $\cA\cB$-design}
   \label{alg:algoFW}
\begin{algorithmic}
   \STATE {\bfseries Input:} arm set $\cA\subset\R^d$, transductive set $\cB\subset\R^d$, maximum iterations $n$
   \STATE  {\bfseries Initialize:} $w \leftarrow{} (1, 1, \ldots, 1)\in\R^A, V \leftarrow{} I_d, t \leftarrow{} 0$
   \WHILE{$t<n$}
        \STATE $\ta\in\argmax_{a\in\cA}\max_{b\in\cB}\langle a,b\rangle_{V^{-1}}^2$%\normm{b}^2_{(V+a_i a_i^\top)^{-1}}$
        \STATE $V \leftarrow{} V + \ta \ta^\top$
        \STATE $w \leftarrow{} \frac{t}{t+1}w+\frac{1}{t+1}e_{\ta}$
        \STATE $t \leftarrow{} t+1$
   \ENDWHILE
   \RETURN $w$
\end{algorithmic}
\end{algorithm}

We propose a variant of the previous heuristic that takes into account a count for each element in the transductive set $\cB$. The pseudo-code of our method is displayed in Algorithm~\ref{alg:algoFW2}. % $N\in\N^{|\cB|}$ denotes the vector of counts for all $b\in\cB$.
Sanity check on various MAB instances shows the correctness of our heuristic, its convergence guarantee remains for the future work.

% \begin{algorithm}[ht]
%    \caption{Saddle Frank-Wolfe heuristic for computing generic $\cA\cB$-design}
%    \label{alg:algoFW2}
% \begin{algorithmic}
%    \STATE {\bfseries Input:} arm set $\cA\subset\R^d$, transductive set $\cB\subset\R^d$, maximum iterations $n$
%    \STATE  {\bfseries Initialize:} $w \leftarrow{} (1/A, 1/A, \ldots, 1/A), N \leftarrow{} (1, 1, \ldots, 1), V \leftarrow{} I_d, t \leftarrow{} 0$
%    \WHILE{$t<n$}
%         \STATE $i_{t}\in\argmin_{i\in\{1,\ldots,A\}}\sum_{j=1}^B(-2N[j]\langle a_i,V b_j\rangle^2)$
%         \STATE $j_{t}\in\argmax_{j\in\{1,\ldots,B\}}\normm{b_{j}}^2_{V^{-1}}$
%         \STATE $V \leftarrow{} V + a_{i_{t}}a_{i_{t}}^\top$
%         \STATE $N[j_t] \leftarrow{} N[j_t] + 1$
%         \STATE $w \leftarrow{} \frac{t}{t+1}w+\frac{1}{t+1}e_{i_{t}}$
%         \STATE $t \leftarrow{} t+1$
%    \ENDWHILE
%    \RETURN $w$
% \end{algorithmic}
% \end{algorithm}

\begin{algorithm}[ht]
   \caption{Saddle Frank-Wolfe heuristic for computing generic $\cA\cB$-design}
   \label{alg:algoFW2}
\begin{algorithmic}
   \STATE {\bfseries Input:} arm set $\cA\subset\R^d$, transductive set $\cB\subset\R^d$, maximum iterations $n$
   \STATE  {\bfseries Initialize:} $w \leftarrow{} (1, 1, \ldots, 1)\in\R^A, \tV \leftarrow{} I_d, V \leftarrow{} I_d, t \leftarrow{} 0$
   \WHILE{$t<n$}
        \STATE $\ta\in\argmax_{a\in\cA}  \norm{a}_{V^{-1} \tV V^{-1} }^2$
        \STATE $\tb\in\argmax_{b\in\cB}\normm{b}^2_{V^{-1}}$
        \STATE $V \leftarrow{} V + \ta\ta^\top$
				\STATE $\tV \leftarrow{} \tV + \tb\tb^\top$
        \STATE $w \leftarrow{} \frac{t}{t+1}w+\frac{1}{t+1}e_{\ta}$
        \STATE $t \leftarrow{} t+1$
   \ENDWHILE
   \RETURN $w$
\end{algorithmic}
\end{algorithm}

\paragraph{Entropic mirror descent.}
An entropic mirror descent alternative is used by~\citet{tao2018alba} to compute $\gopt$. The entropic mirror descent approach requires the knowldge of the Lipschitz constant of $\log\det V_w$. Unfortunately, that Lipschitzness property does not seem to hold. \citet{lu2018convex} propose a solution to overcome the Lipschitz issue, but only for $\gopt$-design. Whether it still works for general $\cA\cB$-design remains an open question.

%\begin{algorithm}[ht]
%   \caption{Entropic mirror descent heuristic for computing $\gopt$-design}
%   \label{alg:algoMD}
%\begin{algorithmic}
%   \STATE {\bfseries Input:} arm set $\cA\subset\R^d$, transductive set $\cB\subset\R^d$, tolerance constant $\epsilon$, Lipschitz constant $L$
%   \STATE  {\bfseries Initialize:} $w \leftarrow{} (1/A, 1/A, \ldots, 1/A), t \leftarrow{} 1$
%   \WHILE{$|\max_{a\in\cA}a^\top V_w^{-1} a - d| \geq \epsilon$}
%        \STATE $\gamma \leftarrow{} \frac{2\sqrt{A}}{L} \frac{1}{\sqrt{t}}$
%        \STATE Compute gradient $\nabla^i \leftarrow{} \Tr(V_w^{-1}a_ia_i^\top)$
%        \STATE $w^{a_i} \leftarrow{} \frac{w^{a_i}\exp(\gamma \nabla^i)}{\sum_{i=1}^A w^{a_i}\exp(\gamma \nabla^i)}$
%        \STATE $t \leftarrow{} t+1$
%   \ENDWHILE
%   \RETURN $w$
%\end{algorithmic}
%\end{algorithm}

\end{document}